\documentclass[11pt]{article}
\usepackage[utf8]{inputenc}
\usepackage{amsmath,amssymb,amsfonts,amsthm}
\usepackage{booktabs}
\usepackage{graphicx}
\usepackage{caption}
\usepackage{subcaption}
\usepackage{hyperref}
\usepackage[margin=1in]{geometry}
\usepackage[authoryear]{natbib}
\usepackage{tikz}
\usepackage[edges]{forest}

\newtheorem{proposition}{Proposition}

\newtheorem{corollary}{Corollary}
\newtheorem{definition}{Definition}
       
\title{One Permutation Is All You Need: Fast, Deterministic Feature Importance and Model Stress-Testing}       

\author{Albert Dorador\\
		Hi! PARIS Center -- AI for Science, Business \& Society\\
        \texttt{dorador-chalar@hec.fr}}

\begin{document}
\maketitle

\begin{abstract}
Reliable estimation of feature contributions in machine learning models is essential for transparency, algorithmic fairness, and regulatory compliance, particularly when models are proprietary or otherwise operate as ``black boxes''.  While permutation feature importance is a widely used diagnostic tool, classical implementations rely on repeated Monte Carlo shuffling, introducing significant computational overhead and stochastic instability. In this paper, we show that replacing $B$ random permutations with a single, max-min rank-optimal deterministic permutation maintains or improves correlation with ground-truth feature importance while eliminating estimation variance and reducing model evaluation complexity from $\mathcal{O}(B \cdot n \cdot p)$ to $\mathcal{O}(n \cdot p)$. Under location-scale feature distributions, we formally prove exact recovery of scale-adjusted linear regression coefficients, alongside improved importance estimation under concave model sensitivity relative to random permutations. We then extend this deterministic framework along two complementary dimensions. First, we introduce Systemic Feature Importance (SFI), which integrates empirical feature correlations to quantify indirect feature reliance through proxy variables. Second, we introduce Importance Direction, extending scalar importance to a signed, directional representation by measuring concordance between covariate displacements and output shifts. Extensive empirical validation across nearly 200 simulation scenarios demonstrates superior bias-variance trade-offs in high-dimensional and low signal-to-noise regimes. Finally, two real-world credit risk case studies illustrate how coupling SFI with Importance Direction enables practitioners and regulators to audit models for both the magnitude and sign of hidden reliance on protected attributes, delivering a principled, transparent, and computationally efficient framework for model governance.
\end{abstract}

\bigskip

\section{Introduction}
Feature importance (FI) estimation has emerged as a cornerstone of explainable artificial intelligence (XAI) and responsible machine learning. Across high-stakes domains, from financial risk scoring to clinical, judicial and recruiting decision support, understanding how a model relies on its input variables is essential for model validation, regulatory compliance, and scientific discovery. However, despite its widespread adoption, what precisely ``feature importance'' measures remains a source of ongoing ambiguity in the literature.

\subsection{A Taxonomy of Model-Agnostic Global Feature Importance}

We distinguish between three fundamental notions of FI, each addressing a distinct question:

\begin{itemize}
\item \textbf{Population FI:} ``How important is this feature in the underlying data-generating process?''
\item \textbf{Model Class FI:} ``How replaceable is this feature across a given class of models?''
\item \textbf{Model Instance FI:} ``How much does this specific, fitted model rely on this feature?''
\end{itemize}

All three notions represent valid, complementary perspectives. Crucially, because they answer fundamentally different questions, they require fundamentally different analytical frameworks.

\medskip

The first notion, while intuitively appealing, introduces a major conceptual hurdle: speaking of ``the'' population importance implicitly assumes a unique true model. This identifiability assumption is strongly challenged by the Rashomon effect \citep{Breiman2001b}, which demonstrates that vastly different models can yield virtually identical predictive performance. To reconcile this multiplicity, \citet{Fisher2019} proposed evaluating feature importance across a \textit{Rashomon set} of high-performing models using Breiman-style permutation techniques \citep{Breiman2001}. While theoretically principled, estimating the full Rashomon set is computationally and operationally intensive, and for that reason is seldom adopted in practice. More commonly, practitioners fit flexible non-parametric estimators, such as Random Forests \citep{Breiman2001} or GUIDE trees \citep{Loh2012, Loh2021}, that feature native, internal importance scoring systems. These tools are frequently repurposed for feature selection in predictive pipelines. Conversely, mutual information is a metric typically used for feature selection that can be used in population FI estimation.  This latter approach relies on an information-theoretic metric that quantifies the pairwise statistical dependency between individual covariates and the target variable, independent of any specific predictive model.

\medskip

The second notion ties FI to a specific model architecture, for instance, the class of Random Forests with 500 decision trees. Here, the goal is to quantify how much a specific model class relies on a feature, given the remaining available covariates. Features offering high unique predictive power relative to their peers earn higher scores. The canonical approach for estimating model class FI is Leave-One-Covariate-Out (LOCO) \citep{Lei2018}, which retrains the model class after withholding each feature one at a time and measures the resulting performance drop. While conceptually clear, in a $p$-dimensional feature space, LOCO requires retraining $p$ separate models, making it computationally prohibitive for certain architectures in high-dimensional or large-scale settings. While conditional importance evaluation under standard Gauss-Markov assumptions is well-understood in linear models, non-linear frameworks require efficient approximations. Alternatives include estimating \textit{ghost variables} \citep{Delicado2023}. Rather than retraining the architecture $p$ times as in LOCO, this approach serves as a computationally efficient surrogate for model class FI by replacing each feature with its conditional expectation regressed on the remaining covariates $\mathbb{E}[X_j \mid \mathbf{X}_{-j}]$. Although evaluated through the originally fitted model function, ghost variables target feature replaceability and unique predictive power, aligning them conceptually with LOCO.

\medskip

Given a single, fully trained machine learning model (the ``primary'' model), the third notion of FI evaluates the direct influence of each covariate within that specific fitted function. Because the model architecture and parameters are fixed, methods estimating model instance FI are not concerned with whether a covariate could theoretically be replaced by correlated peers, as doing so would require fitting and ultimately auditing a \textit{different} model. Consequently,  evaluating the \textit{direct} importance of a feature in a fixed model instance requires, in a sense, isolating its effect from background feature correlations.  However, this does not mean that ignoring feature interdependence during permutation is without drawback. Indeed, insight derived from distributionally implausible feature-value combinations should be disregarded, as evaluating models in off-manifold regions can induce extrapolation artifacts \citep{Strobl2008, Hooker2021, Borgonovo2026}.  Nevertheless,  accounting for exact conditional feature distributions often requires high-dimensional density estimation or intensive resampling that can be computationally prohibitive in practice. Standard permutation feature importance (PFI) thus acts as a fast, scalable baseline for model instance FI that preserves marginal distributions but trades off adherence to feature joint distributions for computational efficiency.
\medskip

Figure \ref{fig:taxonomy_tree} summarizes this taxonomy of FI methods across modern machine learning practice.

\begin{figure}[htbp]
\centering
\begin{forest}
  forked edges, % Applies forked edges globally
  for tree={
    draw, rounded corners=3pt,fill=gray!5,
    font=\scriptsize, align=center,
    edge={-stealth, semithick},
    s sep=5pt, % horizontal separation between sibling nodes
    l sep=28pt,  % vertical separation between tree levels
    inner xsep=2pt,  % Tight inner horizontal padding inside boxes
    inner ysep=3pt,     % Inner vertical padding
    fork sep=14pt   % Adjusts the length of the stem before the split
  }
  [Model-Agnostic Global FI
    [Population FI
      [{$\Delta$ Error\\(Rashomon Set)}]
      [{Native Metrics\\(RF / GUIDE)}]
      [{Entropy-based\\(Mutual Info.)}]
    ]
    [Model Class FI
      [{$\Delta$ Error\\(LOCO)}]
      [{$\Delta$ Prediction\\(Ghost Variables)}]
    ]
[Model Instance FI
      [{$\Delta$ Error\\(Classical PFI)}]
      [$\Delta$ Prediction
        [{Direct FI\\(\textbf{DFI} / KernelSHAP)}]
        [{Systemic FI\\(\textbf{SFI})}]
      ] 
    ]
  ]
\end{forest}
\caption{A taxonomy of model-agnostic global feature importance methods categorized by target query and analytical approach. The methods we contribute in this paper are shown in bold.}
\label{fig:taxonomy_tree}
\end{figure}

\subsection{Virtues and Limitations of Classical PFI}

Classical PFI -- which randomly shuffles each feature $X_j$ multiple times to generate a perturbed column $X_j'$ and evaluates the resulting loss in predictive performance (à la Breiman) -- remains among the most widely deployed XAI methods due to several key strengths:
\begin{itemize}
\item \textbf{Conceptual Simplicity:} It directly measures performance degradation under input corruption, revealing how much signal each feature in the model carries.
\item \textbf{Computational Efficiency:} It requires fitting only a single predictive model (the one using all selected covariates).
\item \textbf{Comparability:} It compares predictions on identical model architectures and feature sets with identical marginal distributions, ensuring differences cannot be attributed to changes in those ancillary factors.
\item \textbf{Model Agnosticism:} It applies universally to any tabular model, including black-box models \citep{Fisher2019}.
\item \textbf{Theoretical Guarantees:} It enjoys asymptotic convergence properties via the Law of Large Numbers \citep{Foge2024}.
\end{itemize}

\medskip

Nevertheless, classical permutation approaches pose four intrinsic methodological challenges:
\begin{itemize}
\item \textbf{Stochastic Instability:} Non-deterministic random shuffling introduces seed dependence, yielding inconsistent importance ranks that compromise auditability in high-stakes regulatory settings.  Fixing a seed simply transforms an instability problem into a credibility one, as outcomes can then be to some extent controlled by choosing the seed.
\item \textbf{High Estimator Variance:} Stochastic sampling introduces substantial Monte Carlo variance, which can only be mitigated -- never eliminated -- by running costly repeat permutations ($B \gg 1$).
\item \textbf{Computational Inefficiency:} Permuting every feature $B$ times adds unnecessary computational overhead, with some authors even advocating for $n!$ exhaustive permutations \citep{Fisher2019}.
\item \textbf{Vulnerable Evaluation Metric:} Evaluating importance via performance degradation (e.g., test MSE increase) conflates model reliance with model quality. A feature upon which the model heavily relies might yield a negative importance score if its permutation fortuitously improves a weak model's accuracy, complicating score standardisation and interpretation.
\end{itemize}

The vulnerability of loss-based metrics underscores a fundamental divergence in analytical objectives. When the primary goal is feature selection or performance attribution, evaluating the drop in predictive performance provides valuable insight \citep{Hue2026}.  In such contexts, a negative importance score meaningfully signals a feature that actively degrades generalization. Conversely, when evaluating model instance FI, as defined in this paper, the objective is strictly to quantify internal model mechanics -- auditing what the trained evaluation function actually computes, irrespective of ground-truth correctness. For this purpose, evaluating the absolute shift in predictions (model sensitivity) isolates the model's structural reliance on a covariate, ensuring that heavy dependence on a noisy or detrimental feature is transparently exposed rather than masked by a null or negative performance score.

\medskip

Accordingly, in this work, as it focuses on model instance FI, we adopt absolute prediction shift as our primary metric for global FI. That said, we note that prediction shift and performance degradation represent complementary diagnostic dimensions. Standard loss-based approaches suffer from an inherent identifiability problem: a near-zero performance drop can indicate either that the model ignores the feature entirely, or that the model relies on it without yielding a net accuracy benefit when averaged across the dataset (due to error cancellation or model overfitting). Evaluating prediction shifts immediately removes all ambiguity about which of these two scenarios takes place. While prediction shifts do not explicitly decompose predictive success, feature reliance and performance attribution remain tightly aligned in well-generalized models due to empirical risk minimization. However, when models suffer from overfitting, this alignment breaks down: the model retains high structural reliance on a feature even as its performance attribution vanishes or turns negative. Consequently, the analytical blind spots of the two frameworks are fundamentally asymmetric: a near-zero loss-based score remains ambiguous with multiple confounding interpretations, whereas a high prediction-shift score directly certifies structural reliance and maps to high performance attribution except under overfitting, its singular failure mode.

\medskip

To see why this is the case, consider how error cancellation manifests in loss-based frameworks. At the sample level, a feature may induce opposing loss changes across different instances that sum to zero globally. More subtly, cancellation occurs even at the individual instance level: because standard loss functions (e.g., squared error) are symmetric around the target $y_i$, any perturbation that flips a prediction across $y_i$ while keeping the magnitude of the error constant yields an instance-level loss change of zero, masking significant model reliance that cannot be fixed by applying an absolute value operator at the instance level. In contrast, prediction-shift frameworks apply an absolute displacement operator directly to the output space. Because this quantity is strictly non-negative at every instance, individual prediction shifts cannot cancel out, either across sub-populations or around ground-truth targets, guaranteeing that any non-zero structural reliance accumulates monotonically into the global importance score.

\subsection{Proposed Framework and Key Contributions}

In this paper,  we introduce a Breiman-inspired framework that resolves the fundamental limitations of classical PFI while preserving its core model-agnostic benefits and proposing several extensions. We make three main contributions:
\begin{enumerate}
\item Direct Feature Importance (DFI): By replacing random shuffling with a single, max-min optimal uniform rank-shift permutation ($\pi_{\lfloor n/2 \rfloor}$) our framework achieves full determinism, zero Monte Carlo variance, a single $\mathcal{O}(n \cdot p)$ forward pass, and strictly non-negative scores via an $\ell_p$ prediction-shift metric ($\|\Delta \hat{f}\|_p$). We formally prove several desirable finite-sample properties that this framework possesses, including enhanced estimation fidelity for widely used classes of models such as tree-based architectures and tabular foundation models.

\item Systemic Feature Importance (SFI): We extend DFI by combining our deterministic rank-shift operator with empirical feature correlations that are statistically significant, allowing us to estimate indirect feature reliance through proxy variables. This extension provides a unified methodology for both internal model auditing and systemic model risk stress-testing.

\item Importance Direction: DFI and SFI naturally induce signed directional estimates $\operatorname{Dir}(X_j) \in \{-1, 0, +1\}$. By measuring the average directional agreement between structured covariate displacements ($\Delta x_j$) and induced prediction shifts ($\Delta \hat{f}$), we extend feature importance from unsigned scalar magnitudes to signed, directional representations. When coupled with SFI, this extension provides a critical missing link for algorithmic fairness auditing, exposing both the magnitude and the sign of indirect model reliance on protected attributes.
\end{enumerate}

\subsection{Distinction from Shapley Value Frameworks}

It is instructive to contrast our framework with Shapley value approaches \citep{Shapley1953, Lundberg2017}, which also measure model instance prediction shifts ($\Delta \hat{f}$). While popular for local attributions, constructing a global FI metric via Shapley values introduces severe computational bottlenecks.  Model-agnostic variants like KernelSHAP solve a combinatorial cooperative game across feature coalitions, resulting in an exponential computational complexity ($\mathcal{O}(n \cdot 2^p)$) that often requires Monte Carlo coalition sampling.  Stochastic subsampling of the background dataset introduces additional non-determinism into the resulting global ranks. Evaluating missing feature coalitions also frequently forces model-agnostic SHAP to integrate over marginal feature distributions \citep{Aas2021}, evaluating the model off-manifold -- a necessary trade-off to maintain tractability in some cases.

\medskip

Therefore, the core distinction between these approaches for global (direct) feature importance estimation lies in structural efficiency. Rather than deriving global importance bottom-up through the costly aggregation of localized combinatorial allocations, our framework acts as a native, top-down auditor. By applying a single rank displacement per feature, we isolate global model sensitivity in one deterministic, efficient forward pass ($\mathcal{O}(n \cdot p)$), entirely bypassing stochastic sampling and Monte Carlo approximation error.

%\medskip

Lastly, existing Shapley frameworks cannot cleanly disentangle Direct Feature Importance (DFI) from Systemic Feature Importance (SFI). Because Shapley values allocate a single unified importance score per feature, they force a structural trade-off based on the chosen background distribution. As noted, marginal distributions capture direct reliance but force unrealistic off-manifold evaluations. Conversely, attempting to respect the data manifold via conditional distributions, besides being computationally intensive, inherently conflates DFI with SFI: it mixes direct model reliance with indirect correlation effects into a single metric. Consequently, SHAP cannot explicitly isolate systemic proxy spillovers. Furthermore, while standard local-to-global aggregations like mean absolute SHAP values must discard directional signs to prevent effect cancellation, our framework decouples magnitude from direction entirely, natively outputting a signed, directional representation.

\subsection{Outline of the Paper}

The remainder of this paper is organized as follows: Section~\ref{sec:oneperm} provides the theoretical derivation establishing $\pi_{\lfloor n/2 \rfloor}$ as the max-min optimal deterministic rank permutation and proving its finite-sample estimation properties. Section \ref{sec:methods} formally defines our proposed DFI metrics and outlines our empirical evaluation setup. Section \ref{sec:results} presents benchmark experiments comparing DFI against classical baselines. Section \ref{sec:VItoMST} extends our framework to SFI to capture indirect model reliance via proxy variables.  Section~\ref{sec:direction} introduces Importance Direction, upgrading scalar importance scores to directional importance diagnostics. Section~\ref{sec:casestudies} demonstrates the practical utility of SFI and Importance Direction estimation for model stress-testing and algorithmic fairness auditing across two real-world case studies. Section \ref{sec:conclusion} concludes.

\section{One Optimal Permutation is Worth a Thousand Random Ones}
\label{sec:oneperm}

As we have discussed, traditional permutation-based FI \citep{Breiman2001, Fisher2019} samples $B$ independent permutations uniformly at random for each feature,  computes a measure of model performance degradation each time,  and then averages the results across the $B$ runs, where $B$ in practice is typically at least ten \citep{Alonso2022}, and sometimes more (e.g., 50 or 100).
\medskip

While this method has enjoyed remarkable popularity for the reasons we described previously, its inherent randomness introduces non-zero estimator variance for finite $B$. Formally, this type of estimator of FI is defined as
\begin{equation}
\hat{I}_B = \frac{1}{B} \sum_{b=1}^B g(\pi_b)
\end{equation}

where $g(\pi_b)$ is the importance functional for a single permutation $\pi_b$ (e.g.  MSE increase per feature following the permutation of its values). Thus, under i.i.d. permutation sampling, the variance of this estimator is
\begin{equation}
\operatorname{Var}\left(\hat{I}_B\right) = \frac{\operatorname{Var}(g(\pi))}{B}
\end{equation}

and hence positive for any finite $B$. On the other hand, the variance of a deterministic permutation $\tilde{I} = g(\pi^*)$ is 0,  given the dataset.
\medskip

In addition, it seems clear that not all possible random permutations contribute equally to the estimation of the FI vector: minimally perturbing ones (e.g. only swapping two out of $n$ feature values) contribute less than another one that significantly displaces every single feature value from its original position in rank space. Thus, in general, some of the $B$ permutations are partially wasted. In particular, it is conceivable that a comparable effect could be achieved with a smaller amount of more carefully-chosen permutations.

\subsection{Optimal Deterministic Rank Permutations and MSE Dominance}

Taking these considerations into account,  we wish to replace $B$ random permutations by a single, deterministic and in some sense optimal permutation. By optimal we mean that it perturbs all ranks of a given feature as much as possible and as evenly as possible. This is a reasonable optimality criterion, as we wish to find a permutation (hence a marginal distribution-preserving mechanism) that removes as much signal from the feature at hand as possible, while making sure no local structure is left barely untouched and no individual observation is given an a priori unjustified disproportionate weight.
\medskip

Given a sample size of $n$ observations per feature, we can achieve this goal by shifting the ranks of the feature at hand by $\lfloor n / 2 \rfloor$ (modulo $n$). Let us formalize this idea. We will prove via a max-min argument that this choice of permutation indeed fulfills the optimality criterion that we have intuitively defined before and which we will rigorously define next.

\begin{proposition}[Max-min optimality of $\pi_{\lfloor n / 2 \rfloor}$] \label{prop:optim}
Let $n \geq 2$. For any integers $a, b \in \{0, \ldots,  n-1\}$ define the circular displacement function as
\begin{equation}
d(a, b) := \min(\vert a - b \vert, \, n - \vert a - b \vert)
\end{equation}

For any permutation $\pi$ of $\{0, \ldots, n-1 \}$, define
\begin{equation}
m(\pi) := \min_j d(j, \pi(j)) 
\end{equation}

Then, for every $\pi$ we have $m(\pi) \leq \lfloor n / 2 \rfloor$.

Moreover,  the cyclic shift permutation
\begin{equation}
\pi_k(j) = (j+k) \bmod n
\end{equation}

with $k = \lfloor n / 2 \rfloor$ satisfies $m(\pi_k) = \lfloor n / 2 \rfloor$. Hence $\pi_k$ attains the maximal possible value of the objective function, and is therefore optimal.
\end{proposition}

\textit{Proof sketch.} The maximum possible circular distance between any two indices on $n$ ranks is bounded by $\lfloor n/2 \rfloor$, establishing an immediate global upper bound on $m(\pi)$. The cyclic shift $\pi_{\lfloor n/2 \rfloor}$ attains this bound by holding $d(j, \pi(j))$ strictly constant at $\lfloor n/2 \rfloor$ for all $j$, which forces the minimum shift to equal the upper bound directly (full proof in Appendix \ref{app:pf_optim}).

\medskip

Note that the function $d(a,b)=\min(|a-b|,\,n-|a-b|)$ is the \emph{circular} displacement: it measures distance around the $n$-cycle. For example, if $n=5$ and $j=3$ then the ordinary absolute difference is $|\pi_2(3)-3|=|0-3|=3$, but the circular displacement is $d(3,\pi_2(3))=\min(3,5-3)=2$. Thus the quantity $\lfloor n/2\rfloor$ is indeed the maximal possible circular displacement for every $n$, without requiring $n$ to be even. Indeed, the cyclic shift $\pi_k$ with $k = \lfloor n / 2 \rfloor$ not only maximizes the minimum displacement but makes every circular displacement equal to the maximum possible.
\medskip

Other natural candidate permutations like full rank reversal ($\pi(j) = n - 1- j$) also produce large displacements on average, but they are not equal: middle ranks map to themselves (for odd $n$) or near themselves (for even $n$), which nullifies the permutation effect for those values. Thus,  full rank reversal does not achieve the homogeneity property that we seek, inducing arbitrary non-homogenous contributions from different sample points,  and not making the most efficient use of the entire sample size $n$ at our disposal, as we will formalize later in Proposition \ref{prop:jensens_sensitivity}.
\medskip

Lastly, note that rank shifting can handle categorical features exactly like continuous ones, as all that is needed is to have an ordered sequence of distinct ranks, which is (deterministically) guaranteed even if the values themselves are neither numeric nor distinct.
\medskip

Next we will show that under appropriate conditions a single permutation recovers ground-truth FI scores more accurately in terms of mean squared error (MSE) than traditional FI methods  based on $B$ i.i.d. random permutations,  whether in finite samples or asymptotically.

\begin{proposition}[Component-wise MSE dominance]
\label{prop:MSEdom}
Given a dataset and a fitted model,  fix a feature index $j\in\{1,\dots,p\}$. Let $\mu_j\in\mathbb{R}$ denote the ground-truth importance for feature $j$. 
Let $\hat I_{B,j}$ be the Breiman Monte Carlo estimator for feature $j$ obtained by averaging $B$ i.i.d.\ random permutations, and let $\tilde I_j$ denote a deterministic single-permutation estimator. %(e.g.\ obtained with our rank-shift permutation). 

Write $g_j(\pi)$ for the scalar importance computed for feature $j$ under permutation $\pi$, and let
\[
\sigma_j^2 := \operatorname{Var}_\Pi\big(g_j(\Pi)\big)
\]
be the variance of the importance under a uniformly random permutation \(\Pi\).  Let expectations be taken over the random permutations (and any randomness in the Monte Carlo draw).  Define
\[
\operatorname{MSE}\big(\hat I_{B,j}\big) := \mathbb{E}\big[(\hat I_{B,j}-\mu_j)^2\big],\qquad
\operatorname{MSE}\big(\tilde I_j\big) := \mathbb{E}\big[(\tilde I_j-\mu_j)^2\big].
\]

Then
\[
\operatorname{MSE}\big(\tilde I_j\big) < \operatorname{MSE}\big(\hat I_{B,j}\big)
\]
if and only if
\begin{equation}\label{eq:comp_mse_condition}
\big(\tilde I_j-\mu_j\big)^2 - \big(\mathbb{E}[\hat I_{B,j}]-\mu_j\big)^2 <  \frac{\sigma_j^2}{B}
\end{equation}
%which implies 
%\begin{equation}\label{eq:comp_mse_condition2}
%\big|\tilde I_j-\mu_j\big| - \big|\mathbb{E}[\hat I_{B,j}]-\mu_j\big|
%\;<\; \frac{\sigma_j}{\sqrt{B}}.
%\end{equation}
\end{proposition}

%\textit{Proof sketch.} We apply the bias-variance decomposition to both estimators, noting that the deterministic estimator $\tilde I_j$ has zero variance while the $B$-sample Monte Carlo average $\hat I_{B,j}$ incurs a variance of $\sigma_j^2 / B$. Comparing their total risks directly yields the stated result. The full proof can be found in Appendix \ref{app:pf_MSEdom}. 

\begin{proof}
First note that expectations are taken with respect to the permutation distribution. Thus, $\mathbb{E}[\tilde I_j] = \tilde I_j$ trivially. Now, write the bias-variance decompositions
\[
\operatorname{MSE}(\hat I_{B,j}) = \big(\mathbb{E}[\hat I_{B,j}]-\mu_j\big)^2 + \operatorname{Var}(\hat I_{B,j}),
\qquad
\operatorname{MSE}(\tilde I_j) = \big(\tilde I_j-\mu_j\big)^2 + \operatorname{Var}(\tilde I_j).
\]
Because $\hat I_{B,j}$ is an average of $B$ i.i.d.\ draws of $g_j(\Pi)$ we have $\operatorname{Var}(\hat I_{B,j})=\sigma_j^2/B$. The deterministic estimator has zero sampling variance so $\operatorname{Var}(\tilde I_j)=0$.
Thus the inequality $\operatorname{MSE}(\tilde I_j) < \operatorname{MSE}(\hat I_{B,j})$ is equivalent to
\[
\big(\tilde I_j-\mu_j\big)^2 < \big(\mathbb{E}[\hat I_{B,j}]-\mu_j\big)^2 + \frac{\sigma_j^2}{B}.
\]
%Taking square roots and rearranging gives the final stated form. The displayed inequality \eqref{eq:comp_mse_condition2} follows by applying the elementary bound $\sqrt{a^2+c^2} \le \vert{}a\vert{} + \vert{}c\vert{}$ to $a=\mathbb{E}[\hat I_{B,j}]-\mu_j$ and $c=\sigma_j/\sqrt{B}$, completing the proof.
\end{proof}

\medskip

Proposition \ref{prop:MSEdom} formalizes the intuitive notion we described previously: the single deterministic permutation wins in MSE for a given feature whenever its \emph{excess squared bias} relative to the Monte Carlo estimator is smaller than the Monte Carlo estimator's standard error $\sigma_j^2/B$.  In other words, the bias of our framework could be larger than that of classical PFI and \textit{still} enjoy a lower MSE thanks to avoiding sampling variance. 

\begin{corollary}[Vector (global) MSE dominance]
If one defines the vector-valued estimators $\hat I_B,\tilde I\in\mathbb{R}^p$ and measures MSE in $\ell_2$,
\[
\mathrm{MSE}(\hat I_B) := \mathbb{E}\big[\|\hat I_B-\mu\|_2^2\big],
\qquad
\mathrm{MSE}(\tilde I) := \mathbb{E}\big[\|\tilde I-\mu\|_2^2\big],
\]
and if the permutation-based variances $\sigma_j^2$ are available for each component, then
\[
\mathrm{MSE}(\tilde I) < \mathrm{MSE}(\hat I_B)
\]
if and only if the component-wise inequalities of the proposition combine to give
\[
\sum_{j=1}^p \big(\tilde I_j-\mu_j\big)^2 \;<\;
\sum_{j=1}^p\left(\big(\mathbb{E}[\hat I_{B,j}]-\mu_j\big)^2 + \frac{\sigma_j^2}{B}\right).
\]
\end{corollary}

\begin{proof}
The result follows directly from the definition of $\ell_2$ norm and Proposition \ref{prop:MSEdom}. 
\end{proof}

A few remarks are in order.
\medskip

First, the effect of increasing $B$ as a stabilizing mechanism has only sub-linear effect and exhibits diminishing marginal returns: the cost of each additional permutation stays constant (permuting and evaluating $p$ times the fitted model) while the additional gain decreases.
\medskip

Second,  note that this MSE dominance result does not require the single, deterministic permutation to be optimal in any sense. However, choosing a permutation that is in some sense optimal makes the inequalities in Proposition \ref{prop:MSEdom} more likely to hold in practice for a given dataset.  Concretely, the homogeneous perturbations that our optimal rank shift induces are particularly effective. We formalize this argument in Proposition \ref{prop:jensens_sensitivity}.
\medskip

%Third,  in practice, given a dataset and a fitted model, one can decide whether the deterministic estimator appears preferable for the target budget $B$ by checking whether $||\tilde{I} - \hat{I}_B||_2^2  < ||\hat \sigma^2||_2^2 / B$, where for each $j=1,\ldots,p$,  $\hat \sigma_j^2 := \sum_b(g(\pi_b) - \hat{I}_B)^2 / (B-1)$. This comparison is conservative in the sense that it implicitly gives the Monte Carlo method a head-start by supposing it is arbitrarily close in expectation to the ground-truth  vector $\mu$.
%\medskip

Lastly, it is possible to state and prove analogous results whose assumptions and, hence, results hold only asymptotically as the sample size $n$ grows. In such case,  the single-permutation method may have asymptotically lower MSE than the Monte Carlo method provided that the absolute bias of the former decays faster than that of the latter as $n$ increases. If $\sigma$ decays with $n$, it would also need to be accounted for in that comparison.

\subsection{Diagnostic Sensitivity and Estimation Fidelity}

Besides MSE dominance,  we can show that our proposed approach, under a concave response function,  extracts a greater signal than a random permutation does.

\begin{proposition}[Sensitivity dominance under concave response and $\ell_1$ scoring]
\label{prop:jensens_sensitivity}
Let $f$ be a fitted predictive model and let $x_j \in \mathbb{R}^n$ be a feature vector with ranks $r_1, \dots, r_n$. Define the instance-level prediction shift function as $h(\Delta_i) = |f(x_i) - f(x_i + \Delta_i)|$, where $\Delta_i = x_{\pi(i),j} - x_{i,j}$ is the physical displacement induced by permutation $\pi$. Assume $h(\cdot)$ is non-decreasing with $h(0) = 0$.

Let $g(d_i) := \mathbb{E}_{\Delta \mid d_i}[h(|\Delta_i|)]$ denote the expected prediction shift for a given rank displacement $d_i = |r_i - r_{\pi(i)}|$ over the empirical feature distribution. Assume $g:[0, n-1] \to \mathbb{R}_{\ge 0}$ is strictly concave.

Let $\Pi$ denote an equiprobable random permutation and $\Pi^* = \pi_{\lfloor n/2 \rfloor}$ be the deterministic max-rank shift permutation. Then, for any $n \ge 2$, the expected $\ell_1$ feature importance score (signal strength)  under random permutations is strictly upper-bounded by the score under $\Pi^*$:
\begin{equation}
\mathbb{E}_{\Pi}[I_{\ell_1}(\Pi)] < I_{\ell_1}(\Pi^*)
\end{equation}
\end{proposition}

\textit{Proof sketch.} Under a random permutation, the expected mean rank displacement across instances evaluates to $\mu_{\Pi} = \frac{n^2-1}{3n}$. Applying Jensen's inequality twice bounds the expected score of the random permutation strictly above by $g(\mu_{\Pi})$, which by monotonicity is  dominated by the constant max-shift score $g(\lfloor n/2 \rfloor)$ of $\Pi^*$ for all $n \ge 2$ (full proof in Appendix \ref{app:pf_jensens}).
\medskip

Proposition~\ref{prop:jensens_sensitivity} establishes that under saturating (concave) model responses, classical random permutations suffer from a systematic attenuation bias driven by two distinct mechanisms. First, the expected rank displacement under random shuffling ($\approx n/3$) is strictly smaller than the uniform displacement ($\approx n/2$) guaranteed by $\Pi^*$, as shown in the proof. Second, random permutations introduce high variance into rank displacements wasting evaluation budget on instances with minimal shift. Due to the concavity of $g(\cdot)$, this sensitivity lost on small perturbations cannot be compensated by larger perturbations in other instances.
\medskip

Crucially, this systematic attenuation is non-homogeneous across features. This implies it cannot be remedied by normalizing importance scores to sum to unity. Consider a saturating model $f(\mathbf{x}) = \sigma(w_1 x_1 + w_2 x_2)$ with $|w_1| > |w_2|$ and arbitrary (possibly identical) marginal feature distributions, where $\sigma(\cdot)$ is the standard logistic cdf.  Because $x_1$ exerts a larger influence on model activations, it pushes instances further into the flattening tails of $\sigma(\cdot)$, imparting stronger effective concavity on $g_1(d)$ relative to $g_2(d)$ given a perturbation of the same size.  Consequently, random permutations tend to inflict a harsher Jensen attenuation penalty on the features with higher model reliance. Normalizing random permutation scores bakes this non-homogeneous attenuation into the final vector, artificially compressing (flattening) the relative ratio  $\hat{I}_1 / \hat{I}_2 < \mu_1 / \mu_2$ of estimated to true importance scores.  By eliminating rank-displacement variance entirely,  $\pi_{\lfloor n/2 \rfloor}$ mitigates score compression, reflecting the true difference in feature contributions with greater fidelity.
\medskip

Given that modern tabular predictive modeling is dominated by tree-based ensembles (e.g., XGBoost, Random Forests) and, increasingly, tabular foundation models (TFMs such as TabPFN or TabICL), assuming a concave expected sensitivity $g(d)$ is well-justified in practice. For classical tree-based models, the prediction function $f(\mathbf{x})$ is constructed via orthogonal step-functions over feature space partitions. Once a displacement magnitude $|\Delta|$ is large enough to push a feature value across the outer split boundaries of a tree path, the instance falls into a static terminal leaf node. Beyond these boundary thresholds, further increases in perturbation size yield zero additional shift in predictions. When aggregated over an ensemble of trees, this hard threshold ceiling yields an expected sensitivity curve $g(d)$ with diminishing marginal returns. A similar saturation occurs in modern deep learning architectures and TFMs. These transformer-based systems process tabular data by routing feature representations through self-attention layers governed by the softmax operator. As localized perturbations push feature values outside their typical dense support, the corresponding exponential attention weights asymptote sharply toward $0$ or $1$. Once these weights saturate, local output shifts flatten out. Consequently, both classical tree ensembles and modern TFMs exhibit diminishing marginal returns to perturbation magnitude, making the uniform maximization of displacement under our framework a theoretically optimal sampling design from this perspective as well.

\medskip

Next we show that uniform shifts, besides better preserving score differentials under concavity, are instrumental to unbiased parameter recovery in linear regression, i.e.,  when the primary model is linear in parameters.

\begin{proposition}[Recovery of normalized, scale-adjusted linear regression coefficients]
\label{prop:linear_recovery}
Let $f(\mathbf{x}) = \sum_{j=1}^p \beta_j x_j$ be a linear model over $p$ features whose marginal distributions belong to a common location-scale family. For each feature $j \in \{1, \dots, p\}$, let $\operatorname{MAD}(X_j) := \frac{1}{n}\sum_{i=1}^n |x_{i,j} - \bar{x}_j| > 0$ denote its empirical mean absolute deviation, and let $\pi_j = \pi_k$ be any deterministic, uniform rank-shift operator with shift $k \in \{1, \dots, n-1\}$, yielding mean physical displacement $c_j := \frac{1}{n}\sum_{i=1}^n |x_{i,j} - x_{\pi_j(i), j}|$. Under $\ell_1$ prediction-difference scoring, the normalized feature importance score satisfies:
\[
\tilde{I}_j = \frac{|\beta_j^*|}{\sum_{m=1}^p |\beta_m^*|},
\]
where $\beta_j^* = \beta_j \cdot \operatorname{MAD}(X_j)$ is the scale-adjusted linear regression coefficient.
\end{proposition}

\textit{Proof sketch.} By linearity of $f$, the raw unnormalized importance score factors into $I_{\ell_1, j} = |\beta_j| c_j$. Because all features belong to a common location-scale family, rank-space quantile shifts map to physical displacements strictly proportional to scale, yielding $c_j \propto \operatorname{MAD}(X_j)$. Upon sum-to-one normalization, the proportionality constant is common to all features,  canceling out across numerator and denominator, recovering the exact ratio $\tilde{I}_j = |\beta_j^*| / \sum_k |\beta_k^*|$. Full proof in Appendix \ref{app:pf_linear_recovery}.

\medskip

\textbf{Remark (Necessity of uniform rank-shifting over random permutations):} One might naturally ask whether standard random permutations also satisfy $c_j \propto \operatorname{MAD}(X_j)$.  Under random shuffling, the expected physical displacement corresponds to the Gini Mean Difference, $\mathbb{E}[c_j] = \operatorname{GMD}(X_j) := \mathbb{E}[|X_j - X_j'|]$, which is indeed proportional to $\operatorname{MAD}(X_j)$. Under the corollary's assumption of a shared location-scale family, the constant of proportionality is shared across features, leading to the same cancellation effect,  in expectation.
\medskip

However, random shuffling still suffers from finite-sample Monte Carlo variance, inducing stochastic fluctuations that distort the coefficient ratio for any finite number of permutations $B$. In contrast, our deterministic rank-shift $\pi_{\lfloor n/2 \rfloor}$ guarantees exact scale proportionality in a single evaluation ($B=1$) with zero Monte Carlo noise.

\medskip

\textbf{Remark (Extension to non-linear models):}
\label{rem:nonlinear_extension}
While Proposition~\ref{prop:linear_recovery} formally proves exact parameter recovery under linear models, the advantages of the deterministic, uniform rank-shift $\pi_{\lfloor n/2 \rfloor}$ over random shuffling extend naturally to smooth non-linear architectures $\hat{f}(\mathbf{x})$. For these model classes (e.g., neural networks or Tabular Foundation Models), a first-order Taylor expansion around each instance $\mathbf{x}_i$ decomposes the global expected prediction shift into a first-order sensitivity signal and a higher-order residual:
\[
I_{\ell_1, j} = \frac{1}{n}\sum_{i=1}^n |\hat{f}(\mathbf{x}_{i, \pi_j}) - \hat{f}(\mathbf{x}_i)| = \frac{1}{n}\sum_{i=1}^n \left| \left.\frac{\partial \hat{f}}{\partial x_j}\right|_{\mathbf{x}_i} (x_{\pi_j(i), j} - x_{i, j}) + \mathcal{O}\left(|x_{\pi_j(i), j} - x_{i, j}|^2\right) \right|.
\]
Under standard random shuffling, paired instances sample quantile differences $|u_i - u_{\pi(i)}|$ uniformly across $[0, 1]$. In features with heavy tails, extreme tail-to-tail pairings ($|u_i - u_{\pi(i)}| \approx 1$) produce massive feature value perturbations $|x_{\pi(i), j} - x_{i, j}| \gg 0$, pushing instances into sparse, out-of-distribution regions of the joint feature space (severe off-manifold extrapolation).  
\medskip

Because the Taylor residual grows quadratically with perturbation distance, the severe shifts induced by even a small fraction of extreme tail-to-tail jumps may disproportionately inflate the global empirical mean. Consequently, standard random permutations cause FI scores to reflect extrapolation artifacts alongside true model reliance, to a more severe extent than a uniform rank-shift permutation. Indeed, our proposed $\pi_{\lfloor n/2 \rfloor}$ permutation fixes quantile displacement at exactly $50\%$ along the empirical cumulative distribution function. By avoiding extreme tail-to-tail pairings ($|u_i - u_{\pi(i)}| > 0.5$), $\pi_{\lfloor n/2 \rfloor}$ bounds the physical perturbation magnitude across all observations, reducing quadratic residual inflation and mitigating off-manifold extrapolation artifacts in FI scores.

\section{Methods}
\label{sec:methods}

In this section we formally define our notion of Direct Feature Importance (DFI), and then describe more specific design choices for our proposed FI method, as well as our empirical simulation setup. The direct importance of a feature within a model is understood as the expected immediate change in prediction after the signal in a feature is removed by permuting its values. We formalize this notion next.

\begin{definition}[Direct Feature Importance] \label{def:DFI}
Given the evaluation function of a trained model $M$ denoted by $f_M: \mathbb{R}^{n\times p}\to \mathbb{R}^{n\times q}$ ($q=1$ in standard regression, or $q=C$ in $C$-class classification),  and a dataset $D = (X,y)$ where $X \in \mathcal{S}_X \subseteq \mathbb{R}^{n\times p}$ and $y \in \mathcal{S}_y$,  the estimated direct importance of the $k$-th variable $x_k$ is defined as
\begin{equation}
d_k := \frac{1}{n q} \sum_{i=1}^{nq} d(f_{M_i}(X) - f_{M_i}(X_{k'}))
\end{equation}

where $d(a,b)$ is a suitable metric (e.g.  $d(a,b) = |a-b|$) that can be applied element-wise,  and $X_{k'}$ denotes $X$ replacing $x_k$ by its permuted representation $x_k'$. Lastly, final direct scores are normalized to sum to one.
\end{definition}

We deliberately do not require having access to a test set where the feature permutations and subsequent model evaluations are carried out: access to training data is enough, as the model parameters are estimated using the training set and fixed thereafter, and only prediction shifts are of consequence to estimate DFI for that specific trained model.  Common practice often requires using a test set, but this belief stems from the confusion surrounding the definitions of FI (for instance, whether the model overfits is irrelevant to estimate its reliance on a given feature). Nonetheless, we follow the convention to use a test set to estimate FI in all our experiments so that our results cannot possibly be attributed to a non-standard choice of evaluation set (e.g. different sample sizes or feature distributions).

\subsection{Algorithm Design Choices}

\subsubsection{Choice of Deterministic Permutation}

A key design aspect is the choice of deterministic permutation, as it has a direct impact on the FI scores. Following our theoretical discussion in Section \ref{sec:oneperm}, we consider two types of permutations: a \textit{rank} shift of $\lfloor n / 2 \rfloor$ modulo $n$ (optimal) and an \textit{index} shift of $\lfloor n / 2 \rfloor$ modulo $n$ too (sub-optimal).
\medskip

Indeed, the rank-shift version is optimal as shown in Proposition \ref{prop:optim}, but it requires sorting a length-$n$ array, which takes $\mathcal{O}(n\log n)$ time, while an approximate solution based on indices instead of ranks takes only $\mathcal{O}(n)$ time.
\medskip

In our extensive simulations, the difference in accuracy between the optimal and sub-optimal approaches tends to be small, as is their difference in compute time -- even with 10,000 samples.  We show that either single-permutation scheme is at least as accurate as traditional random permutation schemes (and often more accurate in challenging regimes), while being at least an order of magnitude faster than the classical 10-repetition permutation-based methods (and competitive with single random permutation FI scoring).

\subsubsection{Evaluation Metric}

We consider different scoring metrics to measure the impact that permuting a feature has on prediction: mean absolute error (MAE, the default), MSE and Root MSE (RMSE). Formally, in the case of MAE or MSE, for each feature we compute $(nq)^{-1} \sum_{i=1}^{nq} d(\hat y_i - \hat y_i')$,  where $d(x) = |x|$ with MAE and $d(x) = x^2$ with MSE,  while $\hat y_i$ and $\hat y_i'$ are the primary model's predictions for the $i$-th observation before and after permuting a given feature, respectively (in $C$-class classification, they are length-$C$ vectors).  Importance scores are then normalized to sum to one. In the case of RMSE, the formula we stated is adjusted by simply taking its square root.
\medskip

In order to isolate the effect of replacing multiple random permutations in traditional FI methods by a single, deterministic permutation in the FI method we propose, we will use the MSE of the prediction differences as scoring metric (instead of our MAE default), just like the classical baseline does in regression.  In classification settings,  we still compute the MSE of the difference in class probabilities, but, for comparability purposes,  we deviate from the default scoring metric used by the baseline (drop in classification accuracy) and replace it by the change in (negative) Brier score, which is the MSE of predicted class probabilities compared to their true observed class.
\medskip

Using the default scoring metrics instead (MAE for our methods, MSE / classification accuracy for the baseline) would in fact  benefit our proposed methods compared to the classical baseline (see Appendix \ref{ap:res_mae}), which out-of-the-box show significantly better ground-truth score recovery than scikit-learn's implementation of Breiman-style permutation across nearly 200 scenarios.
\medskip

Unlike other evaluation metrics that are often used (e.g. difference in MSE or classification accuracy before and after permuting a feature) the MAE/MSE/RMSE of the prediction differences ensures non-negative scores that can easily be normalized to a convenient, context-independent scale (e.g.  the closed interval [0,1]) and remain maximally granular regardless of the primary task (regression or classification).

% THE FOLLOWING SUBSECTION IS COMMENTED OUT FOR IMPROVED PAPER FOCUS

%\subsubsection{Optional Feature Prescreening}
%
%Our method optionally supports a prescreening step that detects features whose perturbation has no measurable effect on the model's predictions. While beneficial in settings with many noise variables that can be identified and discarded going forward,  this step involves evaluating a single, deterministic perturbation per feature -- conceptually similar to a lightweight, structured permutation. To avoid any ambiguity in the empirical results section and to ensure a completely ``pure'' one-permutation framework, we disabled prescreening in all experiments comparing our direct importance method to Breiman's classical permutation importance. This guarantees that both methods operate on the full set of features, without any prior filtering, and that our reported advantages do not stem from auxiliary preprocessing steps. The resulting comparison is therefore strictly conservative and isolates the core mechanistic differences between the two approaches.

\subsection{Empirical Simulation Setup}

\subsubsection{Scenarios Considered}

We consider 192 individual scenarios in total, with varying sample size $n$, dimensionality $p$, noise level $\sigma_\varepsilon$ (Gaussian standard deviation), feature correlation $\rho$ and choice of ground-truth linear/nonlinear regression or classification model, each repeated 50 times. Concretely, we consider all combinations that result from the parameter values below, both for regression and binary classification:
\begin{itemize}
\item $n \in \{100, 1000, 10000 \}$
\item $p \in \{10, 100 \}$
\item $\sigma_\varepsilon \in \{0.1, 5 \}$
\item $\rho \in \{0, 0.3 \}$, see Section \ref{sec:corfeats} for more details
\end{itemize}

For the linear response case, we fix a random seed (123) and randomly sample from a multivariate Gaussian distribution with covariate matrix $C$ (the identity matrix in the $\rho=0$ case), and define our response variable as a randomly chosen (seeded) linear function of the design matrix obtained with each combination of the above parameters. We set half of the $p$ covariates as noise features (unrelated with the response).
\medskip

For the nonlinear response case we follow the process described in the next subsection to generate correlated uniformly-distributed features, and use a highly nonlinear function known as ``Friedman function" that is often used for benchmarking purposes. This function is defined as
$$y\equiv f(x_1,x_2,x_3,x_4,x_5) = 10\sin(\pi x_1x_2) + 20(x_3-0.5)^2 + 10x_4 + 5x_5$$
for $ x_j \overset{i.i.d.}{\sim} Unif[0,1]$ for all $j$, although in addition we also consider the case of correlated covariates (see Section \ref{sec:corfeats}). Since the standard Friedman function is defined for exactly five covariates, the remaining number of covariates are set as noise features, which gives the opportunity of testing a much sparser regression problem in the $p=100$ regime.
\medskip

In the case of classification, the response variable is simply binarized as follows: $y_{bin} = 1$ if $y > \text{median}(y)$ and 0 otherwise.
\medskip

Thus, the data-generating process is conceptually consistent across linear/nonlinear and regression/classification scenarios, with the response variable adjusted accordingly to fit each scenario.
\medskip

Lastly, being able to establish the ``ground-truth" FI scores requires using primary models that inherently provide this information. For that reason, we consider generalized linear models with identity or logit link (respectively, for regression or classification) and with or without $\ell_1$ regularization to check if the FI methods considered can estimate well zero or small scores too, respectively. In all cases, we standardize covariates to be on the same scale and then define ground-truth (direct) importance scores as the absolute value of the coefficient estimates, which are then normalized to sum to unity.

\subsubsection{FI Methods Considered}

We test both variants of our proposed method: the one using the optimal rank-shift permutation, and the one using the simpler index-shift permutation, which acts as a deterministic, computationally cheaper approximation to the optimal permutation scheme. As baseline we consider the classical Breiman-style permutation importance methods implemented in standard Python libraries like scikit-learn, both using one random repetition (the fastest possible setup) and the much more common default of ten \citep{Alonso2022}. We use the same MSE-based scoring metric in all cases so that differences in results cannot be attributed to differences in scoring metrics. However, for completeness, we include in Appendix \ref{ap:res_mae} the results obtained using default scoring metrics, which are even more favorable to our proposed methods.

\subsubsection{Correlated Features Design}
\label{sec:corfeats}

In order to evaluate the performance of our proposed method relative to Breiman's under correlated features, we consider a block covariance matrix with stronger correlation among informative features ($\rho = 0.3$) to reflect their shared underlying factor structure,  weaker correlation among noise features ($\rho = 0.3/2 = 0.15$), and zero cross-correlation between the informative and uninformative feature blocks. This type of covariance design is common in the literature, see e.g. \cite{Loh2012}. 
\medskip

Therefore,  we intentionally included (rather mild) correlations among noise features, which can slightly affect ordinary least squares (OLS) coefficient estimates and, consequently, the performance of FI methods. This choice creates challenging scenarios that act as a stress-test for FI approaches: while Breiman's methods exhibit substantially reduced accuracy under these slightly hardened conditions, our DFI methods consistently recover the true feature importance. By incorporating correlated noise features, we not only reflect a plausible structure encountered in real datasets, but also provide a less idealized evaluation framework in which the robustness of our methods is clearly demonstrated.
\medskip

To generate correlated uniform features on $[0,1]$, we first construct correlated Gaussian variables and then apply the probability integral transform. Let $C \in \mathbb{R}^{p \times p}$ denote our custom covariance matrix. We obtain its Cholesky factorization $C = L L^\top$, where $L$ is lower triangular. If $X \in \mathbb{R}^{n \times p}$ has independent standard Gaussian entries, $X_{ij} \sim \mathcal{N}(0,1)$, then $Z = X L^\top$ has rows $Z_i \sim \mathcal{N}(0, C)$.
This follows from
\[
\mathrm{Var}(Z)
= \mathrm{Var}(X L^\top)
= L \, \mathrm{Var}(X) \, L^\top
= L I L^\top
= C,
\]
showing that the linear transformation $L^\top$ injects the desired correlation structure into otherwise independent Gaussian features.
\medskip

To obtain correlated uniform variables, we apply the standard normal CDF componentwise,
\[
U_{ij} = \Phi(Z_{ij}),
\]
which yields $U_{ij} \sim \mathrm{Unif}(0,1)$ by the probability integral transform. Importantly, $\Phi$ is strictly increasing, and monotone transformations preserve all rank-based dependence measures (e.g.  Spearman or Kendall correlations). Thus, the uniform variables $U$ retain the dependence structure implied by $C$ while matching the  marginal distributional assumptions of the Friedman benchmark function.

\subsubsection{Ensuring Breiman's FI Scores are Non-Negative}

As we mentioned previously, classical permutation-based methods measure how much an external performance metric (e.g., mean squared error or classification accuracy) deteriorates when a feature is permuted; we refer to this as a \emph{performance-drop} score. While appropriate when the goal is to quantify contribution to a chosen performance metric,  scores based on performance drops can be unstable and even negative in finite samples: permuting a feature may, by chance or due to model weakness (both due to under- or over- fitting), \emph{improve} the metric. Such negative values may be hard to interpret and complicate normalization to a chosen context-independent scale (e.g.  [0,1]).
\medskip

Due to its widespread use, our chosen Breiman-style implementation of permutation FI is scikit-learn's implementation.  Because our ground-truth importance vectors are non-negative and normalized to sum to one, a compatible representation of Breiman-style importance scores requires transforming these raw effects into the same scale.
\medskip

To remain faithful to the widely used Breiman-style permutation framework, we adopt only the minimally necessary modification to ensure meaningful downstream comparisons -- such as maximum discrepancy with ground-truth importances (which could otherwise exceed 1). Specifically, we set negative permutation effects to zero before normalization.
%This minor post-processing step is conservative and fully consistent with Breiman's intent: it is reasonable to assume the goal of permutation importance was not to reflect cases where random perturbations \emph{improve} predictive performance.
\medskip

Formally, letting $\mathcal{B}_j$ denote the raw permutation effect for feature $j$ à la Breiman, we define
\[
\tilde{\mathcal{B}}_j := \frac{\max\{\mathcal{B}_j, 0\}}{\sum_{k=1}^p \max\{\mathcal{B}_k, 0\}}.
\]
This transformation preserves all positive information while removing negative scores that complicate homogeneous score normalization. This yields well-behaved importance profiles in $[0,1]^p$ and prevents pathological maximum-discrepancy values exceeding $1$, allowing meaningful comparison with ground truth, including bounded  metrics.

\subsubsection{Result Aggregation Across Scenarios}

To aggregate results across scenarios while accounting for both within-scenario variability and between-scenario differences, we compute a combined standard error using a well-established variance-decomposition approach. For each scenario we first obtain its mean estimate and its empirical variance from repeated simulations (50 times each scenario). The total variance across all scenarios is then defined as
\[
\mathrm{Var}_{\text{combined}}
:= \underbrace{\mathbb{E}\!\left[\mathrm{Var}_{\text{within},i}\right]}_{\text{average within-scenario variance}}
\;+\;
\underbrace{\mathrm{Var}\!\left(\mu_i\right)}_{\text{between-scenario variance}}
\]
where $\mu_i$ is the estimate for scenario $i$. The first term captures random fluctuations within each scenario, while the second captures systematic differences in the scenario means. The square root of this total variance gives the combined standard error, and reported error bands use $\pm 2$ times this value. This approach ensures that the final uncertainty quantification reflects both sources of variability, producing a single coherent measure of precision for the aggregated performance metrics.

\section{Results}
\label{sec:results}

\subsection{Regression}

\subsubsection{Linear Response}

\begin{table}[h]
\centering
\caption{Linear regression regime.  Global mean results across 48 scenarios with varying sample size, dimensionality, noise level, feature correlation and choice of ground-truth regression model.}
\label{tab:linreg}
\begin{tabular}{lcccc}
\toprule
Metric & DFI-Opt & DFI-Approx & Breiman (B=1) & Breiman (B=10) \\ \midrule
Ground-truth cor & 0.953 ± 0.008 & 0.941 ± 0.012 & 0.900 ± 0.037 & 0.910 ± 0.035 \\
Max score diff & 0.082 ± 0.019 & 0.093 ± 0.020 & 0.095 ± 0.024 & 0.091 ± 0.024 \\
Time (ms) & 24.9 ± 13.7 & 19.4 ± 10.7 & 17.1 ± 5.3 & 98.4 ± 32.4 \\ \bottomrule
\end{tabular}
\end{table}

Table~\ref{tab:linreg} shows the global results for linear regression across 48 scenarios.  They show that, under a linear regression regime (linear response estimated by OLS) the proposed direct feature importance approach is both more accurate and robust than classical Breiman-style permutation methods: both the method with optimal rank shift and its approximation (using a simpler index shift) display higher correlation with ground-truth scores on average, and beyond the corresponding two-standard-error bands in the case of the former. Maximum score discrepancies across features are on average comparable or slightly lower with our proposed methods as well, although in this case their advantage over classical baselines is not as significant. As Figures \ref{fig:VIS_lm_uncorrelated}, \ref{fig:VIS_Lasso_uncorrelated}, \ref{fig:VIS_lm_corr03}, and \ref{fig:VIS_Lasso_corr03} illustrate, our proposed methods particularly shine (relative to the classical methods) under challenging circumstances: small sample size $n$,  high-dimensionality ($n \approx p$),  low signal-to-noise, or correlated-feature scenarios. This shows that while in idealized conditions both our methods and the classical baselines are similarly accurate, in less ideal scenarios the variability injected by random permutations tends to compromise performance, on average.
\medskip

Despite their superior accuracy -- especially in challenging conditions -- our proposed methods remain efficient (on average, competitive with single-permutation Breiman's method).
\medskip

Overall, our deterministic DFI methods are capable of recovering true feature importance patterns more accurately and, above all, more robustly across scenarios, with comparable or lower computational cost,  and with full seed-free reproducibility.

\subsubsection{Nonlinear Response}

For the Friedman benchmarks we add i.i.d.\ Gaussian observation noise with standard deviation $\sigma_{\varepsilon} \in \{0.1, 5\}$.

\begin{table}[h]
\centering
\caption{Nonlinear regression regime.  Global mean results across 48 scenarios with varying sample size, dimensionality, noise level, feature correlation and choice of ground-truth regression model.}
\label{tab:nonlinreg}
\begin{tabular}{lcccc}
\toprule
Metric & DFI-Opt & DFI-Approx & Breiman (B=1) & Breiman (B=10) \\ \midrule
Ground-truth cor & 0.953 ± 0.007 & 0.941 ± 0.008 & 0.899 ± 0.036 & 0.910 ± 0.034 \\
Max score diff & 0.154 ± 0.019 & 0.176 ± 0.021 & 0.173 ± 0.021 & 0.168 ± 0.023 \\
Time (ms) & 25.3 ± 13.9 & 19.7 ± 11.0 & 17.0 ± 5.1 & 98.2 ± 31.8 \\ \bottomrule
\end{tabular}
\end{table}

As Table~\ref{tab:nonlinreg} illustrates, the results under nonlinear regression are very similar to those under the linear setup. While maximum score differences increase for all methods, our proposed method using a single, deterministic and max-min optimal permutation remains the most accurate and robust on all accounts, while being significantly faster than the common 10-repetition Breiman-style permutation method.
\medskip

Figues \ref{fig:VIS_lm_uncorrelated_NL},  \ref{fig:VIS_Lasso_uncorrelated_NL},  \ref{fig:VIS_lm_corr03_NL}, and \ref{fig:VIS_Lasso_corr03_NL} show that individual per-scenario performance again mirrors the linear regime: our proposed single-permutation methods and the classical baselines are similarly accurate in idealized conditions, while the gap widens significantly in favor of our methods in more challenging (yet realistic) conditions.

\subsection{Classification}

\subsubsection{Linear Response}

We next evaluate all methods in a binary classification setting, keeping the same linear data-generating process as in the regression experiments and binarizing the response as described previously. A logistic regression model serves as the primary model, and importance scores are computed exactly as in the regression setting.

\begin{table}[h]
\centering
\caption{Classification (linear response).  Global mean results across 48 scenarios with varying sample size, dimensionality, noise level, feature correlation and choice of ground-truth classifier.}
\label{tab:linclass}
\begin{tabular}{lcccc}
\toprule
Metric & DFI-Opt & DFI-Approx & Breiman (B=1) & Breiman (B=10) \\ \midrule
Ground-truth cor & 0.962 ± 0.010 & 0.947 ± 0.018 & 0.868 ± 0.052 & 0.882 ± 0.051 \\
Max score diff & 0.066 ± 0.019 & 0.067 ± 0.018 & 0.088 ± 0.026 & 0.083 ± 0.029 \\
Time (ms) & 27.1 ± 14.7 & 21.2 ± 11.5 & 26.2 ± 8.0 & 189.1 ± 60.4 \\ \bottomrule
\end{tabular}
\end{table}

Table~\ref{tab:linclass} reports global performance aggregated across all linear-response classification scenarios. The proposed \emph{DFI-Opt} and \emph{DFI-Approx} methods remain highly accurate, achieving ground-truth correlations of $0.962$ and $0.947$, respectively, substantially outperforming Breiman's permutation importance with $B=1$ ($0.868$) and even $B=10$ ($0.882$).  Maximum score differences show a similar although less pronounced pattern, with our methods remaining below $0.07$ on average, compared to $0.09$ for Breiman with $B=1$.  
\medskip

Runtime comparisons also remain favorable: for $B=1$, Breiman's speed is comparable to our methods,  whereas for $B=10$ it becomes an order of magnitude slower. Overall, these results confirm that the proposed direct importance estimators extend naturally to classification, preserving both accuracy and computational efficiency while avoiding the instability that other permutation-based methods such as Breiman's exhibit even in simple linear setups.
\medskip

Figures \ref{fig:VIS_lm_uncorrelated_CLASS}, \ref{fig:VIS_Lasso_uncorrelated_CLASS}, \ref{fig:VIS_lm_corr03_CLASS}, and \ref{fig:VIS_Lasso_corr03_CLASS} illustrate that in challenging conditions the performance of the classical baselines can deteriorate substantially, reaching low correlation values with ground-truth scores (below 0.40) while our optimal method maintains correlations around 0.90 in those same circumstances.

\subsection{Nonlinear Response}

\begin{table}[h]
\centering
\caption{Classification (nonlinear response).  Global mean results across 48 scenarios with varying sample size, dimensionality, noise level, feature correlation and choice of ground-truth classifier.}
\label{tab:nonlinclass}
\begin{tabular}{lcccc}
\toprule
Metric & DFI-Opt & DFI-Approx & Breiman (B=1) & Breiman (B=10) \\ \midrule
Ground-truth cor & 0.953 ± 0.009 & 0.942 ± 0.013 & 0.863 ± 0.047 & 0.879 ± 0.045 \\
Max score diff & 0.153 ± 0.021 & 0.162 ± 0.020 & 0.173 ± 0.022 & 0.168 ± 0.023 \\
Time (ms) & 27.0 ± 14.6 & 20.9 ± 11.4 & 26.0 ± 7.8 & 185.1 ± 57.3 \\ \bottomrule
\end{tabular}
\end{table}

Table \ref{tab:nonlinclass} shows that the results with a logistic regression primary model used to forecast a nonlinear (binarized) response are similar to the linear response case. However, as we observed in the nonlinear regression case as well,  maximum score differences are inflated, but our proposed methods remain the most accurate in this respect too.
\medskip

In terms estimation efficiency, our methods remain competitive with 1-repetition Breiman importance, while being an order of magnitude faster compared with the more common 10-repetition setup.
\medskip

See Figures \ref{fig:VIS_lm_uncorrelated_NL_CLASS}, \ref{fig:VIS_Lasso_uncorrelated_NL_CLASS}, \ref{fig:VIS_lm_corr03_NL_CLASS}, and \ref{fig:VIS_Lasso_corr03_NL_CLASS} for individual scenario results.

\subsection{Global Result Conclusions}

\subsubsection{Accuracy}

Our proposed methods are on average at least as accurate as Breiman-style FI methods across nearly 200 scenarios, while substantially more so in the more challenging scenarios tested, confirming a superior bias-variance tradeoff.  The difference in mean accuracy is smallest in linear regression problems (0.953 vs. 0.900), while it is largest in binary classification (0.962 vs. 0.868).  Our methods are remarkably more stable across scenarios as well -- even compared with 10-repetition Breiman scoring.

\subsubsection{Runtime}

For small datasets (low $n$ and $p$), our DFI calculation -- even the variant using optimal rank shifts -- is up to 9$\times$ faster than scikit-learn's permutation importance implementation with $B=1$. This suggests that scikit-learn's implementation includes overhead that only amortizes for larger datasets. Indeed, our experiments indicate that such overhead may become advantageous for $n \gtrsim 10,000$ and moderate $p \in [11,99]$, say $p \gtrsim 50$. Regardless, our method remains competitive with even the fastest possible evaluation of Breiman's method (the rather uncommon $B=1$) across all tested regimes (including $n=10,000$ with $p = 100$).
\medskip

Crucially, for the standard $B=10$ ensemble (the typical benchmark for stable scores), our method is typically at least about an order of magnitude faster across all scenarios.
\medskip

This reduced runtime, which stems from requiring just a single evaluation per feature instead of $B$,  is especially relevant when auditing models whose evaluation is particularly compute intense, like tabular foundation models (TFM), making our proposed framework a more viable option.

\section{From Feature Importance to Model Stress-Testing}
\label{sec:VItoMST}

Beyond efficiently producing deterministic, reliable FI scores, our framework naturally
extends to \emph{model stress-testing}. 
\medskip

In this context, stress-testing a model -- or more formally, a (model, data) pair -- refers to evaluating the impact that the degradation of each feature would have on model performance. Data degradation can happen in practice due to several reasons, and some of the most common include covariate shifts, sensor failure, human error, and adversarial attacks.
\medskip

Although superficially the problem at hand might appear identical to estimating variable importance (and it is indeed related), it is fundamentally different, because in order to determine the true risk that a compromised feature poses to a given fitted model we must account for network effects and contagion risk: the degradation of a feature rarely takes place in isolation. In highly interconnected systems (like financial markets or macro-economies), closely related features will likely be impacted to some extent too, creating contagion risk which may increase (or decrease) the \emph{systemic} importance of the original feature of interest.
\medskip

A simple analogy will suffice to illustrate the difference between direct and systemic FI, as the systemic economic importance of a component may exceed its direct contribution. Consider Spain: it accounts for 8.7\% of the European Union's GDP (Eurostat, 2023 figures) so its direct economic importance within the EU could reasonably be estimated as that number; however, it is not hard to imagine that its systemic importance might be substantially higher due to significant trade flows and financial integration with larger European economies like France, Italy or Germany, as well as its role as a key gateway between Europe, North Africa, and Latin America. Therefore, estimating the systemic economic importance of Spain within the EU requires taking into account the strength of its ties with the rest of member states, which is something that direct FI frameworks (e.g. Breiman-style permutation FI) cannot do by construction.
\medskip

Unlike classical sensitivity analyses that treat input changes in isolation (e.g. ``x decreases by 10\%"), systemic importance quantifies the model's vulnerability to correlated covariate corruption or covariate-shift scenarios in which a disturbance to one predictor co-occurs with predictable changes in its correlated neighbors.
\medskip

Observe that model class FI methods like LOCO or ghost variables, despite accounting for feature correlation,  cannot estimate systemic importance: in fact,  by discounting the importance of features that are more correlated with the rest of covariates, they do the opposite. We emphasize these methods are not incorrect, they are simply designed to estimate a different quantity (and thereby answer a different question).
\medskip

By allowing our permutation-based framework to account for feature correlations we seamlessly switch from direct to systemic FI estimation, where more strongly correlated features have their importance scores boosted, not deflated.
\medskip

We define systemic importance as the expected disruption to model predictions when a single feature is perturbed and that perturbation is allowed to \emph{propagate} to other covariates according to their empirical correlations. The resulting systemic importance vector is non-negative and normalized to sum to one, and admits a direct interpretation as the fraction of total predictive disruption attributable to each feature when correlated propagation is accounted for. We formalize this notion next.

\begin{definition}[Systemic Feature Importance (SFI)] \label{def:SFI}
With the same setup as in Definition \ref{def:DFI}, the estimated \emph{systemic importance score} $s_k$ of the $k$-th variable $x_k$ is the model prediction disruption that occurs when $x_k$ is permuted, and this perturbation is propagated to all other covariates.
\medskip

\textbf{1. Propagation Rule:} When the $j$-th feature, $\mathbf{x}_j$, is perturbed (permuted) to $\mathbf{x}_j'$, the value of any other feature $\mathbf{x}_k$ ($k \neq j$) in the original data matrix $X$ is simultaneously updated according to the following order-invariant rule:
\begin{equation}
\mathbf{x}_k' \leftarrow \mathbf{x}_k + \mathrm{cor}(\mathbf{x}_k, \mathbf{x}_j) \times \Delta x_j
\end{equation}
where $\mathrm{cor}(\mathbf{x}_k, \mathbf{x}_j)$ is the empirical correlation between features $k$ and $j$ (e.g. Spearman's $\rho$) and $\Delta x_j := \mathbf{x}_j'-\mathbf{x}_j$ is the vector of changes induced by the permutation of $x_j$. In order to disregard spurious correlations, the propagation only occurs if $|\mathrm{cor}(\mathbf{x}_k, \mathbf{x}_j)|$ exceeds a data-driven tolerance threshold.
\medskip

\textbf{2. Systemic Score:} The raw systemic importance score $s_k^{\mathrm{raw}}$ is the resulting expected change in prediction when only $x_k$ is perturbed and the full propagation rule is applied. The final normalized systemic importance score $s_k$ is calculated as:
\begin{equation}
s_k := \frac{s_k^{\mathrm{raw}}}{\sum_{j=1}^p s_j^{\mathrm{raw}}}
\end{equation}

\textbf{Interpretable Score Decomposition:} For analytical purposes, the normalized systemic importance score $s_k$ can be decomposed into its \emph{direct importance score} ($d_k$) and its \emph{indirect importance score} ($i_k$):
\begin{equation}
s_k = d_k + i_k
\end{equation}
The direct score $d_k$ is the normalized performance degradation from the standard, isolated permutation of $x_k$ (i.e., when no propagation occurs). The indirect score $i_k$ is the difference that captures the network amplification or dampening of $x_k$'s total importance (relative to the rest of covariates), reflecting its interconnectedness with the feature set.
\end{definition}

Put simply, the systemic importance of a feature is equal to its direct importance plus an indirect component which is dynamically computed by propagating the perturbations of all other covariates proportionally to their correlation with the feature of interest, so that correlated covariates amplify or dampen the original (direct) perturbation. 
\medskip

Indeed, to estimate systemic feature importance we perform two steps: first we perturb $x_j$ just like in DFI, but then, before re-evaluating the model, we propagate this perturbation to $x_j$ through the system, shifting every other feature $x_k$ proportionally by its empirical correlation ($\mathrm{cor}(\mathbf{x}_k, \mathbf{x}_j) \times \Delta x_j$). This makes direct feature importance a special case of systemic feature importance with $\mathrm{cor}(\mathbf{x}_k, \mathbf{x}_j) = 0$ for all $k \neq j$.
\medskip

Because our systemic importance measure relies on correlations, all covariates must either be numeric or have a numeric encoding in the primary model: ordinal encoding (the ideal case if there is a meaningful order), one-hot encoding or target-encoding. In that case, our suggested rank-based correlation default metric (Spearman) is still appropriate for the purposes of detecting non-parametric feature association. We emphasize that systemic importance reflects (and is bounded by) model behavior and empirical correlations, rather than causal structure.
%\medskip

Observe that with this simple framework we avoid simulating dynamic systems,  imposing any kind of parametric model or assuming a particular multivariate feature distribution. As such, the objective is not to necessarily replace much more complex tools specifically designed to model and analyze shocks in dynamic systems like Dynamic Stochastic General Equilibrium (DSGE) or Structural Equation Models (SEM). Instead, it is intended to be a fast and transparent post-hoc tool built on top of classical permutation scoring,  which can be applied to any predictive model (all it takes is knowledge of its evaluation function, no need to know any internals) to provide a more holistic view of the importance of each feature in a model taking into account their correlation as a first-order approximation to their dependence structure.

\subsection{Statistical Calibration of the Propagation Threshold}

To ensure that SFI captures genuine structural dependencies rather than spurious numerical artifacts, we implement a non-parametric calibration mechanism to establish a baseline correlation threshold $\tau$. In finite samples, uncorrelated features may exhibit small non-zero empirical correlations due to sampling variability. Establishing $\tau$ via statistical calibration defines a principled noise floor below which minor correlations are truncated to zero.

\subsubsection{Null Matrix Construction and Determinism}
We define the global null hypothesis of total independence, $H_0$, under which the joint feature distribution factors completely into the product of its marginals: $f(x_1, \dots, x_p) = \prod_{j=1}^p f(x_j)$. To simulate a synthetic dataset $\tilde{\mathbf{X}} \in \mathbb{R}^{n \times p}$ conforming to $H_0$, we independently shuffle the row indices of each column in the training matrix $\mathbf{X}$. This procedure preserves the exact empirical marginal distributions $f(x_j)$ while stochastically breaking all pairwise and higher-order inter-feature dependencies. 
\medskip

To maintain strict determinism and reproducibility across execution environments, column shuffles are generated using a fixed internal pseudo-random seed. Note that applying a uniform rank or index shift operator simultaneously across all columns would preserve original row alignments and leave the empirical correlation matrix unchanged. 

\subsubsection{Non-parametric Proxy Detection with False Discovery Rate (FDR) Control}
When auditing models for hidden reliance on protected attributes, the primary risk is failing to detect subtle proxy networks (a Type II error). Traditional Family-Wise Error Rate (FWER) control penalizes the $M = p(p-1)/2$ pairwise comparisons so severely that moderate but genuinely non-zero correlations between sensitive attributes and proxy variables are truncated to zero, giving a false impression of model fairness.  Since our network propagation mechanism relies on the accurate estimation of all feature correlations simultaneously, setting a high threshold to guarantee FWER in high-dimensional settings could severely compromise its utility.
\medskip

To maximize sensitivity to genuine proxy pathways while bounding false discoveries, we calibrate the propagation threshold $\tau$ by controlling the False Discovery Rate (FDR) at target significance level $\alpha \in (0, 1)$ (typically $\alpha = 0.05$). Under $H_0$, the $M$ absolute correlations computed on $\tilde{\mathbf{X}}$ are exchangeable,  forming an empirical null distribution $F_{\mathrm{null}}$. For each observed feature pair $(X_i, X_j)$ in the original matrix $\mathbf{X}$, we compute its empirical null p-value:
\begin{equation}
    p_{ij} = 1 - F_{\mathrm{null}}(|\hat{\rho}_{ij}|) = \frac{1}{M} \sum_{1 \le a < b \le p} \mathbb{I}\left( |\hat{\rho}(\tilde{X}_a, \tilde{X}_b)| \ge |\hat{\rho}_{ij}| \right)
\end{equation}

Let $p_{(1)} \le p_{(2)} \le \dots \le p_{(M)}$ denote the ordered p-values across all $M$ feature pairs. We apply the Benjamini--Hochberg step-up procedure to identify the largest cutoff index $k$:
\begin{equation}
    k = \max \left\{ m \in \{1, \dots, M\} : p_{(m)} \le \frac{m}{M} \alpha \right\}
\end{equation}
If no such index $k$ exists, no pairwise correlations exceed the noise floor and $\tau$ defaults to $1.0$ (disabling systemic propagation). Otherwise, the propagation threshold $\tau$ is set to the magnitude of the $k$-th ordered correlation coefficient:
\begin{equation}
    \tau = |\hat{\rho}|_{(k)}
\end{equation}

Correlations satisfying $|\hat{\rho}_{ij}| > \tau$ are retained as systemic proxy pathways in the propagation matrix $\mathbf{R}_\tau$, whereas those falling below $\tau$ are truncated to zero as sampling numerical artifacts. This non-parametric thresholding mechanism guarantees that among all inter-feature dependencies admitted into the SFI propagation mechanism, the expected proportion of false-positive proxy connections is bounded by $q$, maintaining high empirical power for subtle proxy networks.

\bigskip

Knowing \textit{which} features matter most for a given model -- whether directly or systemically -- is only part of the story; another one is knowing \textit{how} they matter. A first, important step is to understand if the overall association between a given feature and the predicted response is positive or negative, which is what we will focus on in the next section.

\section{Importance Direction Estimation}
\label{sec:direction}

While scalar importance metrics quantify the magnitude of model reliance on a given feature $X_j$, they are intrinsically sign-agnostic: a high importance score indicates that a feature strongly influences model predictions, but reveals nothing about whether an increase in $X_j$ drives predictions up or down overall. In high-stakes domains such as credit risk assessment, clinical diagnostics, and algorithmic fairness auditing, this directional ambiguity represents a critical information gap.
\medskip

To bridge this gap,  we introduce a directional operator $\operatorname{Dir}(X_j) \in \{-1, 0, +1\}$ that upgrades scalar importance scores into directional importance diagnostics without distorting the underlying magnitude calculations and with negligible computational cost, as it relies on intermediate calculations that must be carried out in all cases.

\subsection{Directional Magnitude-Weighted Concordance Operator}
For an observation $\mathbf{x}_i \in \mathbb{R}^p$ and a given permutation $\pi_j$, let $\Delta x_{i,j} := x_{\pi_j(i), j} - x_{i,j}$ denote the physical displacement of feature $j$ under permutation $\pi_j$, and let $\Delta \hat{f}_{i,j} := \hat{f}(\mathbf{x}_{i, \pi_j}) - \hat{f}(\mathbf{x}_i)$ represent the resulting prediction shift.  Crucially, the construction of the permuted vector $\mathbf{x}_{i, \pi_j}$ -- and, consequently, $\Delta x_{i,j}$ and the induced prediction shift $\Delta \hat{f}_{i,j}$ -- adapts to the chosen evaluation context:
\begin{itemize}
    \item DFI: Only coordinate $j$ is permuted while holding all other features fixed, yielding the direct prediction shift $\Delta \hat{f}_{i,j}^{\mathrm{dir}}$.
    \item SFI: The perturbation in coordinate $j$ is linearly propagated across correlated covariates according to empirical feature correlations, yielding the systemic prediction shift $\Delta \hat{f}_{i,j}^{\mathrm{sys}}$.
\end{itemize}
This distinction is vital for model auditing: while a protected attribute $X_j$ excluded from model training yields $\Delta \hat{f}_{i,j}^{\mathrm{dir}} = 0$,  propagating its perturbation under SFI generates non-zero systemic shifts $\Delta \hat{f}_{i,j}^{\mathrm{sys}}$ whenever $X_j$ correlates with active model covariates. This enables our directional framework to uncover both the magnitude and sign of hidden proxy reliance.
\medskip

To determine the overall orientation of feature influence under either evaluation mode, we measure the sample-wide agreement between the sign of the input perturbation and the signed value of the induced prediction shift:
\begin{equation}
\label{eq:dir_operator}
\operatorname{Dir}(X_j) := \operatorname{sgn} \left( \frac{1}{n} \sum_{i=1}^n \operatorname{sgn}\left(\Delta x_{i,j}\right) \cdot \Delta \hat{f}_{i,j} \right)
\end{equation}
where $\operatorname{sgn}(\cdot) \in \{-1, 0, +1\}$ is the standard sign function.

\medskip

The inner product $\operatorname{sgn}(\Delta x_{i,j}) \cdot \Delta \hat{f}_{i,j}$ isolates local concordance: when an increase in feature value ($\Delta x_{i,j} > 0$) causes the prediction to increase ($\Delta \hat{f}_{i,j} > 0$), or a decrease in feature value ($\Delta x_{i,j} < 0$) causes the prediction to decrease ($\Delta \hat{f}_{i,j} < 0$), the local product is positive. If $\Delta x_{i,j}$ and $\Delta \hat{f}_{i,j}$ have opposite signs, the local contribution is negative. Averaging across all $n$ instances and taking the outer sign yields a global orientation:
\begin{itemize}
\item $\operatorname{Dir}(X_j) = +1$: The model exhibits a net positive monotonic relationship with $X_j$.
\item $\operatorname{Dir}(X_j) = -1$: The model exhibits a net negative monotonic relationship with $X_j$.
\item $\operatorname{Dir}(X_j) = 0$: The model response is purely non-monotonic or symmetric across the empirical feature distribution.
\end{itemize}

\textbf{Remark (Magnitude-weighted concordance vs.  unweighted voting):}
\label{rem:magnitude_weighting}
By retaining the signed scalar prediction difference $\Delta \hat{f}_{i,j}$ inside the summation rather than reducing it to an unweighted binary vote $\operatorname{sgn}(\Delta \hat{f}_{i,j}) \in \{-1, 0, +1\}$, the estimator differentiates between minor localized fluctuations and major structural model responses for a given change in input value $\Delta x_{i,j}$.  Conversely, under an unweighted voting scheme, a feature inducing numerous small prediction shifts in one direction (e.g., $\Delta \hat{f}_{i,j} = +10^{-3}$) and a few large shifts in the opposite direction (e.g., $\Delta \hat{f}_{i,j} = -10^{3}$) would be considered as having an overall positive concordance with the model's prediction,  even though the model's average prediction change when feature $j$ is permuted is overwhelmingly negative. 
\medskip

Moreover, weighting by $\Delta \hat{f}_{i,j}$ offers a theoretical advantage, namely consistency with $\ell_1$ DFI scoring: under full positive or negative directional monotonicity (as in linear models), the un-signed sum inside $\operatorname{Dir}(X_j)$ evaluates exactly to the $\ell_1$ DFI score, i.e.,  $S_j := \frac{1}{n} \sum_{i=1}^n \operatorname{sgn}(\Delta x_{i,j}) \cdot \Delta \hat{f}_{i,j} = \pm I_j$, where $I_j = \frac{1}{n} \sum_{i=1}^n \vert{}\Delta \hat{f}_{i,j}\vert{}$.  In more general settings that don't satisfy full directional monotonicity, this formulation bounds the directional sum to the interval $[-I_j, +I_j]$, ensuring that directional orientation is directly tied to the prediction-shift norm generating $I_j$. Concretely, the ratio $\frac{S_j}{I_j} \in [-1, +1]$ can be interpreted as a directional purity index, where $+1$ means pure monotonic positive, $-1$ means pure monotonic negative, and $0$ means pure symmetry or overall effect cancellation. In contrast, using $\operatorname{sgn}(\Delta \hat{f}_{i,j})$ would erode this direct relationship with DFI.  Note that this theoretical advantage extends to SFI and/or $\ell_2$ scoring, as our next Remark on adaptation to quadratic loss metrics illustrates.
\medskip

In the case of highly complex non-linear prediction functions $\hat{f}$ (e.g.,  model ensembles or Tabular Foundation Models), local directional concordance $\operatorname{sgn}(\Delta x_{i,j}) \cdot \Delta \hat{f}_{i,j}$ may vary in sign across the feature domain $\mathcal{X}_j$. In such regimes, $\operatorname{Dir}(X_j)$ uses $n$ sampled instances according to the chosen permutation to evaluate the \emph{net empirical orientation} of the model response:
\begin{itemize}
    \item Dominant monotonic regimes: when $\hat{f}$ exhibits a predominant global trend with localized fluctuations, $\operatorname{Dir}(X_j) \in \{-1, +1\}$ aims to capture that overall association direction.
    \item Symmetric and non-monotonic responses: when $\hat{f}$ exhibits symmetric or opposing directional effects across equal masses of the feature support (e.g., quadratic responses $\hat{f}(x_j) \propto X_j^2$), positive and negative local concordances cancel under sample aggregation, yielding $\operatorname{Dir}(X_j) = 0$. 
\end{itemize}
Thus, a directional score of zero serves as an explicit diagnostic alerting the practitioner that a feature with large scalar importance magnitude operates through strong non-monotonic or interaction-driven mechanics. More generally, $S_j$ can be stored as an intermediate step to flag near-perfect cancellation (and more generally to discriminate between varying intensities of positive and negative association after normalizing by $I_j$ as shown e.g.  in Figure \ref{fig:Debt_directions}).  In those cases, a directional score of zero (or near zero, if the intermediate step discussed is saved) is useful to prompt localized inspection via other post-hoc explainability tools better suited for local -- rather than global --  effect estimation such as Shap or LIME.
\medskip

\textbf{Remark (Adaptation to quadratic loss metrics):}
\label{rem:dir_metrics}
For model evaluation under quadratic loss ($\ell_2^2$ / MSE or $\ell_2$ / RMSE),  the directional score $S_j := \frac{1}{n} \sum_{i=1}^n \operatorname{sgn}(\Delta x_{i,j}) \cdot \Delta \hat{f}_{i,j}$ can be adapted to match the choice of loss function:
\begin{equation}
S_j^{(\mathrm{MSE})} := \frac{1}{n} \sum_{i=1}^n \operatorname{sgn}\left(\Delta x_{i,j}\right) \cdot \operatorname{sgn}\left(\Delta \hat{f}_{i,j}\right) \cdot \left(\Delta \hat{f}_{i,j}\right)^2
\end{equation}
This formulation yields $S_j^{(\mathrm{MSE})} / I_j^{(\mathrm{MSE})} \in [-1, +1]$ for MSE, and $\operatorname{sgn}(S_j^{(\mathrm{MSE})}) \sqrt{\vert S_j^{(\mathrm{MSE})} \vert} / I_j^{(\mathrm{RMSE})} \in [-1, +1]$ for RMSE, preserving exact directional purity bounds across all common scoring metrics. Unless explicitly stated otherwise, we adopt the canonical $\ell_1$ formulation in what follows.
\medskip

\textbf{Remark (Decoupling magnitude and direction):}
\label{rem:factorization}
Multiplying the directional operator $\operatorname{Dir}(X_j) \in \{-1, 0, +1\}$ by the scalar DFI or SFI score $I_j$ yields a \emph{signed directional score}:
\begin{equation}
I_j^{\mathrm{signed}} := \operatorname{Dir}(X_j) \cdot I_j \in [-I_j, +I_j]
\end{equation}
This factorization cleanly decouples \emph{magnitude} (governed by prediction-shift norms) from \emph{orientation} (governed by directional concordance). Crucially, because the absolute magnitude $|I_j^{\mathrm{signed}}| = I_j$ is preserved, all theoretical guarantees established for the scalar estimators in Section \ref{sec:oneperm} -- including exact scale recovery under location-scale distributions and MSE dominance -- remain completely intact. From an analytical standpoint, this signed representation retains full interpretability: the sign immediately identifies the net direction of feature influence, while the absolute score magnitudes can still be normalized to sum to unity for context-agnostic relative ranking.

\subsection{Theoretical Properties}

Under appropriate assumptions, it is straightforward to show that $\operatorname{Dir}(X_j)$ is guaranteed to recover the true overall direction sign under any choice of non-trivial permutation.

\begin{proposition}[Sign Recovery for Linear Models]
\label{prop:sign_recovery_lm}
Let $\hat{f}(\mathbf{x}) = \hat \beta_0 + \sum_{j=1}^p \hat \beta_j x_j$ be a linear prediction function. For non-constant feature $X_j$ and any permutation $\pi_j$ with non-zero feature displacements (i.e., $x_{\pi_j(i), j} \neq x_{i,j}$ for at least one $i$),
\begin{equation}
\operatorname{Dir}(X_j) = \operatorname{sgn}(\hat \beta_j)
\end{equation}
\end{proposition}

\begin{proof}
For a linear model, the prediction difference simplifies to:
$$ \Delta \hat{f}_{i,j} = \hat{f}(\mathbf{x}_{i, \pi_j}) - \hat{f}(\mathbf{x}_i) = \hat \beta_j \left( x_{\pi_j(i), j} - x_{i,j} \right) = \hat \beta_j \Delta x_{i,j}$$
Substituting this into Equation~\ref{eq:dir_operator}:
$$\frac{1}{n} \sum_{i=1}^n \operatorname{sgn}(\Delta x_{i,j}) \cdot \hat \beta_j \Delta x_{i,j}  = \hat \beta_j \cdot \left( \frac{1}{n} \sum_{i=1}^n \vert{}\Delta x_{i,j}\vert{} \right)$$
Since at least one $x_{\pi_j(i), j} \neq x_{i,j}$, the average absolute displacement $C_j := \frac{1}{n} \sum_{i=1}^n \vert{}\Delta x_{i,j}\vert{} > 0$. Taking the outer sign yields:
$$\operatorname{Dir}(X_j) = \operatorname{sgn}\left(\hat \beta_j \cdot C_j\right) = \operatorname{sgn}(\hat \beta_j) \cdot \operatorname{sgn}(C_j) = \operatorname{sgn}(\hat \beta_j)$$
\end{proof}

A similar result can be stated for other underlying model architectures besides linear models, such as monotonic additive models.

\begin{proposition}[Sign Recovery for Monotonic Additive Models]
\label{prop:sign_recovery_mad}
Let $\mathbf{x} = (x_1, \dots, x_p)^T \in \mathbb{R}^p$ denote a single observation vector, and suppose the prediction function decomposes as $\hat{f}(\mathbf{x}) = g_j(x_j) + h(\mathbf{x}_{-j})$ for a non-constant feature $X_j$ with empirical support $\mathcal{X}_j \subset \mathbb{R}$.  Consider any permutation $\pi_j$ with non-zero feature displacements (i.e., $x_{\pi_j(i), j} \neq x_{i,j}$ for at least one $i$).  If $g_j: \mathcal{X}_j \to \mathbb{R}$ is strictly increasing (resp. strictly decreasing), then:
\begin{equation}
\operatorname{Dir}(X_j) = +1 \quad (\text{resp. } -1)
\end{equation}
\end{proposition}

\begin{proof}
Evaluating $\hat{f}$ at observation $\mathbf{x}_i$ and its permuted counterpart $\mathbf{x}_{i, \pi_j}$, the sub-vector $h(\mathbf{x}_{i, -j})$ remains unpermuted and cancels out in the prediction difference:
\[
\Delta \hat{f}_{i,j} = \hat{f}(\mathbf{x}_{i, \pi_j}) - \hat{f}(\mathbf{x}_i) = g_j(x_{\pi_j(i), j}) - g_j(x_{i,j})
\]
For any $a, b \in \mathbb{R}$,  because $g_j: \mathcal{X}_j \to \mathbb{R}$ is strictly increasing (resp.  strictly decreasing), $g_j(b) - g_j(a) = \operatorname{sgn}(b - a) \cdot \vert{}g_j(b) - g_j(a)\vert{}$ (resp. $g_j(b) - g_j(a) = -\operatorname{sgn}(b - a) \cdot \vert{}g_j(b) - g_j(a)\vert{}$). Thus, for every instance $i$:
\[
\operatorname{sgn}(\Delta x_{i,j}) \cdot \Delta \hat{f}_{i,j} = \operatorname{sgn}(\Delta x_{i,j})^2 \cdot |g_j(x_{\pi_j(i), j}) - g_j(x_{i,j})| = |g_j(x_{\pi_j(i), j}) - g_j(x_{i,j})| \ge 0
\]
(resp.  $-|g_j(x_{\pi_j(i), j}) - g_j(x_{i,j})| \le 0$), with strict inequality whenever $x_{\pi_j(i), j} \neq x_{i,j}$. 
\medskip

Since $X_j$ is non-constant under $\pi_j$, at least one term is strictly positive (resp.  negative). Summing across all $n$ instances yields a strictly positive (resp.  negative) aggregate quantity, establishing $\operatorname{Dir}(X_j) = +1$ (resp. $-1$).
\end{proof}

\subsection{Application to Algorithmic Fairness and Compliance Auditing}
In regulatory compliance and algorithmic fairness, scalar importance metrics are fundamentally limited because they cannot distinguish between beneficial and discriminatory model reliance.  For example, in automated credit scoring, a model may exhibit non-zero reliance on a sensitive attribute such as gender, race or age (typically via proxy variables, as direct inclusion is generally prohibited). A scalar metric capable of detecting reliance through proxy variables merely flags that the model relies on the attribute,  but it cannot determine whether the attribute artificially inflates or depresses credit scores for specific demographic groups, which could have different legal implications in each case.  By computing $\operatorname{Dir}(X_{\text{protected}})$, auditors can explicitly verify both the \emph{scale} and \emph{sign} of indirect discrimination, enabling:
\begin{enumerate}
\item Disparate impact verification: a positive systemic direction indicates that a higher value of the protected attribute, propagated through proxy variables included in the model, systematically increase predictions. Whether this benefits or penalizes higher values of the protected attribute (or class 1, if binary) depends on the consequences that a higher prediction entails. For example, in the case of a credit scoring model that estimates probability of default it would be penalizing, as higher predictions indicate a higher risk of default, leading to a reduced likelihood of granting a loan to that individual. The symmetric analysis is true in case of a negative systemic direction.

\item Monotonicity checks: models are often required by regulators to satisfy domain-specific monotonicity constraints (e.g., loan default probability should increase with debt-to-income ratio).  Equation~\ref{eq:dir_operator} provides an automated, single-pass diagnostic to verify compliance across the entire feature space.
\end{enumerate}

\section{Case Studies}
\label{sec:casestudies}

To evaluate the practical utility of our proposed framework beyond controlled benchmark settings, we examine two high-stakes empirical case studies where classical PFI fails to provide actionable diagnostic insight.  Beyond traditional model stress-testing under structural domain shifts, such as macroeconomic shocks or sensor failures, our approach addresses a critical imperative in model governance and financial regulation: evaluating algorithmic fairness against protected attributes. Across both case studies, we first verify that DFI achieves ground-truth recovery comparable to standard PFI while drastically reducing computational runtime. We then demonstrate how SFI, when coupled with Importance Direction, exposes hidden proxy discrimination and algorithmic bias in credit risk assessment, uncovering both the magnitude and sign of indirect reliance on sensitive features.  Ultimately, across both applications, our single-pass deterministic framework delivers exact, reproducible, and directional diagnostics under real-world data distributions.

\subsection{Case Study A (Regression): Debt-to-Income Ratio Estimation}

The Boston HMDA dataset is a well-known dataset with 2380 observations and 12 features collected in the 1990s that derived from the federal Home Mortgage Disclosure Act (HMDA) data, used extensively in econometrics to study mortgage lending patterns, especially racial bias, by analyzing applicant features (e.g. debt ratios, job, age, or race) against loan outcomes (approval/denial). This dataset, available at \url{https://github.com/adc-trust-ai/trust-free}, is often used to explore if lenders treat minorities differently, finding persistent disparities even after controlling for financial factors \citep{Munnell1996, Bostic1997}. 
\medskip

In our case, however, we will use this dataset to test,  on the one hand,  our proposed DFI method in terms of its ability to recover ground-truth scores, and on the other hand, the SFI method that we propose in combination with importance direction estimation, with a view to model fairness.
\medskip

We will estimate the debt-to-income ratio of mortgage applicants in the Boston HMDA dataset, based on 12 personal attributes, spanning 5 numeric and 7 binary.  The response variable \texttt{dir} has the following empirical order statistics: min = 0, Q1 = 0.28, median = 0.33, Q3 = 0.37, and max = 3. It is therefore heavily right-skewed, which typically makes estimation by standard regression methods more challenging -- not an issue for our purposes.
\medskip

Since we wish to be able to define ground-truth FI scores to serve as benchmark in our experiments, we choose unregularized and $\ell_1$-regularized linear models to estimate the debt-to-income ratio of mortgage applicants in the Boston HMDA dataset.  A black-box benchmark -- Random Forest (RF) \citep{Breiman2001} -- is also included to provide a performance reference and to show that our FI methods can be applied to black-box models as well -- and that is arguably their most useful application, in fact.  Although it is debatable whether the FI scores that RF natively provides in most implementations should be considered ``ground-truth" for that model or just an estimate based on a specific criteria choice (typically mean decrease in node impurity), we confirm that our proposed framework achieves a similar correlation with those scores compared to the classical baseline methods, around 0.98 in all cases.
\medskip

All primary models are run with default parameters, which in the case of RF involves an ensemble of 100 trees as implemented in standard Python libraries like scikit-learn.
\medskip

We randomize the order of the rows in the original dataset and perform 10-fold cross-validation to obtain repeated measures of primary model accuracy as well as direct FI method accuracy.

\subsubsection{Direct FI}
In this dataset,  as Table \ref{tab:Debt_master_model} shows,  the performance of Lasso is essentially identical to that of unregularized linear regression. Meanwhile, a black-box model like RF only manages to achieve similar accuracy but employed 100 trees with over 125,000 logic rules in total (although many of those trees were possibly redundant \citep{Dorador2025}).

\begin{table}[!h]
\centering
\caption{Forecasting out-of-fold average primary model results with two standard error bands  in the Boston HMDA dataset}
\label{tab:Debt_master_model}
\begin{tabular}{lccc}
\toprule
Metric & Lasso Regression & Linear Regression & RF \\ \midrule
$R^2$ & 0.49 ± 0.10 & 0.49 ± 0.10 & 0.46 ± 0.08\\
MAE & 0.048 ± 0.002 & 0.048 ± 0.002 & 0.048 ± 0.002 \\
Nonzero coefficients  & 11.8 ± 0.2 & 12.0 ± 0.0 & N/A \\ \bottomrule
\end{tabular}
\end{table}

Tables \ref{tab:Debt_VI_TRUST} and \ref{tab:Debt_VI_OLS} show that our proposed methods,  under MSE scoring,  have a very similar performance to the classical baselines but are more stable and at least an order of magnitude faster when compared to the $B=1$ and $B=10$ classical variants.

\begin{table}[h]
\centering
\caption{Results of different FI methods under $\ell_1$-regularized linear regression primary estimator -- Boston HMDA dataset}
\label{tab:Debt_VI_TRUST}
\begin{tabular}{lcccc}
\toprule
Metric & DFI-Opt & DFI-Approx & Breiman (B=1) & Breiman (B=10) \\ \midrule
Ground-truth cor.  & 0.997 ± 0.001 & 0.997 ± 0.001 & 0.996 ± 0.001 & 0.996 ± 0.001 \\
Max score diff.  & 0.272 ± 0.014 & 0.273 ± 0.015 & 0.271 ± 0.020 & 0.270 ± 0.019 \\
Mean score diff.  & 0.045 ± 0.001 & 0.045 ± 0.002 & 0.045 ± 0.003 & 0.045 ± 0.003 \\
Time (ms) & 0.79 ± 0.01 & 0.63 ± 0.01 & 3.64 ± 0.11 & 21.42 ± 0.20 \\ \bottomrule
\end{tabular}
\end{table}

\begin{table}[h]
\centering
\caption{Results of different FI methods under an unregularized primary linear model  -- Boston HMDA dataset}
\label{tab:Debt_VI_OLS}
\begin{tabular}{lcccc}
\toprule
Metric & DFI-Opt & DFI-Approx & Breiman (B=1) & Breiman (B=10) \\ \midrule
Ground-truth cor.  & 0.997 ± 0.001 & 0.997 ± 0.001 & 0.996 ± 0.001 & 0.996 ± 0.001 \\
Max score diff.  & 0.279 ± 0.019 & 0.279 ± 0.019 & 0.278 ± 0.023 & 0.277 ± 0.023 \\
Mean score diff.  & 0.046 ± 0.003 & 0.047 ± 0.003 & 0.046 ± 0.004 & 0.046 ± 0.004 \\
Time (ms) & 0.69 ± 0.01 & 0.61 ± 0.01 & 3.50 ± 0.06 & 21.31 ± 0.28 \\ \bottomrule
\end{tabular}
\end{table}

\medskip

Tables \ref{tab:Debt_VI_TRUST_mae} and \ref{tab:Debt_VI_OLS_mae} in Appendix \ref{ap:res_mae} show the results employing default scoring functions for our methods and the classical baseline (but in the case of the latter it is the same scoring function in the case of regression). Those results highlight that correlation with ground truth is rather stable across scoring functions, but that is not the case for a more extreme metric like the maximum score difference.
\medskip

Lastly, to further test the stability of the traditional FI methods on real datasets we execute them ten times and check the top 5 features that are reported each time.  We use a 60-40 train-test split to emulate typical use while ensuring we still keep a non-trivial number of test observations ($\sim$1,000) to rule out artifacts related to having an extremely small sample size to permute. Table \ref{tab:Debt_Flicker} in Appendix \ref{ap:flicker} shows that the classical baseline employing only one repetition yields very unstable rankings for the top 5 features: 7 distinct top-5 rankings are produced in 10 runs. Employing $B=10$ repetitions reduces this instability to some degree although it is not fully resolved: 2 distinct top-5 rankings are produced in 10 runs. In stark contrast, our deterministic framework is designed to yield the exact same scores (and, hence, rankings) every single time it is run. In addition, only our optimal-permutation method recovered the ground-truth top 5 components in their correct order. This experiment exposes the instability of classical PFI methods even under non-trivial sample sizes, compromising its responsible use in model audits.

\subsubsection{Systemic FI}

We will test two scenarios: one where we use all of our features to estimate the debt-to-income ratio of mortgage applicants via linear regression, and another scenario where we exclude the sensitive attribute \texttt{black} (1 if the applicant is African American, 0 otherwise). The purpose of this setup is to show that even if a model does not directly use a sensitive feature, the model might still rely on the sensitive attribute via proxy variables, which is something that traditional Breiman-style FI methods are not able to  detect.

\begin{itemize}
\item With a linear regression primary model using all features, the systemic relative importance of feature \texttt{black} in the fitted model is estimated to be 0.16\%, which can be decomposed as the sum of its direct (0.01\%) and indirect (0.15\%) relative importance scores. Thus, its systemic importance is around 16 times larger than its direct importance (which is what classical Breiman-style methods would only be able to estimate). Using MAE instead of MSE, we obtain the following importance decomposition: 1.97\% = 0.06\% + 1.91\%, which in this case implies an even larger multiplier of around 31.

\item With a linear regression primary model that excludes only the protected attribute \texttt{black}, the systemic relative importance of said feature in the fitted model is estimated to be 0.10\%, which is due entirely to its indirect component, as the direct importance of the feature in the model is exactly 0.  Note that the systemic reliance on the protected attribute when it is excluded from the model is still nearly as large as when it is included,  due to proxy effects, which suggests that excluding it from the model is not sufficient to guarantee non-discrimination. Using MAE instead of MSE, we obtain a larger systemic importance score of 1.55\%,  again entirely due to network effects, and similar in size as in the case the protected attribute was part of the primary model. While reliance on the sensitive attribute appears modest in this case, being able to estimate it remains essential. If deemed necessary, corrective measures could include replacing features that are strongly correlated with race by others that exhibit lower correlation while having comparable or higher predictive power conditional on the presence of the rest of covariates.
\end{itemize}

\begin{figure}[!h]
\centering
\includegraphics[width=0.8\linewidth]{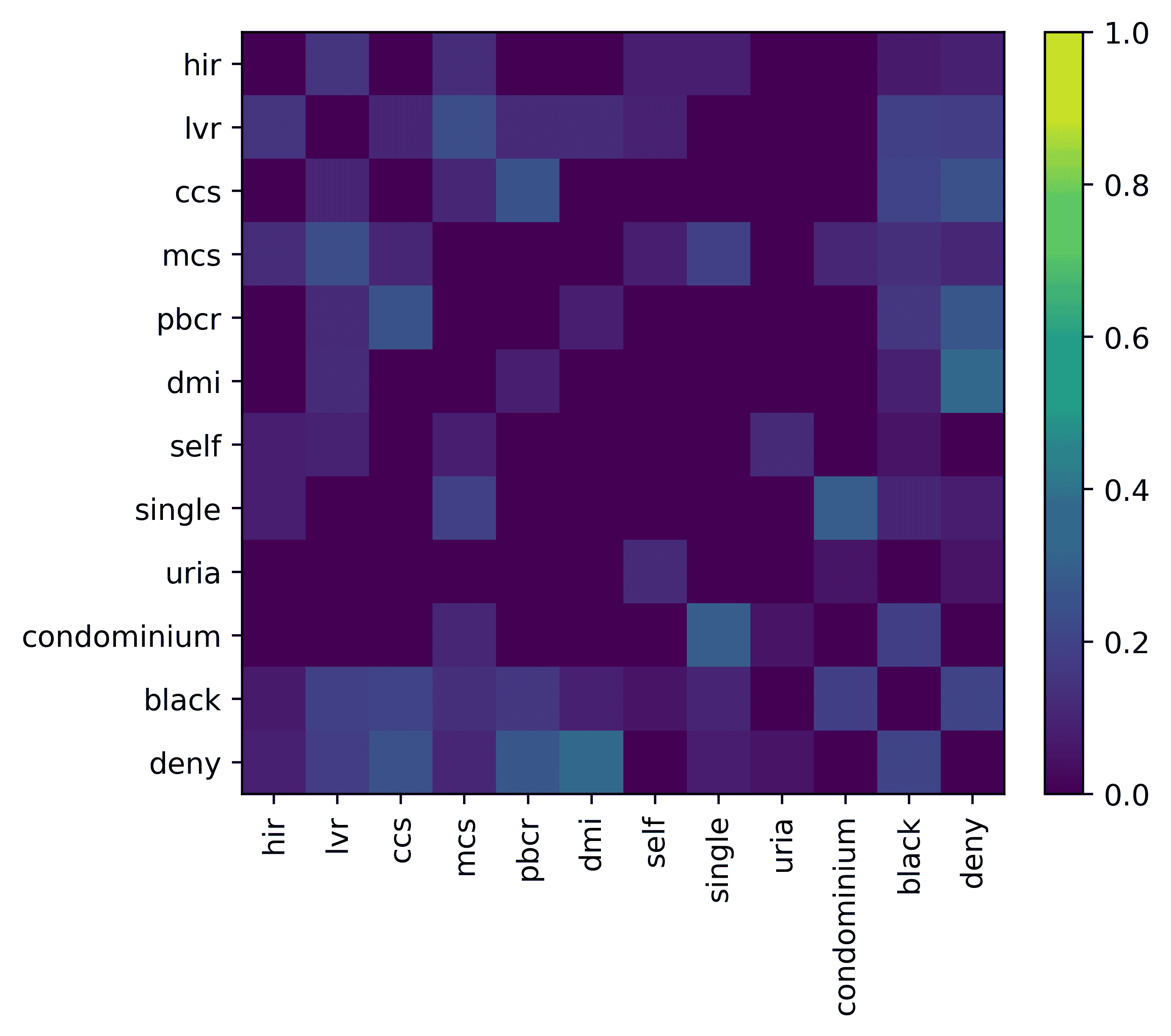}
\caption{Spearman correlation in the Boston HMDA test dataset (empirical tolerance threshold: 0.09, FDR significance level: 0.05)}
\label{fig:Debt_heatmap}
\end{figure}

Hence, even if the sensitive attribute \texttt{black} is not used directly by the model,  our systemic importance method identified that the model still relies on it indirectly through correlated covariates beyond any spurious correlation doubts (FDR significance level: 0.05),  such as \texttt{lvr} (ratio of size of loan to assessed value of property),  \texttt{css} (consumer credit score from 1 to 6, where a lower value represents a better score) or \texttt{condominium} (whether the applicant lives in a condominium as opposed to a detached single-family home).  Depending on the choice of metric (MSE or MAE), our method quantified this systemic reliance on the protected attribute to be between 0.10\% and 1.55\% of the total predictive reliance of the model, despite not being present in the model as such. This reliance on a protected attribute via proxy variables (Figure \ref{fig:Debt_heatmap}) would be an undetected model risk and potential regulatory non-compliance when assessed using traditional FI methods.

\medskip

Observe that the protected attribute \texttt{black} shows Spearman correlations of about 0.15 or higher with 4 of the 11 other features included in the dataset: \texttt{css}, \texttt{lvr} ,  \texttt{condominium}, and \texttt{pbcr} (public bad credit record). This means that the model can ``reconstruct" a sizable proportion of the racial information using those proxies, even when the race variable is explicitly excluded from the model. Taking into account all features having Spearman correlation with \texttt{black} exceeding the Benjamini-Hochberg-adjusted 95th quantile from a non-parametric null distribution to define its significant proxies,  we note that 7 out of 11 covariates used by the model have a statistically significant Spearman correlation with the protected attribute, and,  collectively, those proxies accumulate 28.6\% of the true FI in a model that excluded the sensitive attribute.
\medskip

\begin{figure}[!h]
\centering
\includegraphics[width=\linewidth]{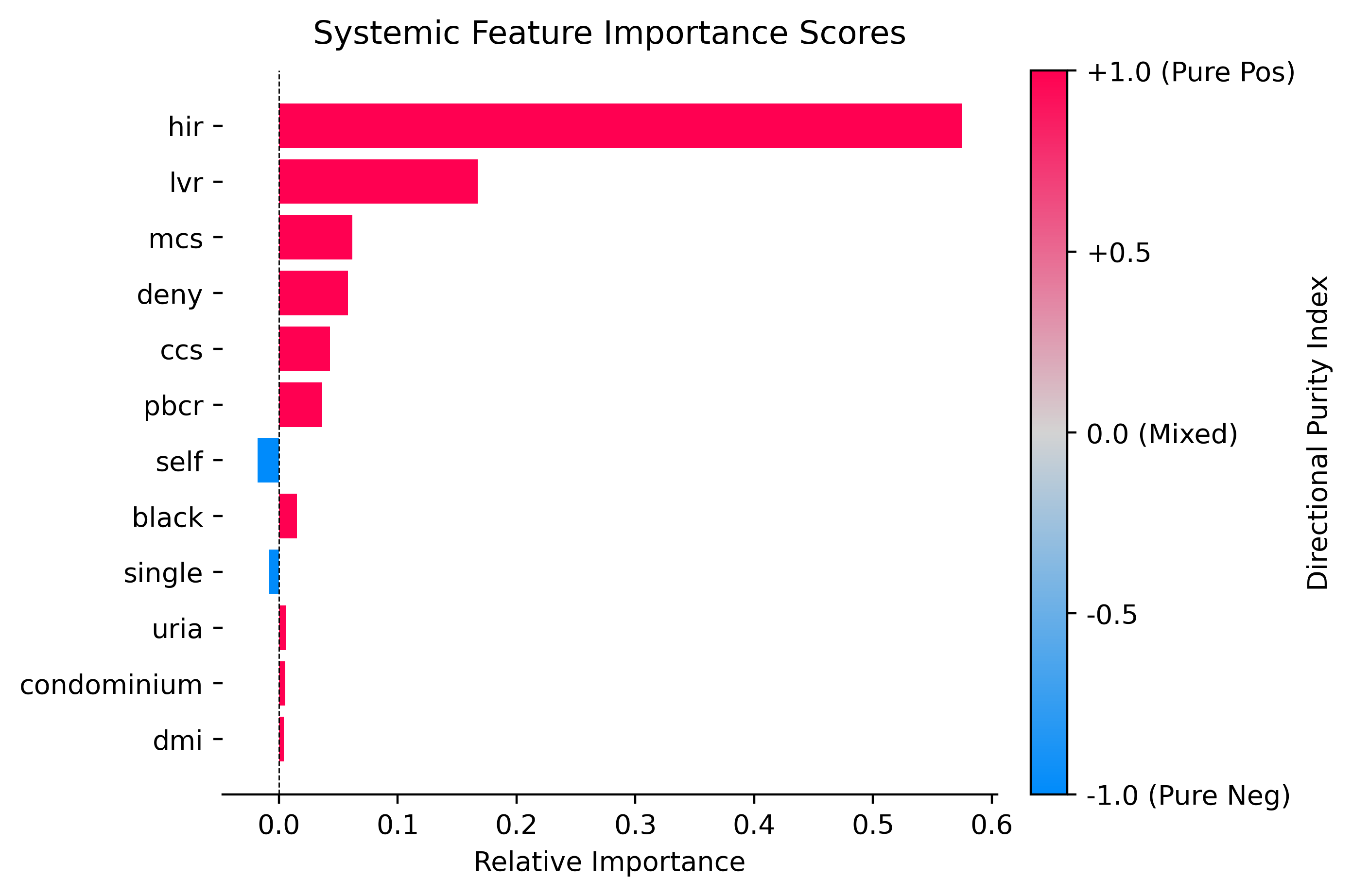}
\caption{Oriented Systemic Feature Importance scores of Lasso regression with \texttt{black} excluded from the model (Boston HMDA dataset)}
\label{fig:Debt_directions}
\end{figure}

Having established the systemic model reliance on the protected attribute of interest, we now use the framework proposed in Section \ref{sec:direction} to estimate the overall direction of systemic importance for the relevant protected attribute.  As Figure \ref{fig:Debt_directions} illustrates,  the overall association direction between the model's reliance on \texttt{black} and its predicted debt-to-income ratio is \textbf{positive}. This means that even though the protected attribute is excluded from the model it is still indirectly relied upon to influence predictions upward,  making access to credit \textit{harder} for this demographic group everything else being equal, as a higher debt-to-income ratio tends to be associated with a higher probability of loan rejection in standard credit models.

\subsection{Case Study B (Classification): Credit Scoring}
The German Credit dataset (Statlog Project) is a classic machine learning benchmark with 1000 records and 20 covariates, used to classify loan applicants as high or low credit risks, featuring mixed numerical (7) / categorical (13) features like credit history, duration, amount, housing, job, and demographics. This dataset is available on the UCI Machine Learning Repository, as well as at \url{https://github.com/adc-trust-ai/trust-free} with more convenient decoded feature names.
\medskip

Like we did in the regression case study of the previous section, we will use this dataset to test,  on the one hand,  our proposed direct FI method in terms of the usual accuracy metrics, and on the other hand, the systemic FI method that we propose coupled with importance direction estimation,  with a view to model fairness.
\medskip

As it is standard with this benchmark dataset, we will train a classifier to learn to classify loan applicants as either low (\texttt{class} = 0) or high (\texttt{class} = 1) risk based on their personal attributes. In our records, only 30\% are classified as high risk, so there is a noticeable class imbalance.  The misclassification cost matrix that accompanies this dataset establishes that a false negative (FN) is 5 times as costly as a false positive (FP), which may suggest using a custom metric such as $A = (5FN + FP)^{-1}$ that reflects that exact misclassification cost asymmetry. However, as it is standard practice, we will report the ROC-AUC as main metric of predictive performance. 
\medskip

Again, being able to define ground-truth FI scores requires using a model class in which they are clearly identifiable, such as linear models (in log-odds). Hence we consider $\ell_1$-regularized logistic regression (with a strong penalty so that most coefficients are given a zero coefficient) and unregularized logistic regression (so that all feature coefficients are kept nonzero). We also consider RF as a black-box benchmark to illustrate one of the main use cases of permutation-based FI methods, namely, its applicability to arbitrary black-box models for tabular data.  As in the previous regression case study, we confirm that our proposed methods achieve a similar correlation with the RF-native FI scores compared to the classical baseline methods (0.50-52 in all cases). The rest of the experimental setup is identical to that in the previous case study.

\subsubsection{Direct FI}
In this case,  as Table \ref{tab:Credit_master_model} shows,  the performance of the strongly-regularized logistic regression model is surprisingly competitive given that it only used 10\% of the available features, on average. As in the previous case study, the RF model only managed to achieve a performance comparable to the (unregularized) linear baseline but employed 100 trees with nearly 12,000 logic rules in total.

\begin{table}[!h]
\centering
\caption{Forecasting out-of-fold average primary model results with two standard error bands  in the German credit dataset}
\label{tab:Credit_master_model}
\begin{tabular}{lccc}
\toprule
Metric & $\ell_1$-Logistic & Logistic & RF \\ \midrule
ROC-AUC & 0.73 ± 0.03 & 0.79 ± 0.03 & 0.79 ± 0.03\\
Nonzero coef.  & 2.2 ± 0.4 & 20 ± 0.0 & N/A \\ \bottomrule
\end{tabular}
\end{table}

The first thing we observe when comparing Tables \ref{tab:Credit_VI_LogL1} and \ref{tab:Credt_VI_Log} is that feature importance is easier to recover in sparse models, everything else being equal. This is not surprising as having fewer features to score and rank should simplify the task exponentially. Tables \ref{tab:Credit_VI_LogL1} and \ref{tab:Credt_VI_Log} also show that our proposed methods (using MSE / negative Brier scoring as in all experiments) have comparable or better performance with respect to classical baselines while being more stable and up to 57 times faster in relation to the $B=10$ classical algorithm.
\medskip

Tables \ref{tab:Credit_VI_LogL1_mae} and \ref{tab:Credt_VI_Log_mae} in Appendix \ref{ap:res_mae} show the results employing default scoring functions for our methods and the classical baseline.  In that case, results remain favorable in the regularized case and even more favorable in the unregularized case compared to using MSE scoring.

\begin{table}[h]
\centering
\caption{Results of different FI methods under an $\ell_1$-regularized logistic regression primary estimator -- German Credit dataset}
\label{tab:Credit_VI_LogL1}
\begin{tabular}{lcccc}
\toprule
Metric & DFI-Opt & DFI-Approx & Breiman (B=1) & Breiman (B=10) \\ \midrule
Ground-truth cor.  & 0.996 ± 0.002 & 0.997 ± 0.002 & 0.999 ± 0.001 & 0.998 ± 0.001 \\
Max score diff.  & 0.072 ± 0.025 & 0.069 ± 0.024 & 0.035 ± 0.020 & 0.044 ± 0.021 \\
Mean score diff.  & 0.007 ± 0.003 & 0.007 ± 0.002 & 0.003 ± 0.002 & 0.004 ± 0.002 \\
Time (ms) & 1.26 ± 0.03 & 1.14 ± 0.06 & 8.79 ± 0.09 & 65.05 ± 0.44 \\ \bottomrule
\end{tabular}
\end{table}

\begin{table}[h]
\centering
\caption{Results of different FI methods under an unregularized logistic regression primary model -- German Credit dataset}
\label{tab:Credt_VI_Log}
\begin{tabular}{lcccc}
\toprule
Metric & DFI-Opt & DFI-Approx & Breiman (B=1) & Breiman (B=10) \\ \midrule
Ground-truth cor.  & 0.875 ± 0.015 & 0.900 ± 0.013 & 0.806 ± 0.036 & 0.773 ± 0.077 \\
Max score diff.  & 0.255 ± 0.031 & 0.200 ± 0.025 & 0.196 ± 0.053 & 0.196 ± 0.031 \\
Mean score diff.  & 0.029 ± 0.002 & 0.025 ± 0.002 & 0.035 ± 0.004 & 0.036 ± 0.003 \\
Time (ms) & 1.19 ± 0.03 & 1.10 ± 0.03 & 8.73 ± 0.17 & 64.19 ± 0.70 \\ \bottomrule
\end{tabular}
\end{table}

Table \ref{tab:Credit_Flicker} in Appendix \ref{ap:flicker} shows that across 10 independent runs on the German Credit dataset, both our optimal and approximate FI methods recovered the exact ground-truth top-5 features in the correct order in all runs. On the other hand,  both classical baselines employing one or even ten repetitions yielded very unstable rankings for the top 5 features across 10 independent runs,  producing 10 and 6 distinct top-5 rankings, respectively.

\subsubsection{Systemic FI}
As in the previous case study in the context of regression,  we will test two scenarios: one where we use all of our features, and another scenario where we guarantee the model does not use the sensitive attribute, in this case \texttt{Sex-Marital\_status}, which encodes information about sex and marital status of a borrower (e.g., `Female: single'). The purpose of this setup is to show that even if a model does not directly use a sensitive feature, the model might still rely on the sensitive attribute via proxy variables, which is something that traditional Breiman-style FI methods are not able to detect, as they ignore feature correlations.

\begin{itemize}
\item With a logistic regression primary model using all features,  the systemic relative importance of feature \texttt{Sex-Marital\_status} in the fitted model is estimated to be 3.46\%, which can be decomposed as the sum of its direct (4.95\%) and indirect (-1.49\%) relative importance scores. Note that in this case the network effects reflected in the `indirect' contribution are negative, diminishing the normalized systemic importance of this attribute relative to its normalized direct importance. Using MAE instead of MSE, we obtain the following importance decomposition: 5.29\% = 6.57\% - 1.28\%.

\item With a logistic regression primary model that only excludes the protected feature, the systemic relative importance of said feature in the fitted model is estimated to be 0.09\% via MSE, and 0.87\% via MAE. Hence, in this case the reliance on the sensitive feature appears modest and of similar proportions as in the previous use case.
\end{itemize}

\begin{figure}[!h]
\centering
\includegraphics[width=0.8\linewidth]{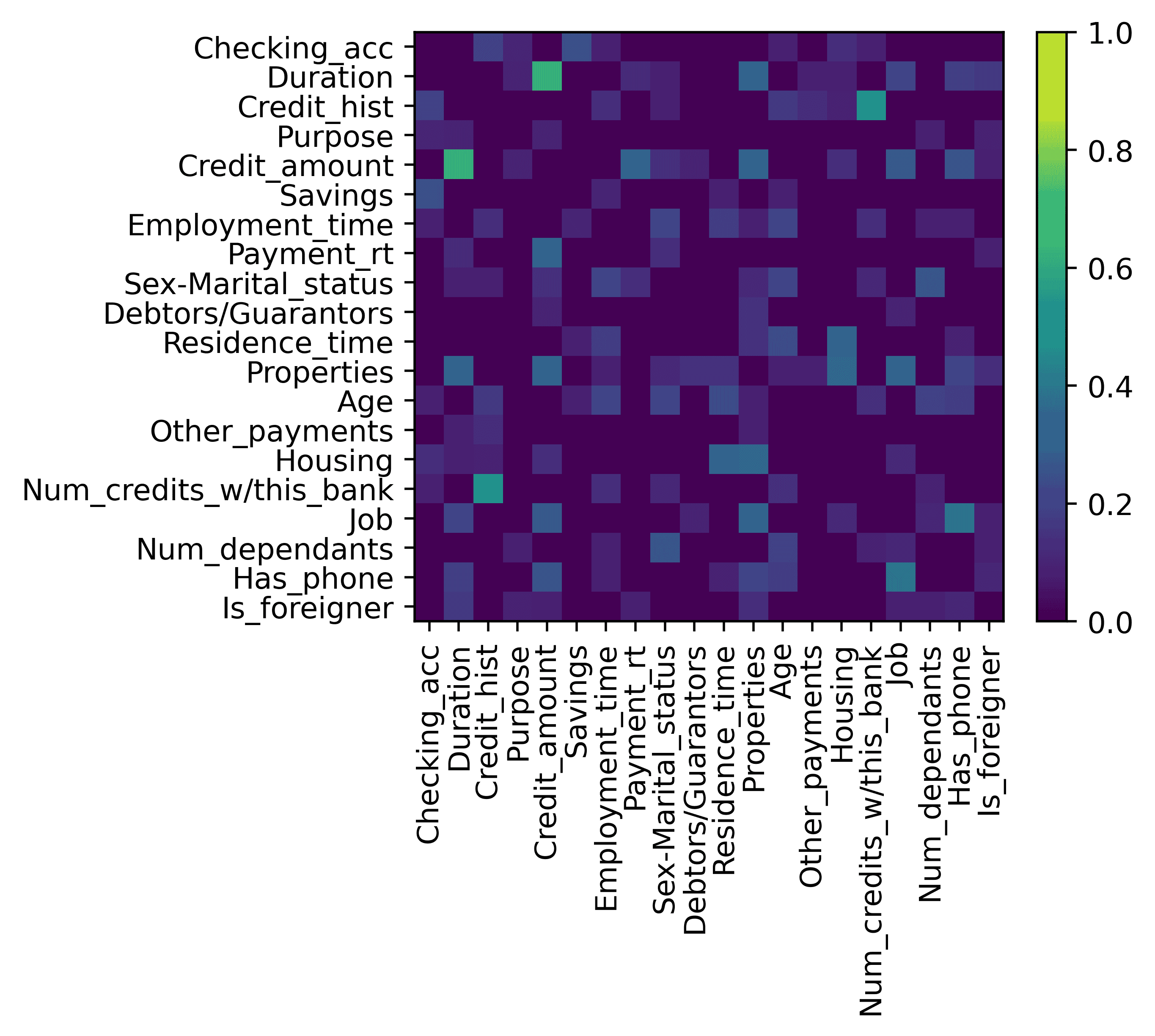}
\caption{Spearman correlation in the German Credit test dataset (empirical tolerance threshold: 0.13, FDR significance level: 0.05)}
\label{fig:Credit_heatmap}
\end{figure}

Hence, in this case study, our proposed systemic importance extension to our single-permutation FI method allowed us to audit the sparse fitted model and uncover a modest yet potentially relevant reliance on the protected attribute despite its explicit inclusion.
\medskip

\begin{figure}[!h]
\centering
\includegraphics[width=\linewidth]{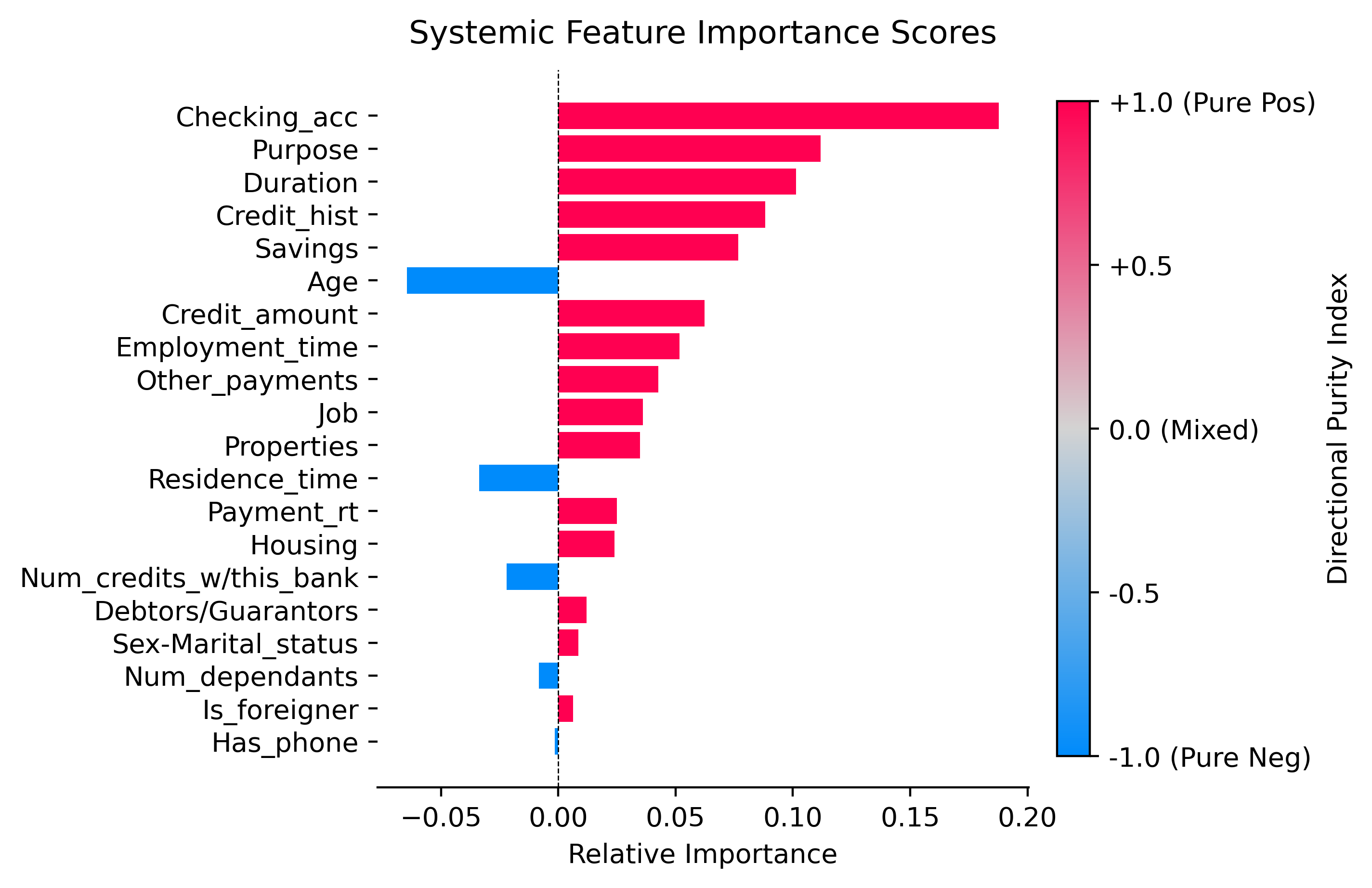}
\caption{Oriented Systemic Feature Importance scores of Lasso regression with \texttt{Sex-Marital\_status} excluded from the model (German Credit dataset)}
\label{fig:Credit_directions}
\end{figure}

Figure \ref{fig:Credit_directions} shows that the systemic reliance on sex and marital status has a  \textbf{positive} sign. In a credit scoring context this means that the model is penalizing the group with most `positive' target-encoded value, which is group A92 (female applicants). Indeed, the model, despite excluding explicit sex and marital status data still relies on such information to discriminate against women, as being in this demographic group is associated with a higher predicted probability of default according to the audited model (although the magnitude of this effect is small, less than 1\%).

\section{Conclusion}
\label{sec:conclusion}

As machine learning models become increasingly central to large-scale,  high-stakes decision-making in finance and macroeconomic policy, the tools used to interpret and stress-test them must meet rigorous standards of stability, efficiency, and reliability. In this paper, we challenged the long-standing convention that permutation-based feature importance requires repeated random sampling to be effective.
\medskip

We demonstrated that a single, deterministic, and max-min optimal rank-shift permutation is sufficient to recover meaningful feature importance scores, offering decisive theoretical and practical advantages over standard stochastic baselines. By eliminating sampling variance, Direct Feature Importance (DFI) yields strictly deterministic and reproducible scores -- an essential prerequisite for model validation, auditing, and regulatory compliance. Crucially, DFI achieves speedups of up to two orders of magnitude relative to classical frameworks, reducing model evaluation complexity from $\mathcal{O}(B \cdot n \cdot p)$ to $\mathcal{O}(n \cdot p)$. This computational efficiency makes DFI naturally scalable to high-dimensional models, inference-heavy architectures such as tabular foundation models, and real-time stress-testing pipelines. Across nearly 200 simulation scenarios, DFI consistently matches or exceeds Monte Carlo permutation methods in ground-truth recovery, delivering particularly strong gains in challenging low signal-to-noise and high-dimensional regimes.
\medskip

Beyond contributing a faster, more reliable alternative to classical permutation-based methods, we introduced Systemic Feature Importance (SFI) to address the limitations of evaluating features in isolation under non-trivial covariate dependence. By integrating a statistically calibrated correlation network with non-parametric False Discovery Rate control, SFI provides a principled, compute-efficient framework for quantifying a model's vulnerability to propagated feature perturbations. Case studies on the Boston HMDA and German Credit datasets demonstrate SFI's practical utility in algorithmic fairness auditing, exposing how models can silently inherit reliance on protected attributes through interconnected proxy networks -- a risk that standard permutation importance cannot detect.
\medskip

Finally, we extended this framework by introducing Importance Direction, which decouples importance magnitude from directional orientation at virtually zero additional computational cost. By measuring concordance between covariate displacements and output shifts, our approach transitions feature importance from unsigned scalar magnitudes to signed, directional representations. When combined with SFI, directional estimation allows practitioners and regulators to audit not only the magnitude of indirect model reliance on protected attributes via proxy variables, but also its net orientation.
\medskip

In summary, this work delivers a unified, model-agnostic framework for simultaneous feature attribution and systemic model risk assessment. By transitioning variable importance from randomized estimation to deterministic, network-aware stress-testing, we offer a diagnostic toolkit that is faster, fully reproducible, and better aligned with the interconnected nature of real-world data. Future work will explore extending the SFI framework beyond pairwise linear associations to capture non-monotonic and high-order dependence structures, further expanding its scope for regulatory compliance and automated model governance.

\bigskip

\bibliographystyle{plainnat}
\bibliography{references}

\clearpage
\appendix

\section{Proof of Theoretical Results}
\label{app:proofs}

\subsection{Proposition \ref{prop:optim} (Max-min optimality of $\pi_{\lfloor n / 2 \rfloor}$)}
\label{app:pf_optim}

\begin{proof}
Observe that for any $a,b \in \{0, \ldots n-1\}$ we have
\begin{equation}
d(a, b) \leq \lfloor n / 2 \rfloor
\end{equation}

because the maximum possible value of $\min(t, n-t)$ for integer $t \in \{0, \ldots n-1\}$ is $\lfloor n / 2 \rfloor$. Consequently, for any permutation $\pi$,
\begin{equation}
m(\pi) = \min_j d(j, \pi(j)) \leq \lfloor n / 2 \rfloor
\end{equation}

Hence, $\lfloor n / 2 \rfloor$ is an upper bound on the objective. We will show that choosing $\pi_k$ with $k = \lfloor n / 2 \rfloor$ attains said upper bound.

Let $k = \lfloor n / 2 \rfloor$ and consider the permutation (a cyclic shift) $\pi_k(j) = (j+k) \bmod n$. Note that for each $j$,
\begin{equation}
\vert \pi_k(j) - j \vert = k \text{ \,\,  or \, \, } \vert \pi_k(j) - j \vert = n - k
\end{equation}

Thus,  the circular displacement is
\begin{equation}
d(j, \pi_k(j)) = \min(k, n-k)
\end{equation}

but by definition of $k = \lfloor n / 2 \rfloor$ we have $\min(k, n-k) = k$. Therefore,
\begin{equation}
d(j, \pi_k(j)) = k \text{ \, for every } j
\end{equation}

and hence 
\begin{equation}
m(\pi_k) = \min_j d(j, \pi(j)) = k = \lfloor n / 2 \rfloor
\end{equation}
\end{proof}

%\subsection{Proof of Proposition \ref{prop:MSEdom} (Component-wise MSE dominance)}
%\label{app:pf_MSEdom}

\subsection{Proof of Proposition \ref{prop:jensens_sensitivity} (Sensitivity Dominance under Concave Response and $\ell_1$ Scoring)}
\label{app:pf_jensens}
\begin{proof}
Let $r = (1, 2, \dots, n)^\top$ be the vector of feature ranks, and let $d(\pi) = (|r_1 - r_{\pi(1)}|, \dots, |r_n - r_{\pi(n)}|)^\top \in \mathbb{R}_{\ge 0}^n$ be the rank displacement vector induced by a permutation $\pi$. 

By definition of the rank-space sensitivity function $g(d_i) = \mathbb{E}_{\Delta \mid d_i}[h(|\Delta_i|)]$, the expected empirical $\ell_1$ feature importance score under a permutation $\pi$ can be written in terms of rank displacements as:
\begin{equation}
I_{\ell_1}(\pi) = \frac{1}{n} \sum_{i=1}^n g(d_i(\pi))
\end{equation}
where $g$ is non-decreasing (because expectation is a monotonic operator), $g(0)=0$, and strictly concave on $[0, n-1]$.

Under an equiprobable random permutation $\Pi \sim \text{Unif}(S_n)$, the target rank $\pi(i)$ for any single instance at rank $i$ is marginally uniformly distributed over $\{1, \dots, n\}$ with probability $1/n$. The expected rank displacement for instance $i$ is:
\begin{equation}
\mathbb{E}_{\Pi}[d_i(\Pi)] = \frac{1}{n} \sum_{j=1}^n \vert{}i - j\vert{}
\end{equation}

Splitting the sum at $j = i$:
\begin{equation}
\sum_{j=1}^n \vert{}i - j\vert{} = \sum_{j=1}^{i-1} (i - j) + \sum_{j=i+1}^n (j - i)
\end{equation}
Note the first term equals $\frac{(i-1)i}{2} = \frac{i^2 - i}{2}$, while the second term equals $\frac{(n-i)(n-i+1)}{2} = \frac{(n-i)^2 + (n-i)}{2}$. Adding them together and dividing by $n$ gives:
\begin{equation}
\mathbb{E}_{\Pi}[d_i(\Pi)] = \frac{i^2 - i + (n - i)^2 + (n - i)}{2n}
\end{equation}

Now, to find the expected average displacement across the entire dataset under a random permutation, we average $\mathbb{E}_{\Pi}[d_i(\Pi)]$ over all $i = 1, \dots, n$:
\begin{equation}
\label{eq:dataset_avg}
\mu_{\Pi} = \frac{1}{n} \sum_{i=1}^n \mathbb{E}_{\Pi}[d_i(\Pi)] = \frac{1}{2n^2} \sum_{i=1}^n \left[ (i^2 - i) + (n-i)^2 + (n-i) \right]
\end{equation}

Note that, by reindexing $k = i-1$ on the left and $k = n-i$ on the right, we can show that $\sum_{i=1}^n (i^2 - i) = \sum_{i=1}^n ((n-i)^2 + (n-i))$. Thus, the sum in \eqref{eq:dataset_avg} simplifies to:
\begin{equation}
\mu_{\Pi} = \frac{1}{n^2} \sum_{i=1}^n (i^2 - i) = \frac{1}{n^2} \left( \sum_{i=1}^n i^2 - \sum_{i=1}^n i \right)
\end{equation}

Using the standard identities $\sum_{i=1}^n i^2 = \frac{n(n+1)(2n+1)}{6}$ and $\sum_{i=1}^n i = \frac{n(n+1)}{2}$:
\begin{equation}
\sum_{i=1}^n i^2 - \sum_{i=1}^n i = \frac{n(n+1)(2n+1)}{6} - \frac{3n(n+1)}{6} = \frac{n(n+1)(2n - 2)}{6} = \frac{n(n^2 - 1)}{3}
\end{equation}

Dividing by $n^2$:
\begin{equation}
\mu_{\Pi} = \frac{1}{n^2} \cdot \frac{n(n^2 - 1)}{3} = \frac{n^2 - 1}{3n}
\end{equation}

Taking expectation over $S_n$ and applying Jensen's inequality twice -- first over the random ensemble $\Pi$, and second over the non-constant displacements $d_i(\Pi)$ across instances --yields a strict upper bound due to the strict concavity of $g$:
\begin{equation}
\mathbb{E}_{\Pi}[I_{\ell_1}(\Pi)] = \frac{1}{n} \sum_{i=1}^n \mathbb{E}_{\Pi}[g(d_i(\Pi))] < g\left( \frac{1}{n} \sum_{i=1}^n \mathbb{E}_{\Pi}[d_i(\Pi)] \right) = g\left( \frac{n^2 - 1}{3n} \right)
\end{equation}

Now consider the deterministic cyclic shift $\Pi^* = \pi_{\lfloor n/2 \rfloor}$. By construction, $\Pi^*$ pairs each rank $r_i$ with its cyclic antipode, assigning a constant rank displacement $d_i(\Pi^*) = \lfloor n/2 \rfloor$ uniformly to all $n$ instances. Thus:
\begin{equation}
I_{\ell_1}(\Pi^*) = \frac{1}{n} \sum_{i=1}^n g(d_i(\Pi^*)) = g\left( \lfloor n/2 \rfloor \right)
\end{equation}

For all $n \ge 2$, $\lfloor n/2 \rfloor > \frac{n^2 - 1}{3n}$, and because $g$ is non-decreasing, evaluating $g$ at these points preserves the direction of the inequality:
\begin{equation}
\mathbb{E}_{\Pi}[I_{\ell_1}(\Pi)] < g\left( \frac{n^2 - 1}{3n} \right) \le g\left( \lfloor n/2 \rfloor \right) = I_{\ell_1}(\Pi^*)
\end{equation}
which establishes $\mathbb{E}_{\Pi}[I_{\ell_1}(\Pi)] < I_{\ell_1}(\Pi^*)$ for all $n \ge 2$.
\end{proof}

\subsection{Proof of Corollary \ref{prop:linear_recovery} (Recovery of Normalized, Scale-Adjusted Linear Regression Coefficients)}
\label{app:pf_linear_recovery}
\begin{proof}
For a linear model $f(\mathbf{x}) = \sum_{j=1}^p \beta_j x_j$, the absolute change in prediction for observation $i$ when feature $j$ is permuted via $\pi_j$ is given by:
\[
\Delta f_{i,j} = |f(\mathbf{x}_i) - f(\mathbf{x}_{i, \pi_j})| = |\beta_j \cdot (x_{i,j} - x_{\pi_j(i), j})| = |\beta_j| \cdot |x_{i,j} - x_{\pi_j(i), j}|.
\]

Taking the empirical mean yields the raw unnormalized $\ell_1$ importance score for feature $j$:
\[
I_{\ell_1, j} = \frac{1}{n} \sum_{i=1}^n |\beta_j| \cdot |x_{i,j} - x_{\pi_j(i), j}| = |\beta_j| \left( \frac{1}{n} \sum_{i=1}^n |x_{i,j} - x_{\pi_j(i), j}| \right) = |\beta_j| c_j.
\]

Applying sum-to-one normalization across all $p$ features yields the relative score:
\[
\tilde{I}_j = \frac{I_{\ell_1, j}}{\sum_{m=1}^p I_{\ell_1, m}} = \frac{|\beta_j| c_j}{\sum_{m=1}^p |\beta_m| c_m}.
\]

Because the marginal feature distributions belong to a common location-scale family, each feature $X_j$ can be represented as $X_j = \mu_j + \sigma_j Z_j$, where $Z_j \sim F_0$ is a standardized base distribution shared across all features. Under any uniform rank-shift operator $\pi_k$ with offset $k \in \{1, \dots, n-1\}$, the order statistics satisfy $x_{(i), j} - x_{\pi_k((i)), j} = \sigma_j (z_{(i)} - z_{\pi_k((i))})$. Consequently, the mean physical displacement evaluates to:
\[
c_j = \sigma_j \left( \frac{1}{n} \sum_{i=1}^n |z_{(i)} - z_{\pi_k((i))}| \right) = \sigma_j \cdot \kappa_k,
\]
where $\kappa_k > 0$ is a scale constant determined strictly by the rank offset $k$ and the quantile structure of $F_0$, and is invariant across all $j \in \{1, \dots, p\}$. Furthermore, since $\operatorname{MAD}(X_j) = \sigma_j \cdot \operatorname{MAD}(Z)$, we can write $c_j = \kappa_k' \cdot \operatorname{MAD}(X_j)$, where $\kappa_k' = \kappa_k / \operatorname{MAD}(Z) > 0$.

\medskip

Substituting $c_j = \kappa_k' \cdot \operatorname{MAD}(X_j)$ into the normalized importance expression yields:
\[
\tilde{I}_j = \frac{|\beta_j| \cdot \kappa_k' \cdot \operatorname{MAD}(X_j)}{\sum_{m=1}^p |\beta_m| \cdot \kappa_k' \cdot \operatorname{MAD}(X_m)} = \frac{|\beta_j| \cdot \operatorname{MAD}(X_j)}{\sum_{m=1}^p |\beta_m| \cdot \operatorname{MAD}(X_m)} = \frac{|\beta_j^*|}{\sum_{m=1}^p |\beta_m^*|}.
\]
The shared constant $\kappa_k'$ cancels out completely, demonstrating that exact recovery of scale-adjusted coefficients holds for any uniform rank-shift offset $k$.

\end{proof}

\clearpage

\section{Results with Default Scoring Metrics}
\label{ap:res_mae}

\begin{table}[!h]
\centering
\caption{Linear regression regime.  Global results across 48 scenarios with varying sample size, dimensionality, noise level, feature correlation and choice of ground-truth regression model}
\label{tab:linreg_mae}
\begin{tabular}{lcccc}
\toprule
Metric & DFI-Opt & DFI-Approx & Breiman (B=1) & Breiman (B=10) \\ \midrule
Ground-truth cor & 0.996 ± 0.002 & 0.992 ± 0.004 & 0.900 ± 0.037 & 0.910 ± 0.035 \\
Max score diff & 0.007 ± 0.002 & 0.011 ± 0.003 & 0.095 ± 0.024 & 0.091 ± 0.024 \\
Time (ms) & 25.0 ± 13.7 & 19.3 ± 10.7 & 17.9 ± 5.4 & 104.8 ± 33.1 \\ \bottomrule
\end{tabular}
\end{table}

\begin{table}[!h]
\centering
\caption{Nonlinear regression regime.  Global results across 48 scenarios with varying sample size, dimensionality, noise level, feature correlation and choice of ground-truth regression model}
\label{tab:nonlinreg_mae}
\begin{tabular}{lcccc}
\toprule
Metric & DFI-Opt & DFI-Approx & Breiman (B=1) & Breiman (B=10) \\ \midrule
Ground-truth cor & 0.997 ± 0.001 & 0.993 ± 0.003 & 0.899 ± 0.036 & 0.910 ± 0.034 \\
Max score diff & 0.013 ± 0.004 & 0.015 ± 0.005 & 0.173 ± 0.021 & 0.168 ± 0.023 \\
Time (ms) & 24.1 ± 13.2 & 18.7 ± 10.3 & 17.3 ± 5.2 & 101.9 ± 32.0 \\ \bottomrule
\end{tabular}
\end{table}

\begin{table}[!h]
\centering
\caption{Classification (linear response).  Global results across 48 scenarios with varying sample size, dimensionality, noise level, feature correlation and choice of ground-truth classifier}
\label{tab:linclass_mae}
\begin{tabular}{lcccc}
\toprule
Metric & DFI-Opt & DFI-Approx & Breiman (B=1) & Breiman (B=10) \\ \midrule
Ground-truth cor & 0.991 ± 0.004 & 0.982 ± 0.008 & 0.838 ± 0.070 & 0.867 ± 0.071 \\
Max score diff & 0.017 ± 0.006 & 0.019 ± 0.005 & 0.079 ± 0.021 & 0.076 ± 0.026 \\
Time (ms) & 26.7 ± 14.4 & 20.8 ± 11.3 & 20.1 ± 6.2 & 126.6 ± 41.6 \\ \bottomrule
\end{tabular}
\end{table}

\begin{table}[!h]
\centering
\caption{Classification (nonlinear response).  Global results across 48 scenarios with varying sample size, dimensionality, noise level, feature correlation and choice of ground-truth classifier}
\label{tab:nonlinclass_mae}
\begin{tabular}{lcccc}
\toprule
Metric & DFI-Opt & DFI-Approx & Breiman (B=1) & Breiman (B=10) \\ \midrule
Ground-truth cor & 0.995 ± 0.002 & 0.987 ± 0.009 & 0.798 ± 0.070 & 0.837 ± 0.064 \\
Max score diff & 0.018 ± 0.004 & 0.020 ± 0.005 & 0.165 ± 0.033 & 0.156 ± 0.029 \\
Time (ms) & 26.4 ± 14.3 & 20.8 ± 11.3 & 20.0 ± 6.2 & 125.6 ± 41.6 \\ \bottomrule
\end{tabular}
\end{table}

\begin{table}[!h]
\centering
\caption{Results of different FI methods under an $\ell_1$-regularized linear regression primary estimator with default scoring metrics  -- Boston HMDA dataset}
\label{tab:Debt_VI_TRUST_mae}
\begin{tabular}{lcccc}
\toprule
Metric & DFI-Opt & DFI-Approx & Breiman (B=1) & Breiman (B=10) \\ \midrule
Ground-truth cor & 0.952 ± 0.005 & 0.956 ± 0.004 & 0.943 ± 0.011 & 0.944 ± 0.011 \\
Max score diff & 0.225 ± 0.016 & 0.219 ± 0.015 & 0.472 ± 0.047 & 0.471 ± 0.045 \\
Mean score diff & 0.045 ± 0.003 & 0.044 ± 0.002 & 0.079 ± 0.008 & 0.079 ± 0.007 \\
Time (ms) & 1.60 ± 0.03 & 1.45 ± 0.03 & 5.06 ± 0.11 & 33.84 ± 0.91 \\ \bottomrule
\end{tabular}
\end{table}

\begin{table}[!h]
\centering
\caption{Results of different FI methods under an unregularized primary linear model with default scoring metrics  -- Boston HMDA dataset}
\label{tab:Debt_VI_OLS_mae}
\begin{tabular}{lcccc}
\toprule
Metric & DFI-Opt & DFI-Approx & Breiman (B=1) & Breiman (B=10) \\ \midrule
Ground-truth cor & 0.997 ± 0.000 & 0.998 ± 0.000 & 0.999 ± 0.000 & 0.999 ± 0.000 \\
Max score diff & 0.185 ± 0.012 & 0.190 ± 0.015 & 0.074 ± 0.026 & 0.073 ± 0.025 \\
Mean score diff & 0.031 ± 0.002 & 0.032 ± 0.002 & 0.013 ± 0.004 & 0.013 ± 0.004 \\
Time (ms) & 0.49 ± 0.03 & 0.39 ± 0.01 & 5.80 ± 0.15 & 42.03 ± 0.99 \\ \bottomrule
\end{tabular}
\end{table}

\begin{table}[!h]
\centering
\caption{Results of different FI methods under an $\ell_1$-regularized logistic regression primary estimator with default scoring metrics -- German Credit dataset}
\label{tab:Credit_VI_LogL1_mae}
\begin{tabular}{lcccc}
\toprule
Metric & DFI-Opt & DFI-Approx & Breiman (B=1) & Breiman (B=10) \\ \midrule
Ground-truth cor & 1.000 ± 0.000 & 1.000 ± 0.000 & 0.996 ± 0.002 & 0.996 ± 0.002 \\
Max score diff & 0.022 ± 0.010 & 0.013 ± 0.008 & 0.077 ± 0.028 & 0.077 ± 0.028 \\
Mean score diff & 0.002 ± 0.001 & 0.001 ± 0.001 & 0.008 ± 0.003 & 0.008 ± 0.003 \\
Time (ms) & 0.82 ± 0.04 & 0.67 ± 0.01 & 5.07 ± 0.10 & 32.56 ± 0.40 \\ \bottomrule
\end{tabular}
\end{table}

\begin{table}[!h]
\centering
\caption{Results of different FI methods under an unregularized logistic regression primary model -- German Credit dataset}
\label{tab:Credt_VI_Log_mae}
\begin{tabular}{lcccc}
\toprule
Metric & DFI-Opt & DFI-Approx & Breiman (B=1) & Breiman (B=10) \\ \midrule
Ground-truth cor & 0.930 ± 0.014 & 0.935 ± 0.017 & 0.682 ± 0.082 & 0.658 ± 0.148 \\
Max score diff & 0.060 ± 0.009 & 0.045 ± 0.005 & 0.180 ± 0.073 & 0.191 ± 0.096 \\
Mean score diff & 0.012 ± 0.001 & 0.010 ± 0.001 & 0.042 ± 0.011 & 0.042 ± 0.013 \\
Time (ms) & 0.75 ± 0.01 & 0.66 ± 0.01 & 4.95 ± 0.08 & 32.48 ± 0.22 \\ \bottomrule
\end{tabular}
\end{table}

\clearpage

\section{Stability Across Repeated Independent Runs}
\label{ap:flicker}

\begin{table}[h]
\centering
\caption{Top 5 features across 10 independent runs. Ground truth: $[ 0,  11,  6,  3,  1]$ -- Boston HMDA.}
\label{tab:Debt_Flicker}
\begin{tabular}{lcccc}
\toprule
Top 5 indices & DFI-Opt & DFI-Approx & Breiman (B=1) & Breiman (B=10) \\ \midrule
$[ 0, 11, 6, 3, 1]$ & 10  & 0 & 0 & 0 \\
$[ 0, 11, 6, 3, 4]$ & 0  & 10 & 0 & 0 \\
$[ 0, 11, 1, 3, 6]$ & 0  & 0 & 0 & 6 \\
$[ 0, 11, 1, 3, 2]$ & 0  & 0 & 3 & 4 \\
$[ 0, 11, 3, 1, 6]$ & 0  & 0 & 2 & 0 \\
$[ 0, 11, 6, 1, 3]$ & 0  & 0 & 1 & 0 \\
$[ 0, 11, 1, 4, 3]$ & 0  & 0 & 1 & 0 \\
$[ 0, 11, 1, 3, 5]$ & 0  & 0 & 1 & 0 \\
$[ 0, 11, 4, 3, 6]$ & 0  & 0 & 1 & 0 \\
$[ 0, 11, 3, 1, 4]$ & 0  & 0 & 1 & 0 \\ \bottomrule
\end{tabular}
\end{table}

\begin{table}[h]
\centering
\caption{Top 5 features across 10 independent runs. Ground truth: $[ 0, 3, 2, 1, 5]$ -- German Credit.}
\label{tab:Credit_Flicker}
\begin{tabular}{lcccc}
\toprule
Top 5 indices & DFI-Opt & DFI-Approx & Breiman (B=1) & Breiman (B=10) \\ \midrule
$[ 0, 3, 2, 1, 5]$ & 10  & 10 & 0 & 0 \\
$[ 0, 2, 1, 3, 5]$ & 0 & 0 & 0 & 4 \\
$[ 0, 1, 2, 3, 5]$ & 0 & 0 & 0 & 2 \\
$[ 0, 1, 3, 2, 5]$ & 0 & 0 & 0 & 1 \\
$[ 0, 1, 2, 7, 5]$ & 0 & 0 & 0 & 1 \\
$[ 0, 2, 1, 5, 3]$ & 0 & 0 & 0 & 1 \\
$[ 0, 1, 2, 3, 7]$ & 0 & 0 & 0 & 1 \\
$[ 0, 2, 1, 7, 5]$ & 0 & 0 & 1 & 0 \\
$[ 0, 1, 2, 8, 5]$ & 0 & 0 & 1 & 0 \\
$[ 0, 3, 1, 5, 7]$ & 0 & 0 & 1 & 0 \\
$[ 0, 3, 2, 1, 7]$ & 0 & 0 & 1 & 0 \\
$[ 0, 1, 5, 3, 2]$ & 0 & 0 & 1 & 0 \\
$[ 0, 2, 5, 1, 7]$ & 0 & 0 & 1 & 0 \\
$[ 0, 1, 2, 7, 3]$ & 0 & 0 & 1 & 0 \\
$[ 0, 2, 1, 5, 7]$ & 0 & 0 & 1 & 0 \\ 
$[ 0, 2, 5, 19, 18]$ & 0 & 0 & 1 & 0 \\
$[ 0, 5, 1, 2, 18]$ & 0 & 0 & 1 & 0 \\ \bottomrule
\end{tabular}
\end{table}

\clearpage

\section{Additional Figures and Plots}
\label{ap:plots}

\begin{figure}[!h]
\centering
\includegraphics[width=1\linewidth]{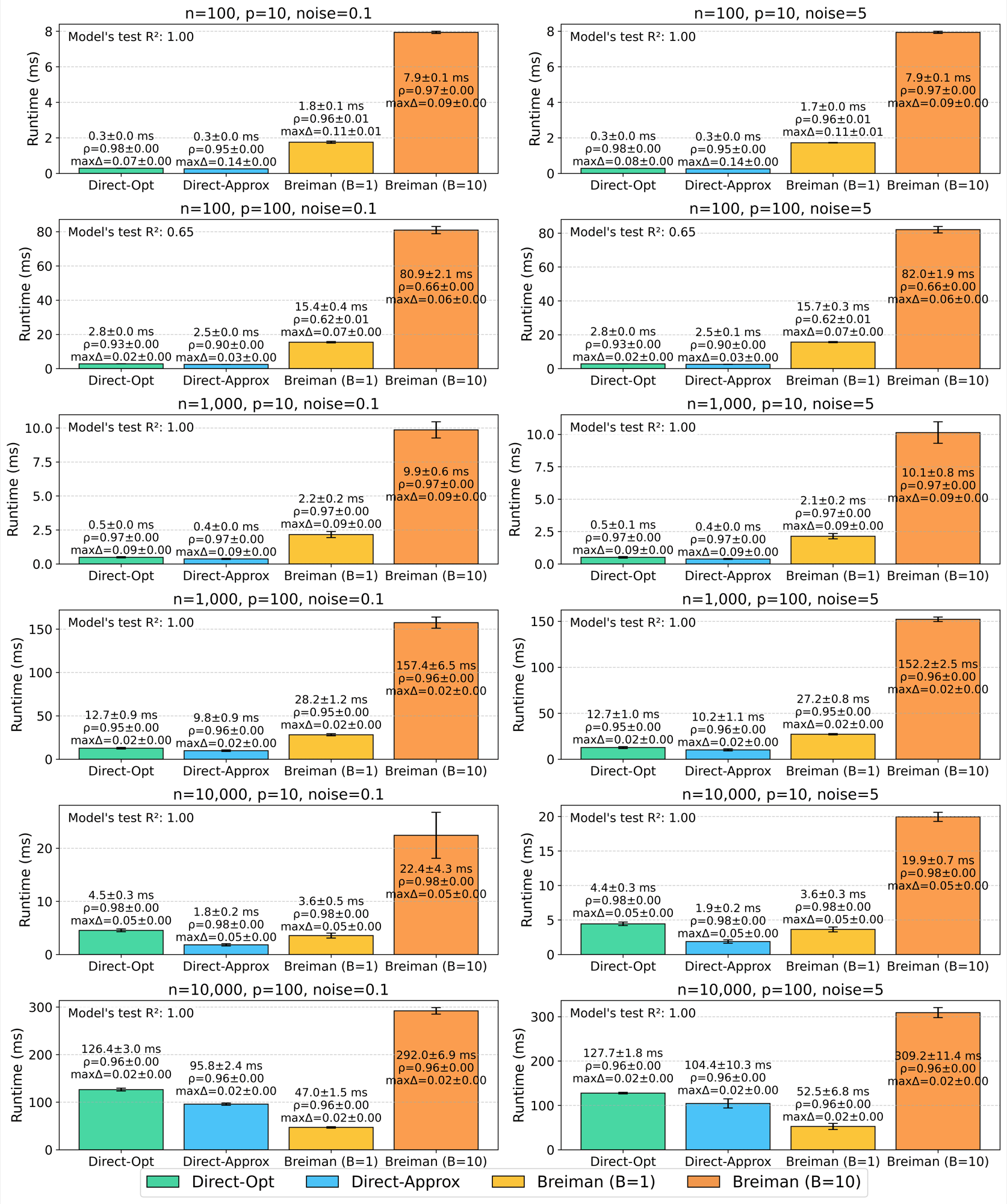}
\caption{Performance of different variable importance methods in linear regression with uncorrelated features and linear response under varying sample size, dimensionality and noise}
\label{fig:VIS_lm_uncorrelated}
\end{figure}

\begin{figure}[!h]
\centering
\includegraphics[width=1\linewidth]{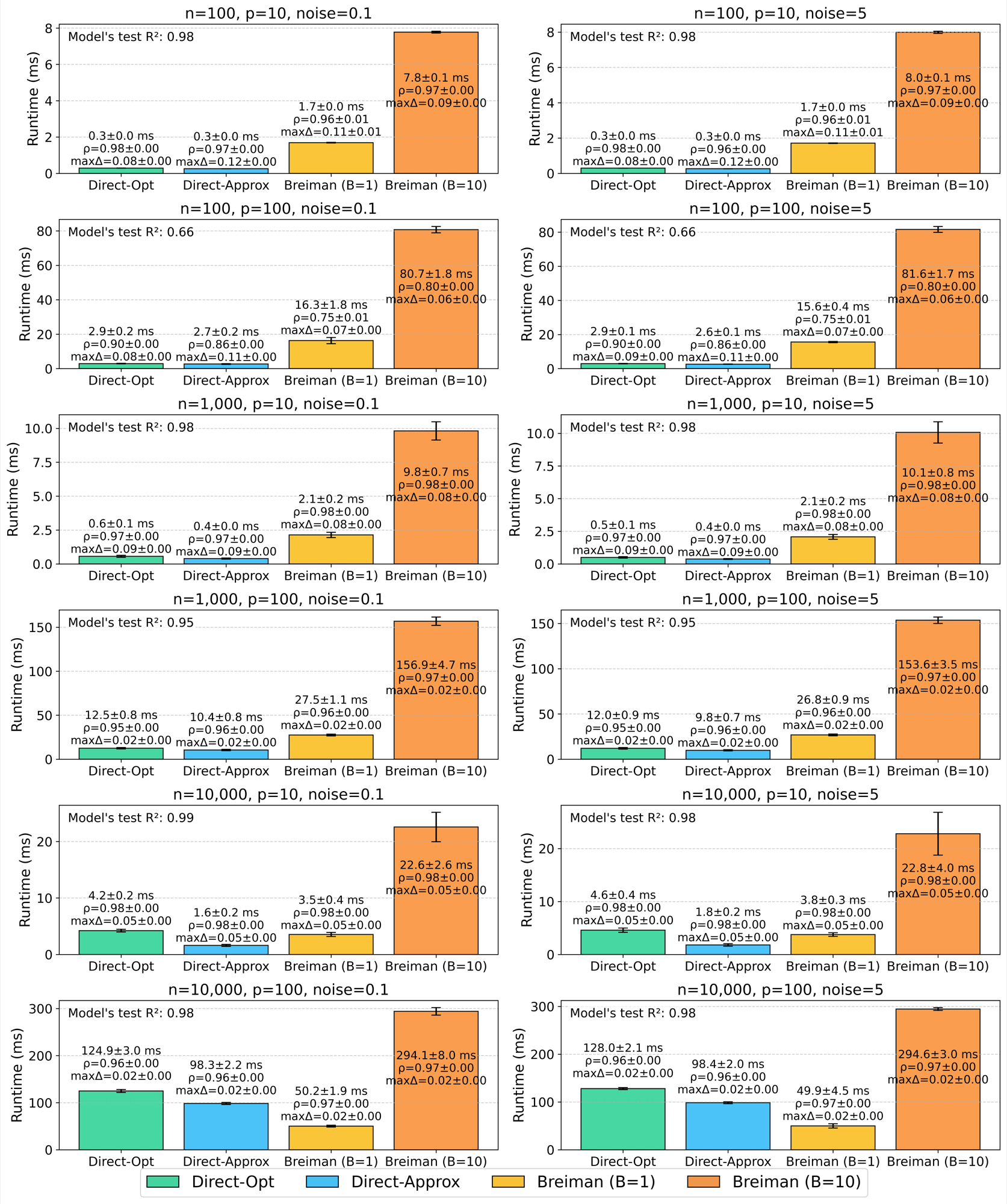}
\caption{Performance of different variable importance methods in $\ell_1$-linear regression with uncorrelated features and linear response under varying sample size, dimensionality and noise}
\label{fig:VIS_Lasso_uncorrelated}
\end{figure}

\begin{figure}[!h]
\centering
\includegraphics[width=1\linewidth]{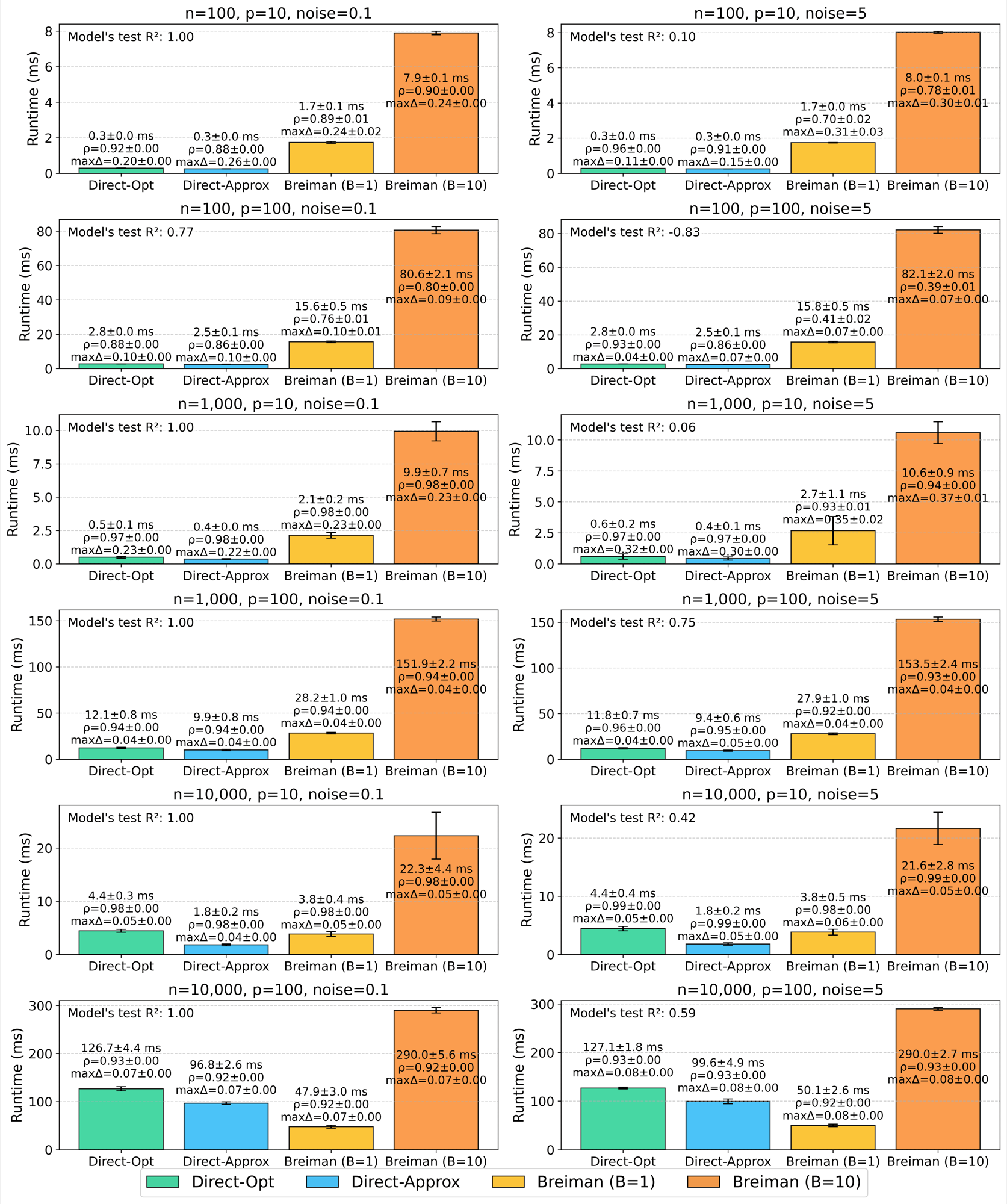}
\caption{Performance of different variable importance methods in linear regression with correlated features and linear response under varying sample size, dimensionality and noise}
\label{fig:VIS_lm_corr03}
\end{figure}

\begin{figure}[!h]
\centering
\includegraphics[width=1\linewidth]{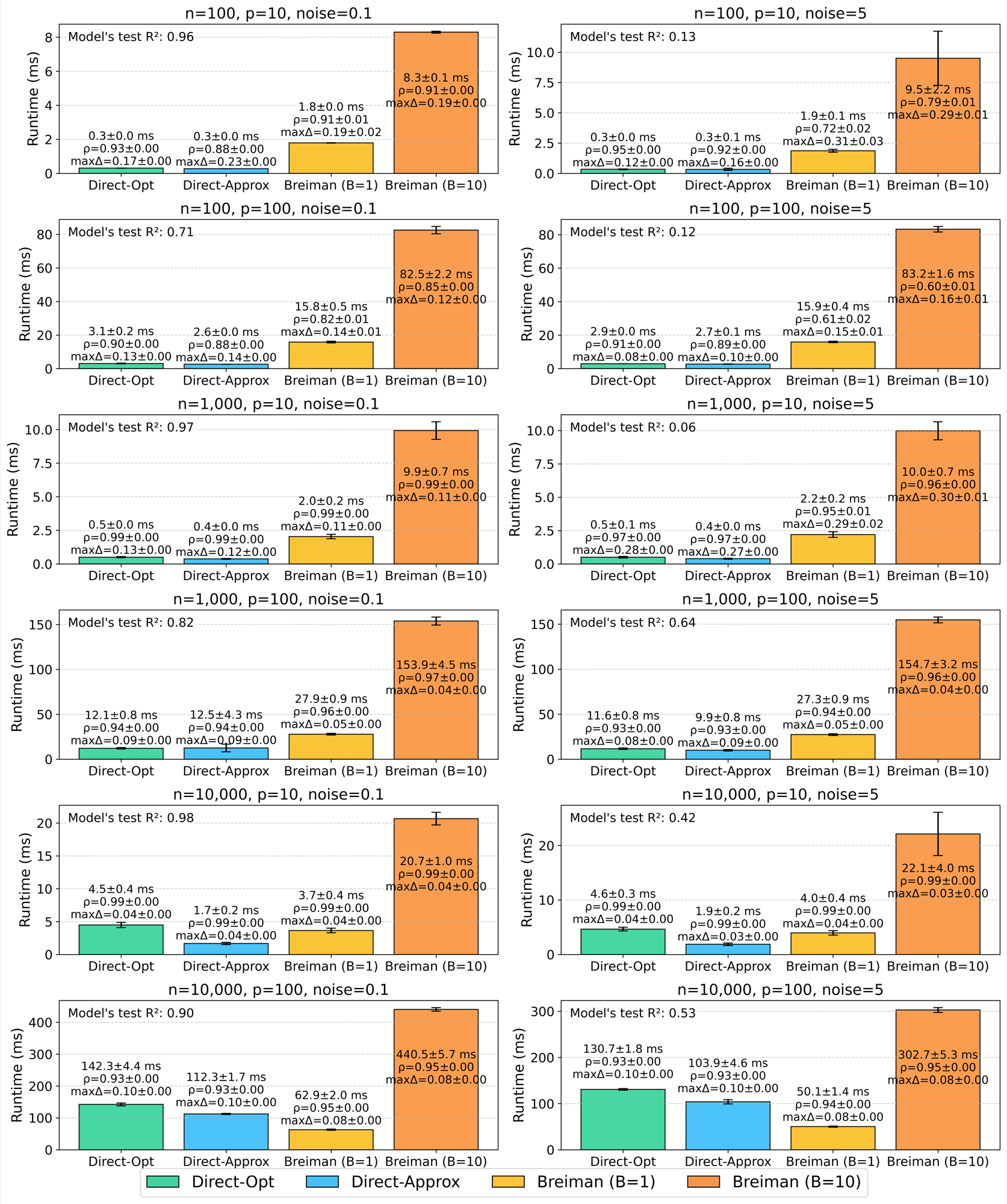}
\caption{Performance of different variable importance methods in $\ell_1$-linear regression with correlated features and linear response under varying sample size, dimensionality and noise}
\label{fig:VIS_Lasso_corr03}
\end{figure}

\begin{figure}[!h]
\centering
\includegraphics[width=1\linewidth]{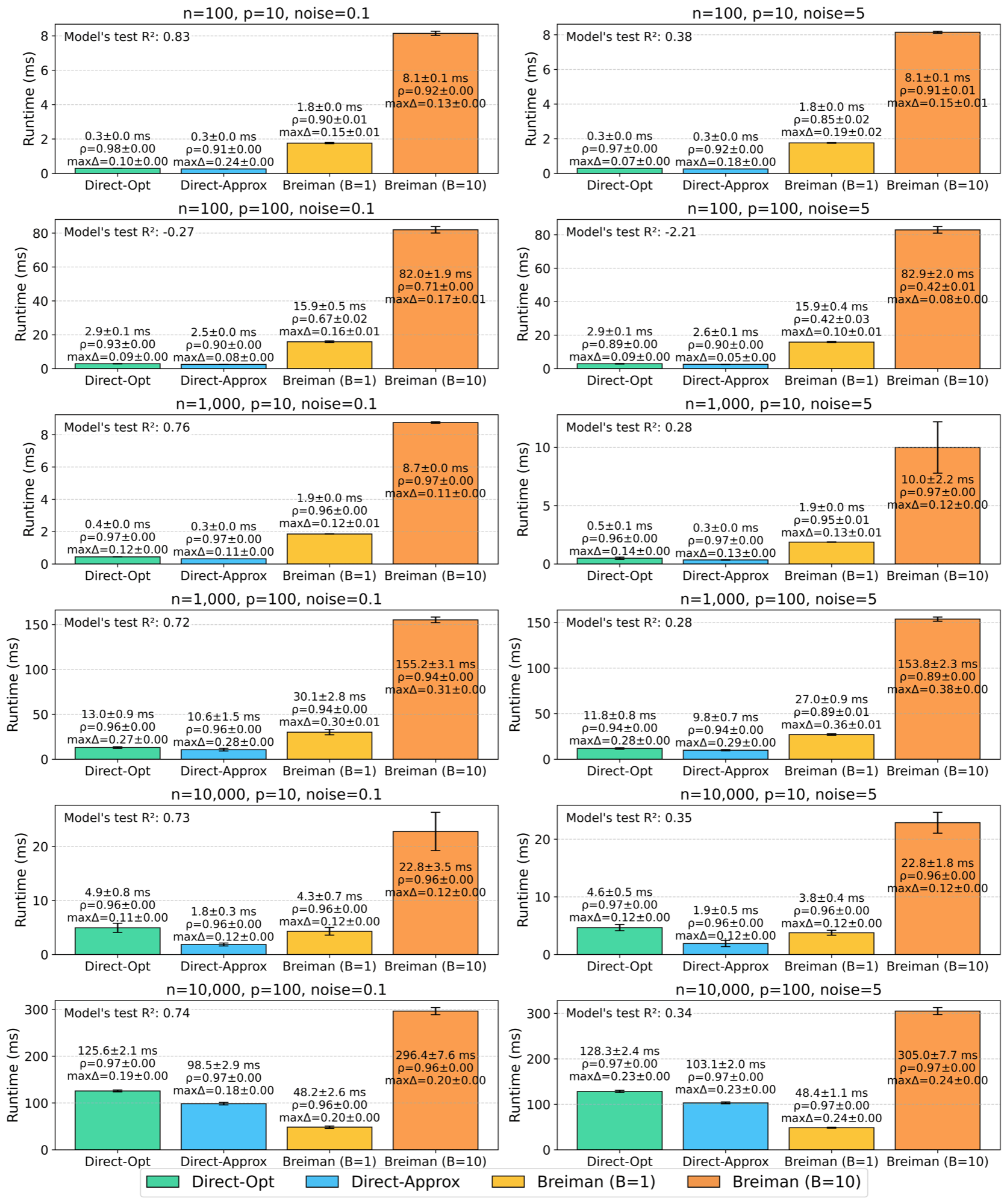}
\caption{Performance of different variable importance methods in linear regression with uncorrelated features and nonlinear response under varying sample size, dimensionality and noise}
\label{fig:VIS_lm_uncorrelated_NL}
\end{figure}

\begin{figure}[!h]
\centering
\includegraphics[width=1\linewidth]{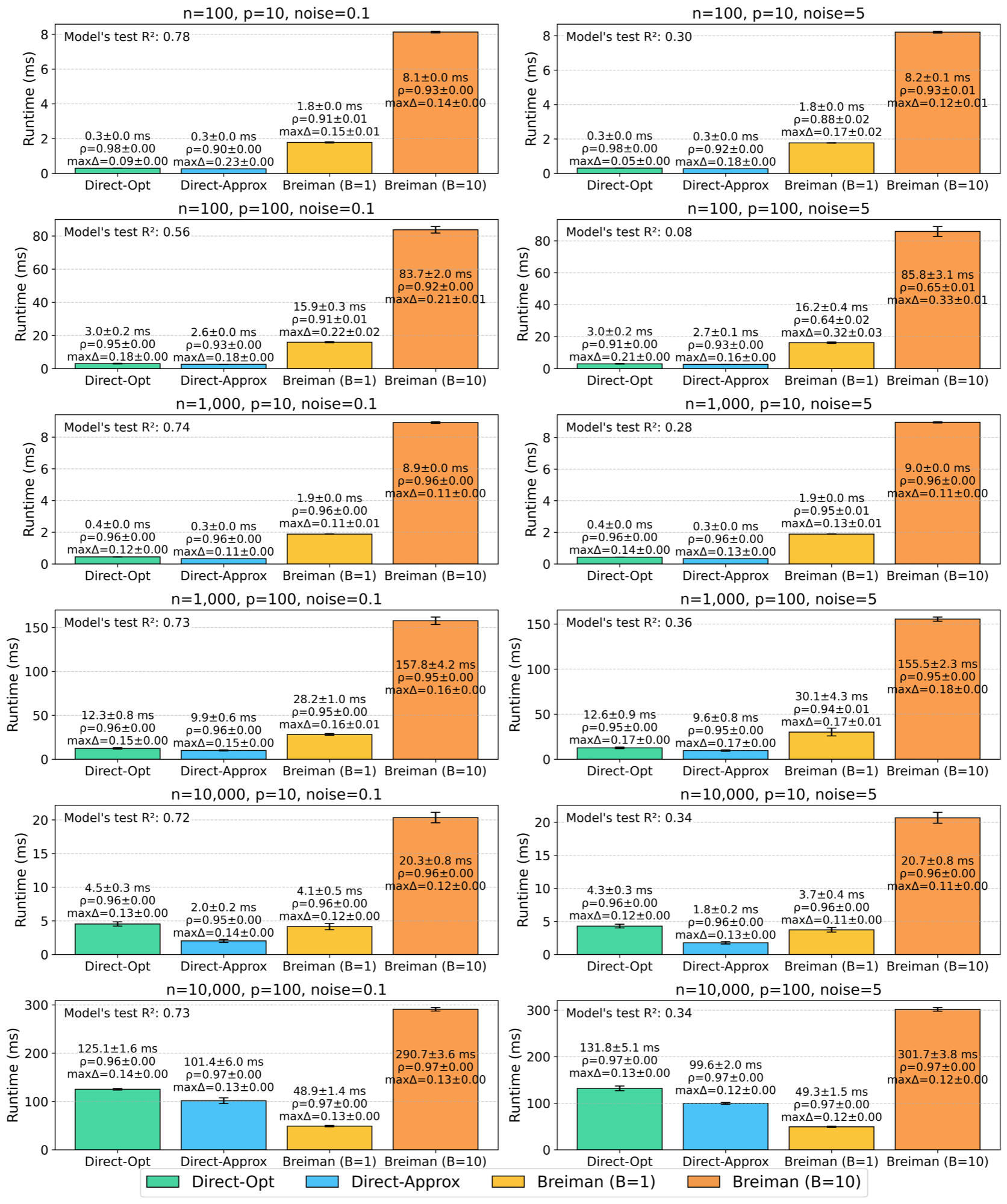}
\caption{Performance of different variable importance methods in $\ell_1$-linear regression with uncorrelated features and nonlinear response under varying sample size, dimensionality and noise}
\label{fig:VIS_Lasso_uncorrelated_NL}
\end{figure}

\begin{figure}[!h]
\centering
\includegraphics[width=1\linewidth]{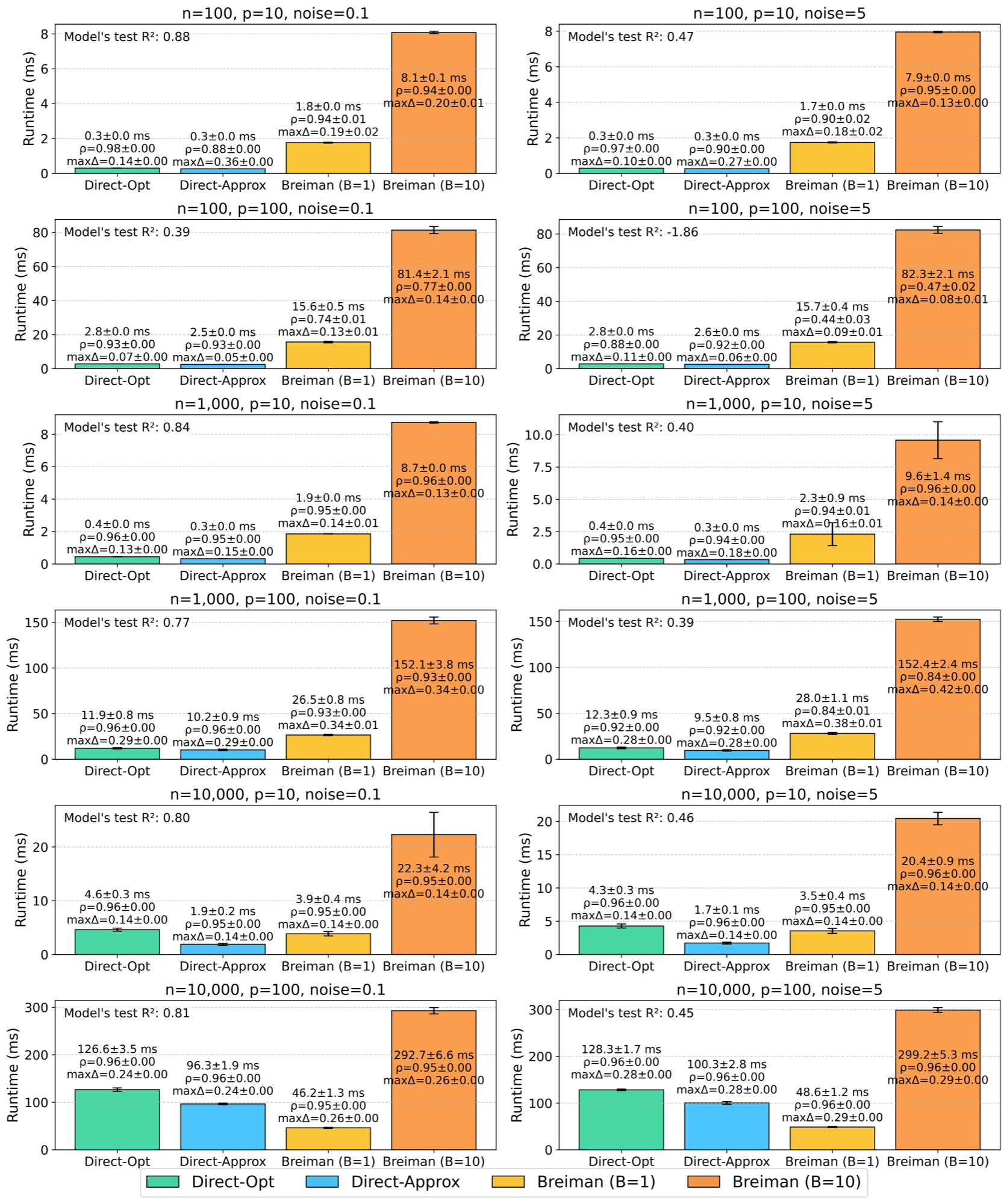}
\caption{Performance of different variable importance methods in linear regression with correlated features and nonlinear response under varying sample size, dimensionality and noise}
\label{fig:VIS_lm_corr03_NL}
\end{figure}

\begin{figure}[!h]
\centering
\includegraphics[width=1\linewidth]{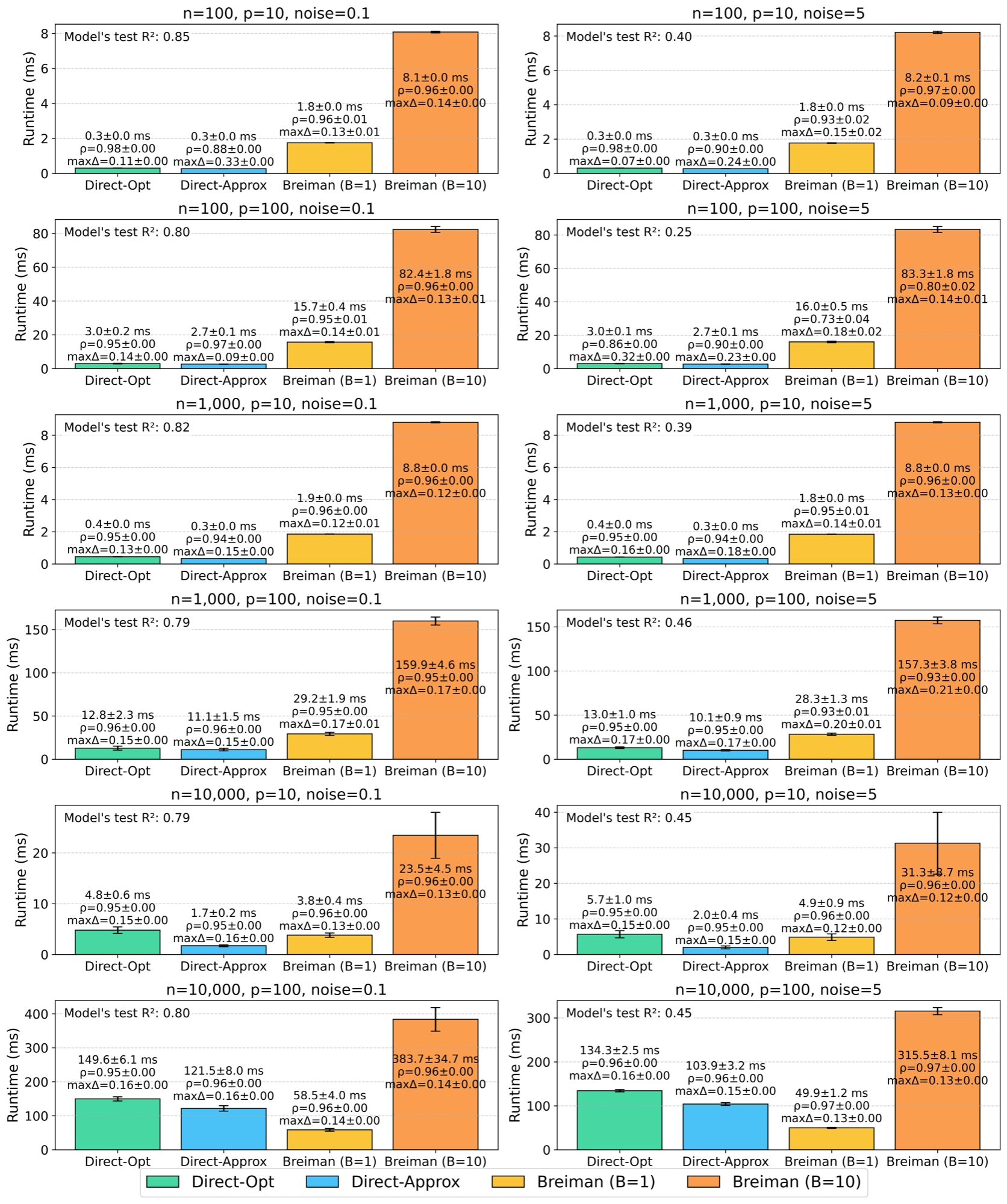}
\caption{Performance of different variable importance methods in $\ell_1$-linear regression with correlated features and nonlinear response under varying sample size, dimensionality and noise}
\label{fig:VIS_Lasso_corr03_NL}
\end{figure}

%CLASSIFICATION

\begin{figure}[!h]
\centering
\includegraphics[width=1\linewidth]{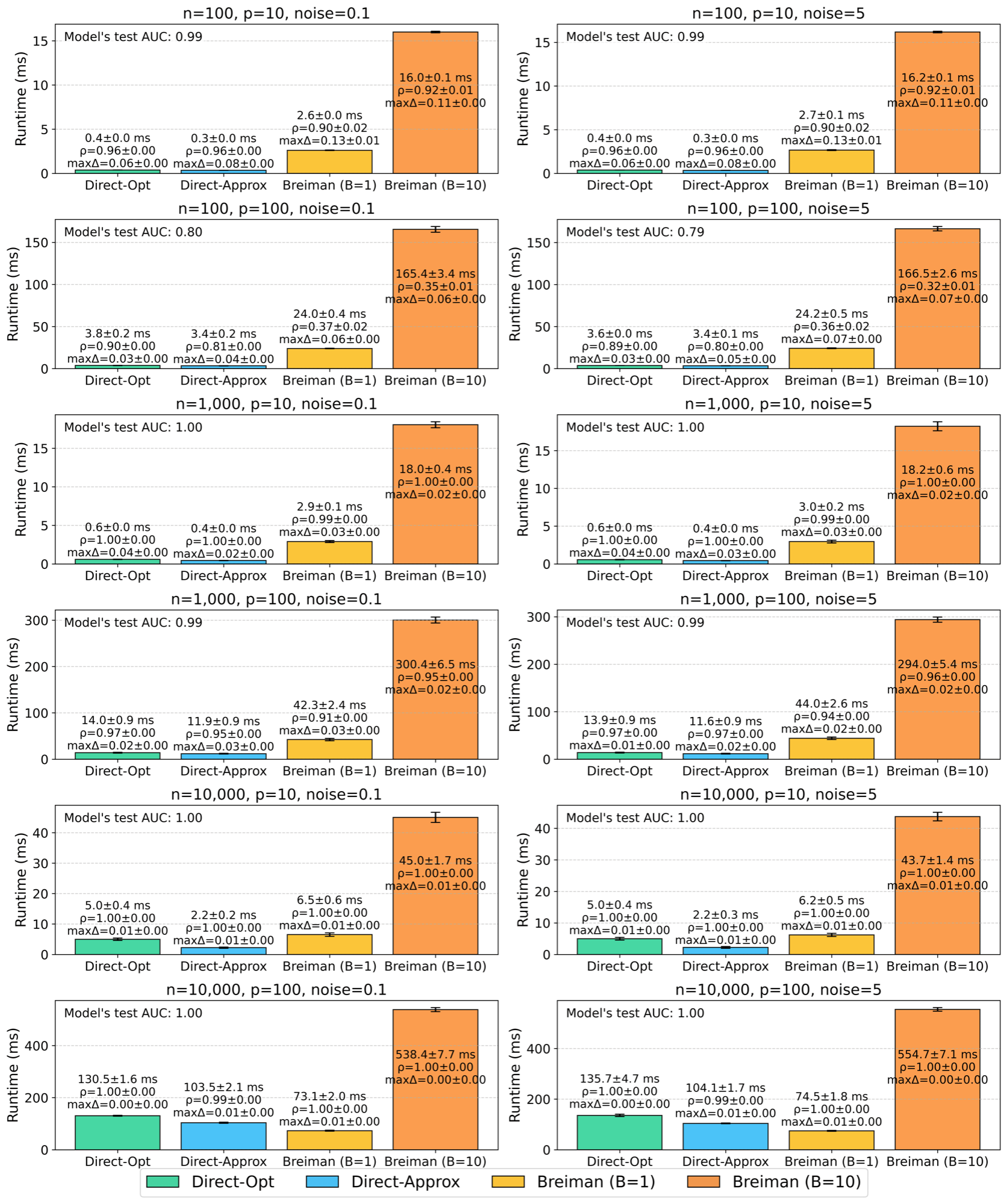}
\caption{Performance of different variable importance methods in logistic regression with uncorrelated features and linear (binarized) response under varying sample size, dimensionality and noise}
\label{fig:VIS_lm_uncorrelated_CLASS}
\end{figure}

\begin{figure}[!h]
\centering
\includegraphics[width=1\linewidth]{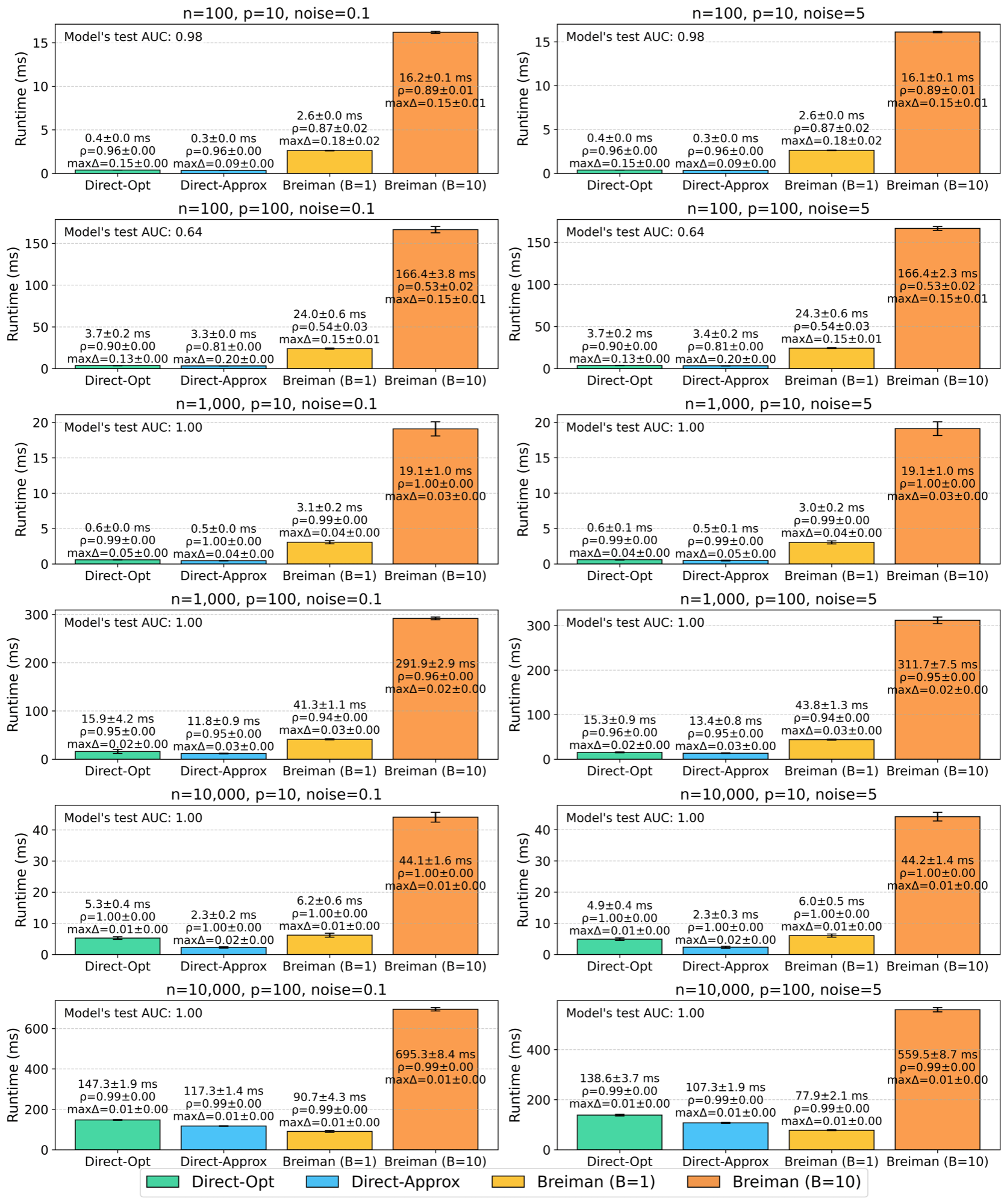}
\caption{Performance of different variable importance methods in $\ell_1$-logistic regression with uncorrelated features and linear (binarized) response under varying sample size, dimensionality and noise}
\label{fig:VIS_Lasso_uncorrelated_CLASS}
\end{figure}

\begin{figure}[!h]
\centering
\includegraphics[width=1\linewidth]{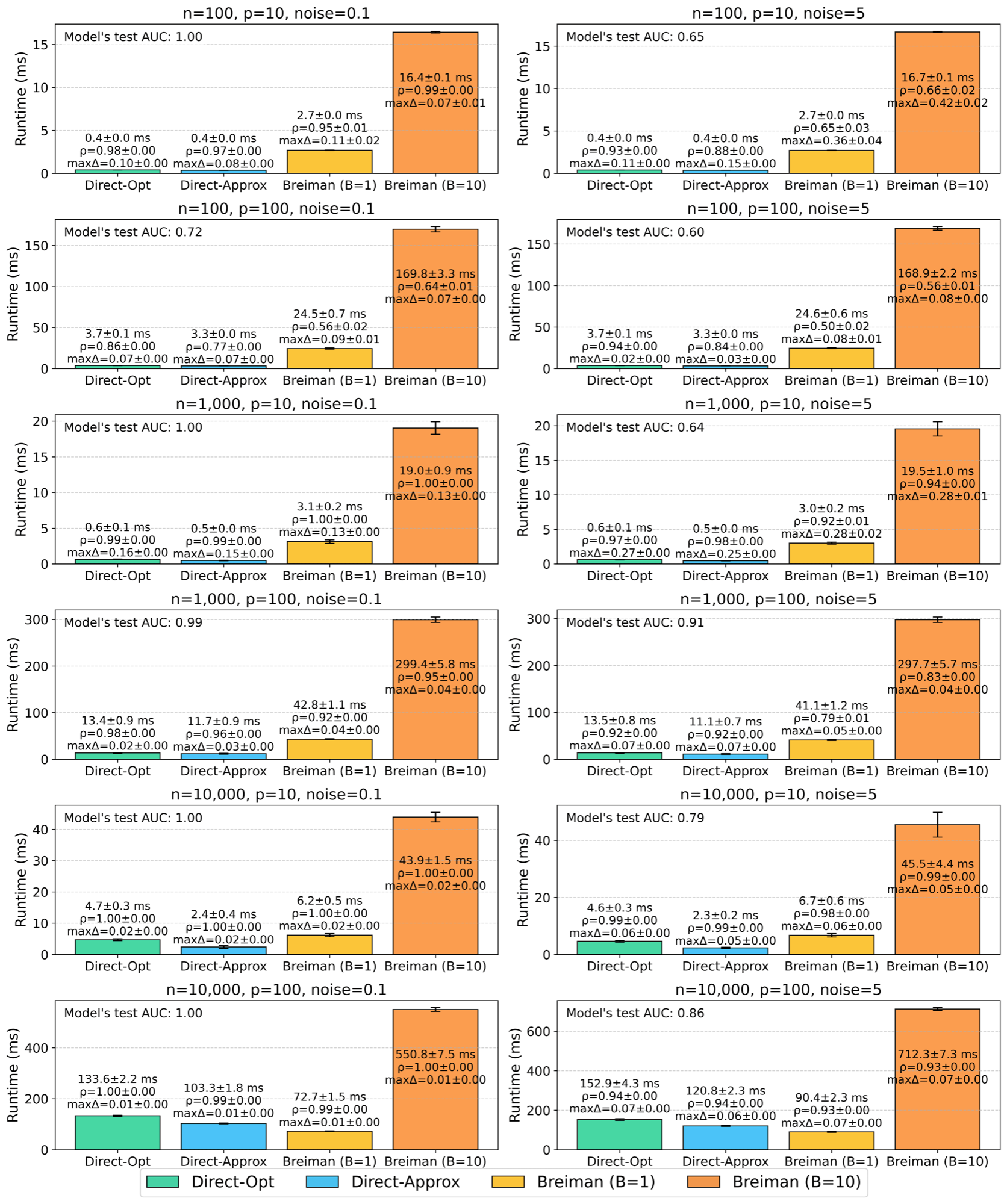}
\caption{Performance of different variable importance methods in logistic regression with correlated features and linear (binarized) response under varying sample size, dimensionality and noise}
\label{fig:VIS_lm_corr03_CLASS}
\end{figure}

\begin{figure}[!h]
\centering
\includegraphics[width=1\linewidth]{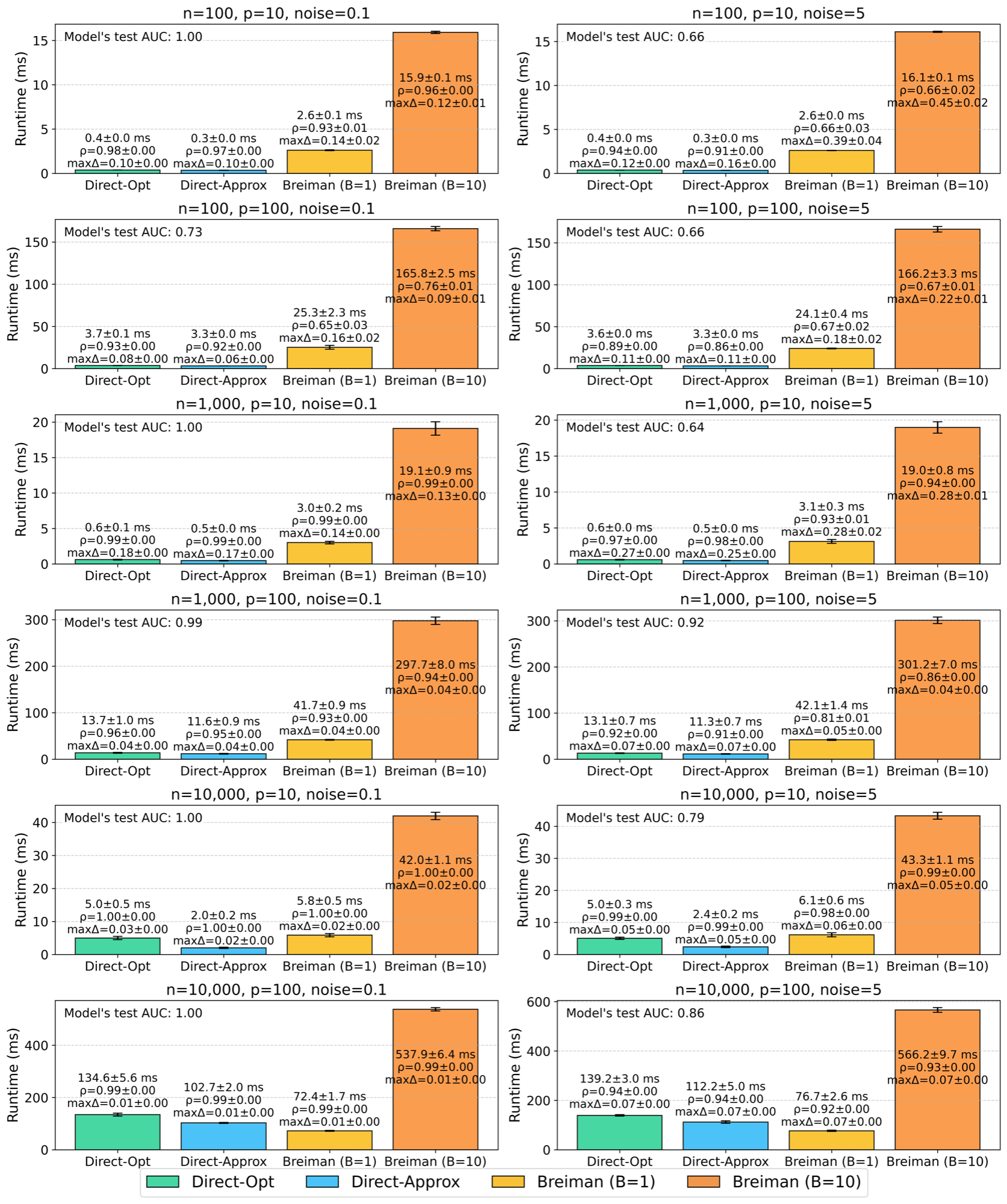}
\caption{Performance of different variable importance methods in $\ell_1$-logistic regression with correlated features and linear (binarized) response under varying sample size, dimensionality and noise}
\label{fig:VIS_Lasso_corr03_CLASS}
\end{figure}

\begin{figure}[!h]
\centering
\includegraphics[width=1\linewidth]{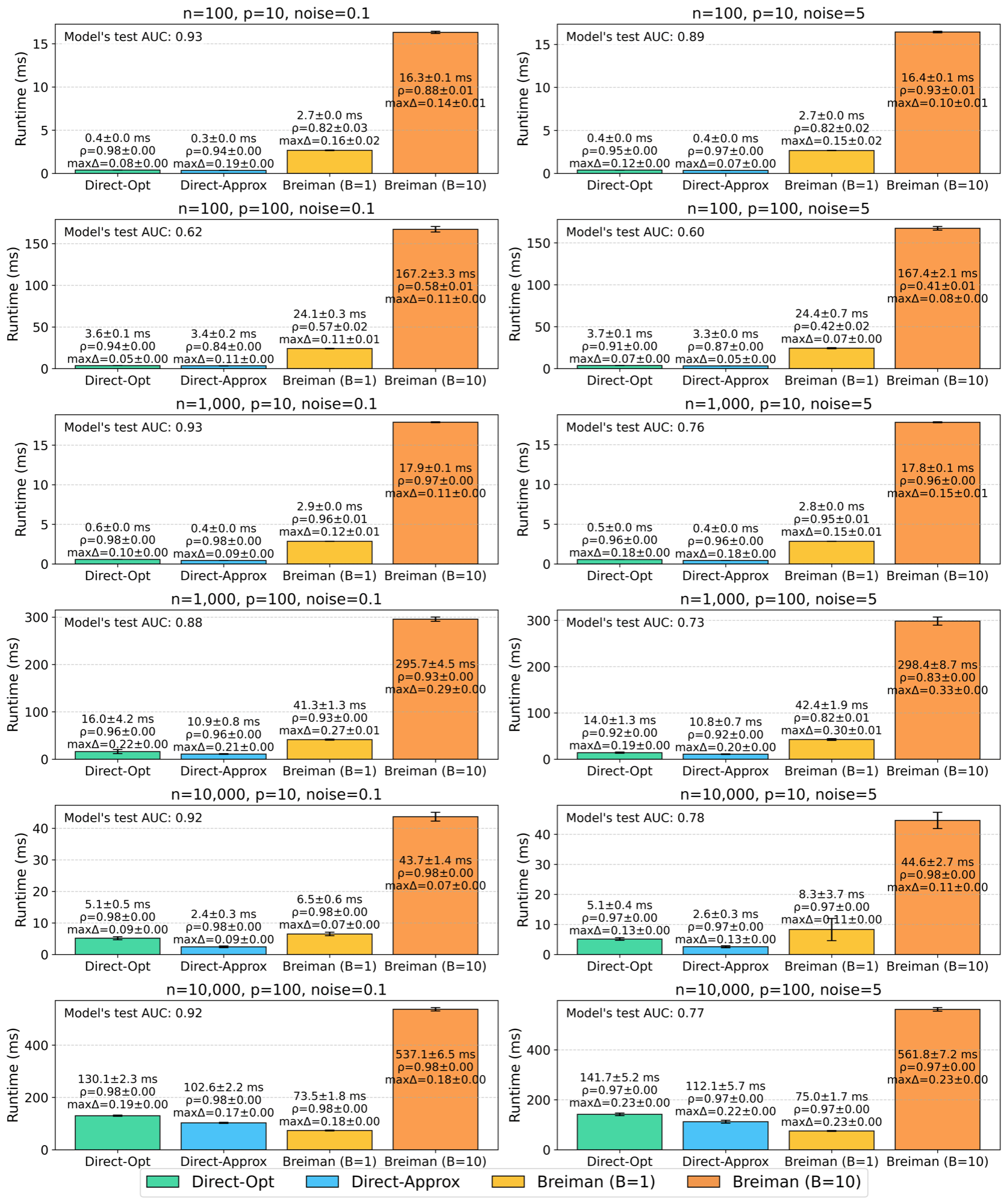}
\caption{Performance of different variable importance methods in logistic regression with uncorrelated features and nonlinear (binarized) response under varying sample size, dimensionality and noise}
\label{fig:VIS_lm_uncorrelated_NL_CLASS}
\end{figure}

\begin{figure}[!h]
\centering
\includegraphics[width=1\linewidth]{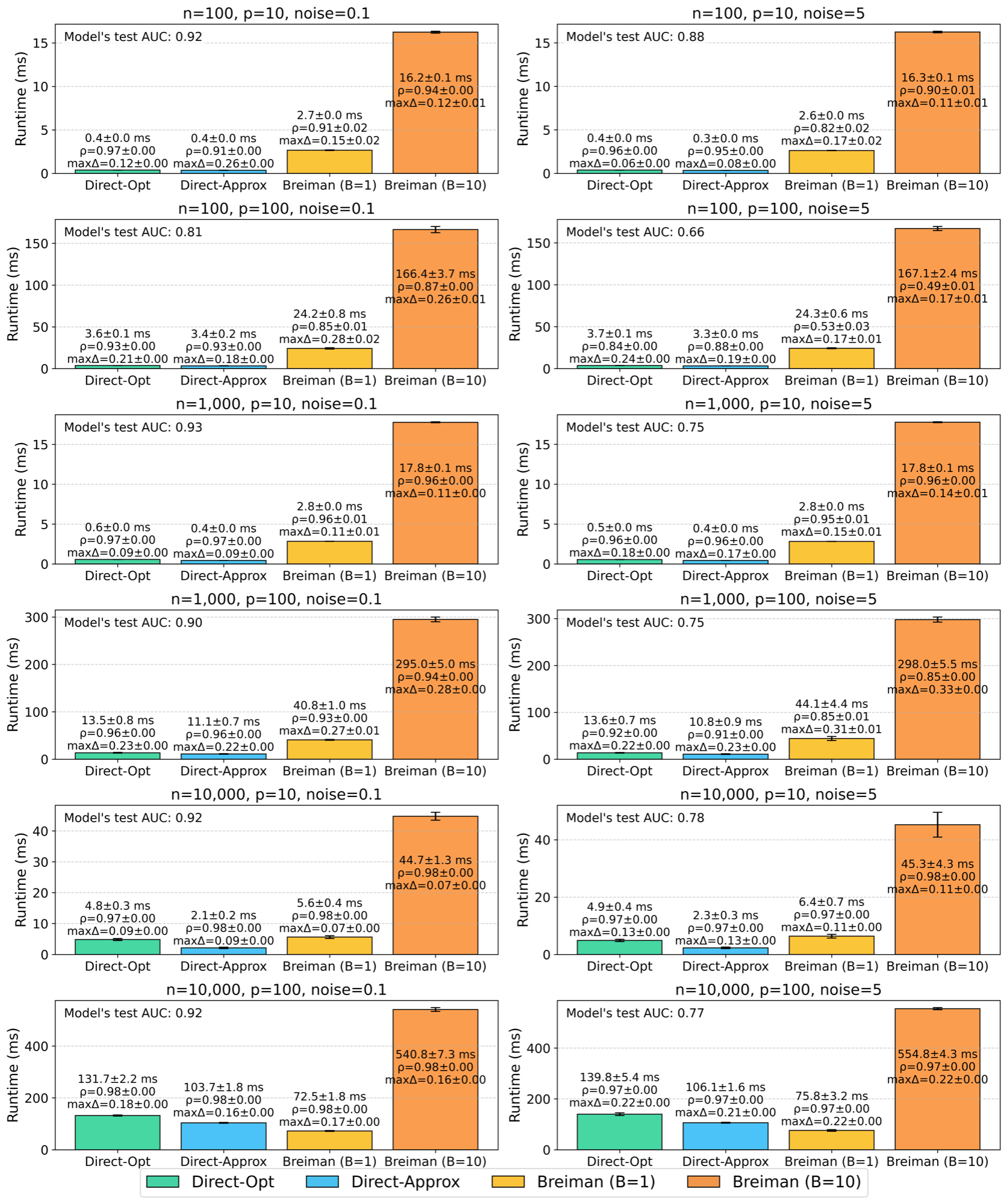}
\caption{Performance of different variable importance methods in $\ell_1$-logistic regression with uncorrelated features and nonlinear (binarized) response under varying sample size, dimensionality and noise}
\label{fig:VIS_Lasso_uncorrelated_NL_CLASS}
\end{figure}

\begin{figure}[!h]
\centering
\includegraphics[width=1\linewidth]{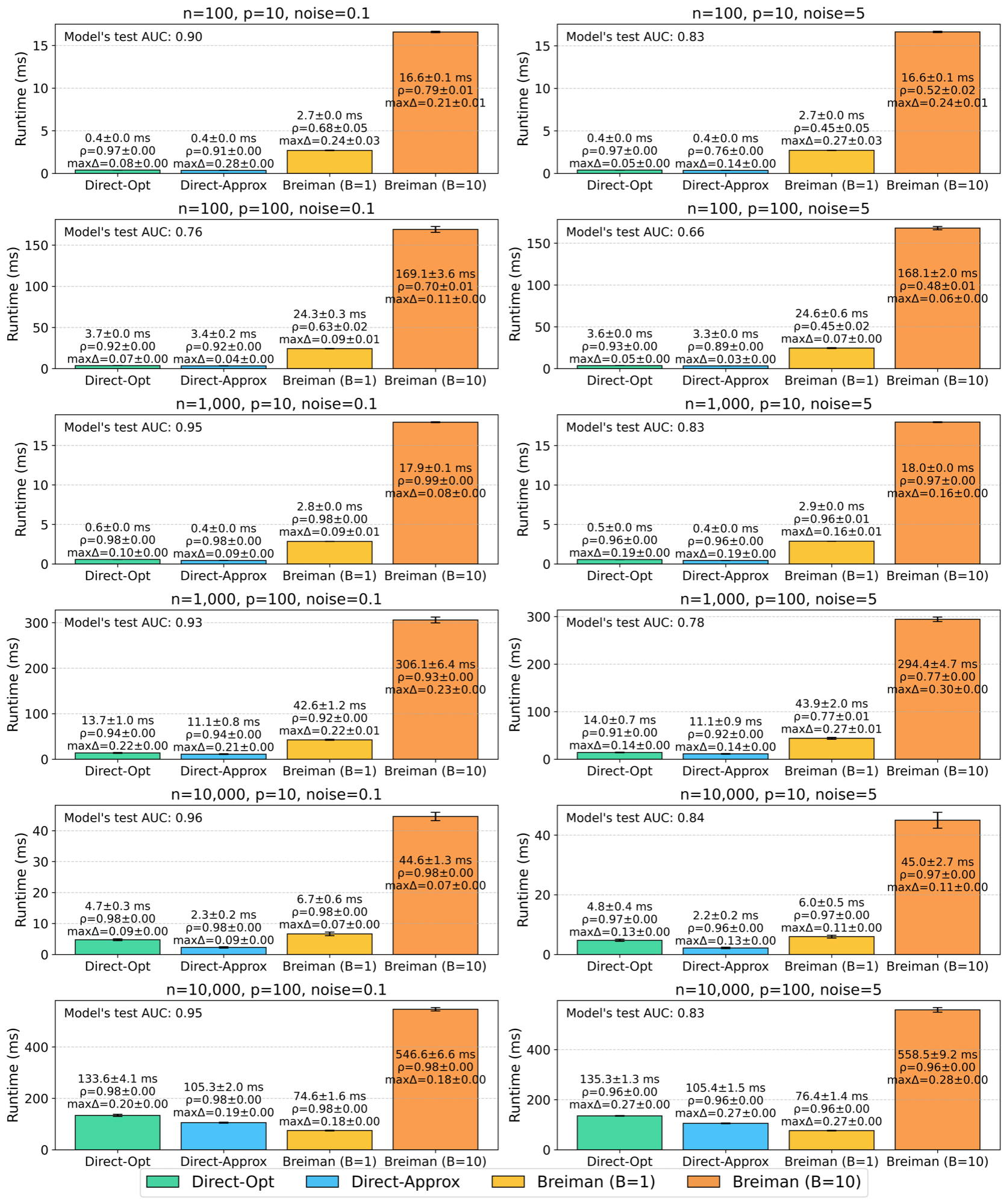}
\caption{Performance of different variable importance methods in logistic regression with correlated features and nonlinear (binarized) response under varying sample size, dimensionality and noise}
\label{fig:VIS_lm_corr03_NL_CLASS}
\end{figure}

\begin{figure}[!h]
\centering
\includegraphics[width=1\linewidth]{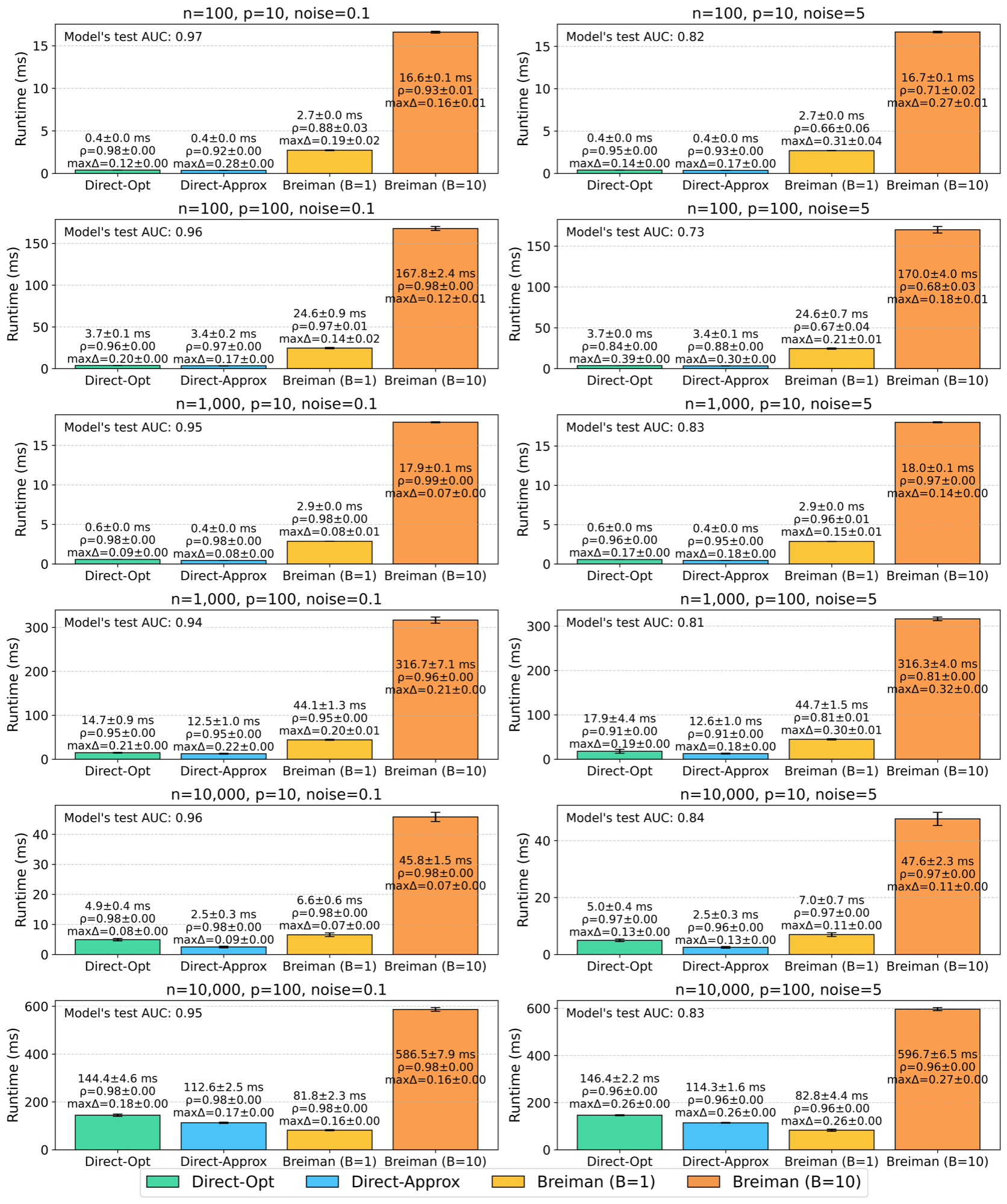}
\caption{Performance of different variable importance methods in $\ell_1$-logistic regression with correlated features and nonlinear (binarized) response under varying sample size, dimensionality and noise}
\label{fig:VIS_Lasso_corr03_NL_CLASS}
\end{figure}

\end{document}